\documentclass[preprint, 3p, 12pt]{elsarticle}

\usepackage{amssymb}
\usepackage{amsmath}
\usepackage{algorithm}%
\usepackage{algorithmicx}%
\usepackage{algpseudocode}%
\usepackage{booktabs}%
\usepackage{multirow}
\usepackage[caption=false,font=footnotesize]{subfig}
\usepackage{float}
\usepackage{xcolor}
\usepackage[hyphens]{url}

\newdefinition{definition1}{Analysis}

\journal{Applied Soft Computing}

\begin{document}

\begin{frontmatter}

\title{A State Alignment-Centric Approach to Federated System Identification: The FedAlign Framework}

\author[itu]{Ertu\u{g}rul Ke\c{c}eci}
\ead{kececie@itu.edu.tr}

\author[itu]{M\"{u}jde G\"{u}zelkaya}
\ead{guzelkaya@itu.edu.tr}

\author[ai]{Tufan Kumbasar\corref{cor1}}
\ead{kumbasart@itu.edu.tr}

\affiliation[itu]{
    organization={Faculty of Electrical and Electronics Engineering, Istanbul Technical University},
    city={Istanbul},
    country={Türkiye}
}

\affiliation[ai]{
    organization={Artificial Intelligence and Intelligent Systems Laboratory, Istanbul Technical University},
    city={Istanbul},
    country={Türkiye}
}

\cortext[cor1]{Corresponding author}

\begin{abstract}
This paper presents FedAlign, a Federated Learning (FL) framework, designed for System Identification (SYSID) of linear State-Space Models (SSMs) by aligning state representations. Local workers can learn linear SSMs with equivalent representations but different parameter basins. We demonstrate that directly aggregating these local SSMs via FedAvg results in a global model with altered system dynamics. FedAlign overcomes this problem by employing similarity transformation matrices to align state representations of local SSMs, thereby establishing a common parameter basin that retains the dynamics of local SSMs. FedAlign computes similarity transformation matrices via two distinct approaches. In FedAlign-A, we represent the global SSM in controllable canonical form (CCF). We use control theory to analytically derive similarity transformation matrices that convert each local SSM into this form. Yet, establishing global SSM in CCF brings additional alignment challenges in multi-input multi-output SYSID, as CCF representation is not unique, unlike in single-input single-output SYSID. In FedAlign-O, we address the alignment challenges by reformulating the local parameter basin alignment problem as an optimization task. We set the parameter basin of a local worker as the common parameter basin and solve least square problems to obtain the transformation matrices needed to align the remaining local SSMs. The experiments conducted on synthetic and real-world datasets show that FedAlign outperforms FedAvg, converges faster, and provides improved global SSM stability thanks to local parameter basins' alignment.
\end{abstract}




\end{frontmatter}



\section{Introduction}\label{sec1}

The primary objective of the System Identification (SYSID) problem is to represent system behaviors based on time-series data by estimating a model \cite{ljung2010perspectives, keesman2011system}. Although recent research has begun integrating machine learning models into the SYSID to represent complex system dynamics \cite{chiuso2019system, pillonetto2025deep}, State Space Model (SSM) estimation techniques remain popular in SYSID. This is largely due to their reliance on linear relationships and the availability of robust, well-established estimation methods \cite{dankers2016, zhou2006}. Historically, SYSID studies have used data from a single source, but emerging studies indicate that sample efficiency can be boosted by employing multiple data sources from systems with analogous dynamics \cite{xin2022identifying, zhang2023multi, papusha2014collaborative}.

Federated Learning (FL) facilitates the training of a global model through decentralized clients, each utilizing their private data to train local models, while these models are merged and redistributed within the center server. FL naturally offers a suitable framework specifically for SYSID tasks, involving multiple time-series data collected from similar systems in distributed environments. The work \cite{wang2023fedsysid} introduced the application of FL to SYSID, showing promising results under data homogeneity. Yet, in practical scenarios, local systems often exhibit heterogeneity in their dynamics, which limits the effectiveness of conventional FL methods such as FedAvg. To address this, several clustered FL approaches have been proposed. \cite{toso2023} tackles system heterogeneity using prior dataset knowledge to form clusters, while \cite{kececi2024} employs incremental clustering without requiring such knowledge. Moreover, in FL-SYSID, even when clients observe similar dynamics, locally trained SSMs can differ significantly due to variations in state coordinate systems. This results in dynamically equivalent but misaligned representations, which can severely degrade the quality of global model aggregation. While recent advances such as \cite{ainsworth2023git} have explored representation alignment in neural networks, this issue remains unaddressed in the context of FL-SYSID. It is worth underlining that general-purpose FL variants like FedProx \cite{li2020federated}, FedAvgM \cite{hsu2019measuring}, and FedFusion \cite{fedfusion} aim to mitigate data heterogeneity and client drift through algorithmic enhancements. Yet, these approaches do not consider a key structural challenge in SYSID: the alignment of internal state representations across independently trained local models.

In this paper, we introduce FedAlign, an FL framework tailored for SYSID tasks by aligning state representations of linear SSMs. We show that direct aggregation of local SSMs with equivalent representations, i.e., different parameter basins with identical dynamics, via FedAvg leads to distorted global SSM dynamics due to misaligned local parameter basins. FedAlign addresses this problem by aligning state representations of local linear SSMs. FedAlign establishes a common parameter basin for the global SSM in the center server and utilizes similarity transformation matrices to convert parameter basins of local SSMs into the common parameter basin. By aligning local parameter basins, FedAlign assures that the global SSM maintains the dynamics of local SSMs. FedAlign offers two distinct methods to compute similarity transformation matrices in the FL-SYSID framework:
\begin{itemize}
    \item FedAlign-A: We depict the global SSM within the controllable canonical form (CCF). We utilize well-established linear control theory to analytically derive similarity transformation matrices that convert local SSMs into CCF representation.  
    \item FedAlign-O: We address local parameter alignment challenges as an optimization-based problem. We set the parameter basin of a local worker as the common parameter basin. We solve least square problems using generated pseudo data to compute similarity transformation matrices, aligning remaining local SSMs with the common parameter basin. It is important to note that FedAlign-O does not determine any strict form for the global SSM.
\end{itemize}
For both FedAlign-A and FedAlign-O, we present all the design details for solving the SYSID problems of Single-Input Single-Output (SISO) and Multi-Input Multi-Output (MIMO) systems. While using MIMO data, FedAlign-A provides additional alignment challenges as the CCF is not unique for MIMO systems, in contrast to SISO systems. However, FedAlign-O mitigates these challenges by not strictly forcing the CCF representation for the global SSM. 

To validate the proposed FL-SYSID frameworks, we present comprehensive comparative results on synthetic and real-world SYSID datasets. We start by analyzing the two possible alignment challenges faced during SYSID problems. Through experiments carried out on the synthetic SISO dataset, we evaluated how local SYSID performance impacts FL-SYSID effectiveness. Moreover, we assess the impact of different creations of similarity transformation matrices on the SYSID performance of FedAlign by using two synthetic MIMO datasets. Finally, we compare the SYSID performances of FedAlign and FedAvg on the real-world SISO and MIMO datasets, illustrating that FedAlign achieves higher performance while converging faster and improving global SSM's stability thanks to effective local parameter basin alignment.

The paper is organized as follows: Section \ref{prelim} gives an overview of SYSID. Section \ref{align} addresses the alignment issues within the FL-SYSID framework by using similarity transformation matrices. Section \ref{fedalign} introduces the complete proposed FedAlign framework, including the design steps of FedAlign-A and FedAlign-O. Section \ref{exp} provides comprehensive comparative analyses of FedAlign. Section \ref{conc} presents the drawn conclusions as well as suggestions for future work.

\section{Problem Definition of System Identification} \label{prelim}

Consider the following nonlinear system
\begin{equation}
\label{nonlinear}
\begin{aligned}
        \boldsymbol{x}_{k+1} &= f(\boldsymbol{x}_k,\boldsymbol{u}_k,\boldsymbol{w}_k) \\
        \boldsymbol{y}_k &= h(\boldsymbol{x}_k,\boldsymbol{u}_k,\boldsymbol{v}_k), \quad k=1,2,\dots,K
\end{aligned}
\end{equation}
where $\boldsymbol{x}_k = (x_{1,k}, \dots, x_{nx,k})^T$ represents the state vector, $\boldsymbol{u}_k = (u_{1,k}, \dots, u_{nu,k})^T$ denotes the input vector, $\boldsymbol{y}_k = (y_{1,k}, \dots, y_{ny,t})^T$ refers the output vector with $\boldsymbol{w}_k = (w_{1,k}, \dots, w_{nx,t})^T$ and $\boldsymbol{v}_k = (v_{1,k}, \dots, v_{ny,t})^T$ being input and output noises, respectively. The functions $f(\cdot)$ and $h(\cdot)$ capture the nonlinearities.

In an SYSID problem with a dataset $D = \{\boldsymbol{u},\boldsymbol{y}\}$, \eqref{nonlinear} can be approximated by an SSM as
\begin{equation} \label{LTI}
\begin{aligned}
    \tilde{\boldsymbol{x}}_{k+1} &= \tilde{A}\tilde{\boldsymbol{x}}_{k} + \tilde{B}\boldsymbol{u}_{k}, \quad \tilde{\boldsymbol{x}}_1 = \boldsymbol{x}_1 \\
    \tilde{\boldsymbol{y}}_k &= \tilde{C}\tilde{\boldsymbol{x}}_{k} + \tilde{D}\boldsymbol{u}_{k}, \quad k=1,2,\dots,K
\end{aligned}
\end{equation}
where $\boldsymbol{\tilde{x}_k}$ refers estimated states while  $\boldsymbol{\tilde{y}_k}$ refers estimated outputs. In \eqref{LTI}, the state, input, output and feedthrough matrices are denoted by $\tilde{A} \in \mathbb{R}^{nx \times nx}$, $\tilde{B} \in \mathbb{R}^{nx \times nu}$, $\tilde{C} \in \mathbb{R}^{ny \times nx}$, and $\tilde{D} \in \mathbb{R}^{ny \times nu}$, respectively. Prediction Error Minimization (PEM) or N4SID methods \cite{dankers2016, zhou2006} can be utilized to identify these system matrices, given as $\tilde{\Theta} = \{\tilde{A}, \tilde{B}, \tilde{C}, \tilde{D}\}$. For a comprehensive discussion on SYSID, we direct readers to \cite{ljung2010perspectives}.

\section{The Alignment Problem of FL-SYSID} \label{align}

We begin by showing that in an FL framework, the aggregation process of local SSMs with different parameter basins leads to the global model, exhibiting altered system dynamics. Following that, we demonstrate that a common-parameter basin can be formed by aligning state representations of local SSMs leveraging similarity transformation matrices. When local SSMs are aggregated in this common parameter basin, inherent system characteristics of local SSMs are maintained in the global SSM.

\subsection{The Impact of Local Parameter Basin Differences}

An FL-SYSID framework involves $M$ decentralized local workers,
learning SYSID tasks using their private datasets $\boldsymbol{D} = \{D^1, D^2, \dots, D^M\}$. Following local training at each communication round, local workers send their models to the center server. These local SSMs are aggregated via well-known FedAvg \cite{mcmahan2017communication} to calculate the global SSM as follows: 
\begin{equation}
\label{directavg}
    \tilde{\Theta} = \frac{1}{M}\sum_{i=1}^M \tilde{\Theta}^{(i)} 
\end{equation}
where $\tilde{\Theta}^{(i)}$ and $\tilde{\Theta}$ denote system matrices of local SSM for $W^i$ and system matrices for the global SSM, respectively. After the aggregation process, local workers receive $\tilde{\Theta}$ from the center server.

The efficiency of FedAvg relies on the assumption that local workers can learn system dynamics represented with similar or even the same parameter basins from their private datasets. Nevertheless, given the inherent nonuniqueness of SSMs in \eqref{LTI}, an SSM can be expressed in equivalent representations exhibiting identical dynamics with distinct parameter basins, e.g., Controllable Canonical Form (CCF) and Observable Canonical Form (OCF). If local SSMs are in these equivalent representations, directly averaging these models as in FedAvg, may result in a global SSM with different system properties, such as system gain, time-domain response, eigenvalues, controllability, and observability, potentially leading to an unstable model. 

\begin{definition1} \label{Init}
Consider a FedAvg framework with two local workers to demonstrate how different parameter basins affect the global SSM dynamics. Assume local SSMs of $W^1$ and $W^2$ are in equivalent representations but with different parameter basins, $W^1$ in CCF and $W^2$ in OCF. Their state matrices are expressed as:
\begin{equation}
\begin{aligned}
    \tilde{A}^{(1)}& = \left[\begin{array}{ccc}
       0  & 1 & 0 \\
       0  & 0 & 1 \\
       -a_3 & -a_2 & -a_1
    \end{array}\right],  \tilde{A}^{(2)} = \left[\begin{array}{ccc}
       0 & 0 & -a_3 \\
       1 & 0 & -a_2 \\
       0 & 1 & -a_1
    \end{array}\right]
\end{aligned}
\end{equation}
where the coefficients $a_1$, $a_2$, and $a_3$ define the characteristic polynomial
\begin{align}
    p(\lambda)=\det (\lambda I-A)=0,
\end{align}
whose roots are the eigenvalues ($\lambda$) of the SSMs. When calculating the state matrix of the global SSM, $\tilde{A}$, through direct averaging:
\begin{align}
    \tilde{A} = \frac{\tilde{A}^{(1)} + \tilde{A}^{(2)}}{2}=\left[\begin{array}{ccc}
       0  & 0.5 & -\frac{a_3}{2} \\
       0.5  & 0 & \frac{(1-a_2)}{2} \\
       -\frac{a_3}{2} & \frac{(1-a_2)}{2} & -a_1
       \end{array}\right]
\end{align}
the characteristic equation of $\tilde{A}$ will differ from those of the local SSMs, causing a global SSM with altered dynamics. As a result, the global SSM may become unstable even when the individual local SSMs are stable, highlighting the risk of directly averaging equivalent SSMs with different parameter basins.\end{definition1}

Although \textbf{Analysis-1} illustrates an edge case, it clearly demonstrates the issues of aggregating equivalent local SSMs. It shows that FedAvg can obtain global SSM with altered dynamics and jeopardize the global SSM's stability. Therefore, a local parameter basin alignment process must be used before computing the global SSM via FedAvg.

\subsection{Aligning Local Parameter Basins}

A similar challenge for neural networks has been addressed in \cite{ainsworth2023git}. The authors focus on $L$-layer MLP defined as:
\begin{equation}
    f(x, \Theta) = z_{L+1}, \ z_{l+1} = \sigma(\Theta_{W_l}z_l + \Theta_{b_l}), \ z_1 = x.
\end{equation}
where $\Theta$ is the weights set and $\sigma$ refers to the activation function. They demonstrate that by applying a permutation matrix $P$ to the weights and biases,
\begin{equation}
    \Theta_{W_l}^{'} = P\Theta_{W_l}, \ \Theta_{b_l}^{'} = P\Theta_{b_l}, \ \Theta_{W_{l+1}}^{'} = \Theta_{W_{l+1}}P^T
\end{equation}
the transformed model, $\Theta'$, remains functionally equivalent, i.e. $f(x,\Theta) = f(x,\Theta')$. The paper introduces various techniques to calculate $P$ that align different parameter basins into a common parameter basin. However, these methods are unsuitable for SSMs, as aligning state representations is necessary when transforming different parameter basins. 

\textbf{Analysis 1} depicted that the global SSM exhibits altered system dynamics when local SSMs represented through different parameter basins are aggregated. Aligning state representations of local SSMs is a must to form a common parameter basin. Instead of focusing on aligning local parameters, similarity transformation between state vectors of local SSMs can be utilized. A state vector $\boldsymbol{x}$ is transformed to alternative representation by
\begin{equation}
    \boldsymbol{x} = T \boldsymbol{x'}
\end{equation}
where $T$ and $\boldsymbol{x'}$ are similarity transformation matrix and transformed state vector, respectively.

\begin{definition1} \label{Init2} 
In the same FL setup from Analysis 1, where local SSMs exhibit identical dynamics, a linear transformation between $\boldsymbol{\tilde{x}}^{(2)}$ and $\boldsymbol{\tilde{x}}^{(1)}$ is defined by
\begin{equation} \label{simT}
    \boldsymbol{\tilde{x}}^{(2)} = T\boldsymbol{\tilde{x}}^{(1)}
\end{equation}
with $T$ being any nonsingular matrix, known as similarity transformation \cite{wisa2024} (Section \ref{fedalign} will introduce two distinct methods to compute $T$). A common parameter basin is formed by aligning state vectors with \eqref{simT}, linking $\tilde{\Theta}^{(1)}$ and $\tilde{\Theta}^{(2)}$ through the following equations.
\begin{equation}
\label{matsim}
\begin{aligned}
    \tilde{A}^{(1)} &= T^{-1}\tilde{A}^{(2)}T, \quad \tilde{B}^{(1)} = T^{-1}\tilde{B}^{(2)} \\
    \tilde{C}^{(1)} &= \tilde{C}^{(2)}T, \quad  \tilde{D}^{(1)} = \tilde{D}^{(2)}
\end{aligned}
\end{equation}
Subsequently, the global SSM, $\tilde{\Theta} = \{\tilde{A}, \tilde{B}, \tilde{C}, \tilde{D}\}$ is calculated as follows:
\begin{equation}
\begin{aligned}
\tilde{A} &= \frac{\tilde{A}^{(1)} + T^{-1}\tilde{A}^{(2)}T}{2}, \quad \tilde{B} = \frac{\tilde{B}^{(1)} + T^{-1}\tilde{B}^{(2)}}{2}, \\
\tilde{C} &= \frac{\tilde{C}^{(1)} + \tilde{C}^{(2)}T}{2}, \quad \tilde{D} = \frac{\tilde{D}^{(1)} + \tilde{D}^{(2)}}{2}.
\end{aligned}    
\end{equation}
The resulting state matrix for the global SSM takes the following form:
\begin{align} 
    \tilde{A}  = \left[\begin{array}{ccc}
       0  & 1 & 0 \\
       0  & 0 & 1 \\
       -a_3 & -a_2 & -a_1
       \end{array}\right],
\end{align}
We illustrated that the CCF representation on the common parameter basin is maintained for the global SSM. In contrast to the merging by direct averaging, as in \eqref{directavg}, the global SSM obtained with \eqref{matsim} preserves the same dynamics as local SSMs. Furthermore, the global SSM maintains identical eigenvalues as the local SSMs, thereby guaranteeing stability if the local SSMs are stable thanks to the state representation alignment of the local SSMs.
\end{definition1}

\section{FedAlign: The FL framework for SYSID tasks} \label{fedalign}

This section introduces the FedAlign framework that builds a global SSM by aligning state representations of local SSMs, ensuring the global SSM exhibits similar dynamics to those of local SSMs. FedAlign accomplishes this by leveraging similarity transformation matrices. During each communication round, the center server employs similarity transformations ($T$) to form a common parameter basin. It then computes a global SSM that retains the properties of local SSMs by averaging them within this common parameter basin.
\begin{equation}
\label{genericT}
\begin{aligned}
    \tilde{A} &= \frac{1}{M}\sum_{i=1}^M T_i^{-1}\tilde{A}^{(i)}T_i, \quad \tilde{B} = \frac{1}{M}\sum_{i=1}^M T_i^{-1}\tilde{B}^{(i)} \\
    \tilde{C} &= \frac{1}{M}\sum_{i=1}^M \tilde{C}^{(i)}T_i, \quad \tilde{D} = \frac{1}{M}\sum_{i=1}^M \tilde{D}^{(i)}
\end{aligned}
\end{equation}
At the end of each communication round, the center server converts the global SSM on the common parameter basin to the local basin of each local SSM with 
\begin{equation}
\label{genericInvT}
\begin{aligned}
    \tilde{A}^{(i)} &= T_i \tilde{A}T_i^{-1}, \quad \tilde{B}^{(i)} = T_i\tilde{B} \\
    \tilde{C}^{(i)} &= \tilde{C}T_i^{-1}, \quad \tilde{D}^{(i)} = \tilde{D} 
\end{aligned}
\end{equation}
Afterward, it sends the updated local SSMs to their respective local workers. Algorithm \ref{alg:alg1}\footnote[1]{MATLAB implementation. [Online]. Available: https://github.com/ertugrulkececi/fedalign-fl-sysid} provides a detailed overview of FedAlign's training procedure.

In this paper, we propose two methods for calculating $T$ within the FL-SYSID framework: 1) FedAlign-A, a data-free analytical approach, and 2) FedAlign-O, a data-driven optimization-based method. 

\begin{algorithm}[t]
\caption{FedAlign}\label{alg:alg1}
\begin{algorithmic}[1]
    \State \textbf{Initialization:} number of communication rounds $R$, local iterations $iter$, initialize local SSM parameters $\tilde{\Theta}^{(i)}_0$ for each worker $W^i$, $\forall i \in [M]$
    \State Choose alignment method: \textbf{FedAlign-A} or \textbf{FedAlign-O}
    \For{$r=0,1,2,\dots,R-1$}
        \State Server broadcasts current local SSM parameters $\tilde{\Theta}^{(i)}_r$ to local workers
        \For{each local worker $W^i$ \textbf{in parallel}}
            \State $\tilde{\Theta}^{(i)}_{r+1}\gets$ \textbf{LocalUpdate}$(\tilde{\Theta}^{(i)}_r, iter)$
            \State Send updated local parameters $\tilde{\Theta}^{(i)}_{r+1}$ to server
        \EndFor
        \State Compute $T_i$, $\forall i\in[M]$ (using FedAlign-A or FedAlign-O)
        \State Aggregate aligned local SSMs to obtain $\tilde{\Theta}_{r+1}$ (using \eqref{genericT})
        \State Obtain $\tilde{\Theta}^{(i)}_{r+1}$, $\forall i\in[M]$ (using \eqref{genericInvT})
        \State Server sends $\tilde{\Theta}^{(i)}_{r+1}$ back to respective workers
    \EndFor
\end{algorithmic}
\end{algorithm}

\subsection{FedAlign-A: The Analytical Method}

In FedAlign-A, we represent the common parameter basin within the CCF for the global SSM. Hence, we convert the local parameter basin of each local SSM into the CCF. It is worth noting that CCF is an option, any equivalent representation could be used. Moreover, for SISO ($nu = ny = 1$) and MIMO ($nu, ny > 1$) systems, we calculate similarity transformations using different mathematical formulations. 

\subsubsection{FedAlign-A for SISO SYSID}

For SISO SYSID tasks, upon completion of local training, the center server computes $T_i$, aligning the local parameter basin of $W^i$ with the CCF representation using:
\begin{equation}
\label{CCFtransform}
    T_i=P^{(i)}\left[\begin{array}{ccccc}
a_1 & a_2 & \cdots & a_{nx-1} & 1 \\
a_2 & a_3 & \cdots & 1 & 0 \\
\vdots & \vdots & \ddots & \vdots & \vdots \\
a_{nx-1} & 1 & \cdots & 0 & 0 \\
1 & 0 & \cdots & 0 & 0
\end{array}\right]. 
\end{equation}
Here, the controllability matrix is denoted by $P^{(i)}=[\tilde{B}^{(i)} \ \tilde{A}^{(i)}\tilde{B}^{(i)} \ \dots \ \mbox{$(\tilde{A}^{(i)})$}^{nx-1} \tilde{B}^{(i)}]$ whereas $a_1, a_2, \dots, a_{nx-1}$ refer the coefficient of the $p(\lambda)$ for $W^i$.

\subsubsection{FedAlign-A for MIMO SYSID} 

Similar to the SISO case, $T_i$ is calculated by the center server after local training. Yet, calculating $T_i$ is more challenging and is not unique for MIMO systems \cite{bay1999fundamentals}. 

To define $T_i$, let us first express the input matrix of $W^i$ as:
\begin{equation}
     \tilde{B}^{(i)} = \begin{bmatrix} \tilde{b}^{(i)}_1 & \tilde{b}^{(i)}_2 & \cdots & \tilde{b}^{(i)}_{nu} \end{bmatrix},
\end{equation}
where $\tilde{b}^{(i)}_\ell$ is the column representing $\ell^{\text{th}}$ input ($\ell = 1,2,\dots,nu$). The controllability matrix is 
\begin{equation}
   P^{(i)} = \begin{bmatrix} P^{(i)}_1 & P^{(i)}_2 & \cdots & P^{(i)}_{nu} \end{bmatrix}
\end{equation}
where each block is defined as
\begin{equation}
    P^{(i)}_\ell = \begin{bmatrix} \tilde{b}^{(i)}_\ell & \tilde{A}^{(i)}\tilde{b}^{(i)}_\ell  & \cdots & \left(\tilde{A}^{(i)}\right)^{nx-1} \tilde{b}^{(i)}_\ell \end{bmatrix}.
\end{equation}
We select $\mu_\ell$ independent column from each $P^{(i)}_\ell$ and denote this selection as matrix
\begin{equation}
\begin{split}
    M^{(i)} = \Bigl[\,
    & \underbrace{\tilde{b}^{(i)}_1 \quad \tilde{A}^{(i)}\tilde{b}^{(i)}_1 \quad \cdots \quad \bigl(\tilde{A}^{(i)}\bigr)^{\mu_1-1}\tilde{b}^{(i)}_1}_{\mu_1\text{ columns}}, \\
    & \cdots, \\
    & \underbrace{\tilde{b}^{(i)}_{nu} \quad \tilde{A}^{(i)}\tilde{b}^{(i)}_{nu} \quad \cdots \quad \bigl(\tilde{A}^{(i)}\bigr)^{\mu_{nu}-1}\tilde{b}^{(i)}_{nu}}_{\mu_{nu}\text{ columns}}
    \Bigr]
\end{split}
\end{equation}
where $\sum_{\ell=1}^{nu} \mu_\ell = nx$. We define the inverse of $M^{(i)}$ with row vectors $m^{(i)}$ as follows:
\[
\left(M^{(i)}\right)^{-1} 
= 
\left[
\begin{array}{c}
    m^{(i)}_{1,1} \\
    m^{(i)}_{1,2} \\
    \vdots \\
    m^{(i)}_{1,\mu_1} \\
    \hline
    \vdots \\
    \hline
    m^{(i)}_{nu,1} \\
    m^{(i)}_{nu,2} \\
    \vdots \\
    m^{(i)}_{nu,\mu_{nu}}
\end{array}
\right].
\]
Then, we calculate $T_i$ by using the last row from each partition, the rows denoted $m_{\ell,\mu_{nu}}$ for $\ell = 1, 2, \dots, nu$ \cite{bay1999fundamentals}.
\begin{equation} 
T_i 
= 
\left[
\begin{array}{@{}c@{}}
m^{(i)}_{1,\mu_1} \\
m^{(i)}_{1,\mu_1}\tilde{A}^{(i)} \\
\vdots \\
m^{(i)}_{1,\mu_1}\bigl(\tilde{A}^{(i)}\bigr)^{\mu_1 - 1} \\
\hline
\vdots \\
\hline
m^{(i)}_{nu,\mu_{nu}} \\
m^{(i)}_{nu,\mu_{nu}}\tilde{A}^{(i)} \\
\vdots \\
m^{(i)}_{nu,\mu_{nu}}\bigl(\tilde{A}^{(i)}\bigr)^{\mu_{nu}-1}
\end{array}
\right]^{-1}
\label{eq:myT}
\end{equation}

In the MIMO case of FedAlign-A, CCF transformation introduces an additional challenge due to its non-uniqueness. The structural hyperparameter $\mu_\ell$ must be set before training, as different choices of $\mu_\ell$ lead to distinct representations of $T_i$. This variability impacts how local SSMs align, influencing the stability and accuracy of the global SSM. Certain settings of $\mu_\ell$ may result in $T_i$ with high condition numbers ($\kappa(T_i$)), which can affect numerical stability and lead to deviations in system dynamics after similarity transformation. Thus, the choice of $\mu_\ell$ is critical for achieving effective state alignment, essential for ensuring stable and accurate SYSID. 

\subsection{FedAlign-O: The Optimization-based Method}

To overcome the challenges of aligning all local workers through the transformation into CCF, we propose FedAlign-O, which treats the alignment problem as an optimization task, making it suitable for both SISO and MIMO SYSID tasks.

In FedAlign-O, instead of forcing strict representation (such as CCF in FedAlign-A), we randomly pick an index $j \in [M]$ and 
designate the parameter basin of $W^j$ as the common parameter basin for the global SSM. We align the remaining workers' state representations with the state representation of $W^j$. To achieve this, the center server solves the following least squares problem to estimate each $T_i$ via generated pseudo-states for each $W^i$ ($\boldsymbol{\tilde{x}}^{(i)}_{pseudo}$). 
\begin{equation}
\label{RandomTransform}
    T_{i} = \underset{T}{\text{argmin}} \sum ({\boldsymbol{\tilde{x}}_{pseudo}^{(i)}} - T {\boldsymbol{\tilde{x}}_{pseudo}^{(j)}}) ^2, \forall i \neq j.
\end{equation}
Note that we define $T_j = I_{nx}$ as the parameter basin of $W^j$ defines the common parameter basin.

FedAlign-O addresses potential structural issues with $T_i$ that may arise in FedAlign-A by not enforcing all local workers to be represented in CCF. It also eliminates the need for the $\mu_\ell$ setting in MIMO SYSID. However, it requires the availability of an extra dataset or the generation of pseudo-data at the central server.

\section{Comparative Performance Analysis} \label{exp}

We present extensive experimental results on the SYSID performance of FL-SYSID using FedAlign-A and FedAlign-O compared to FedAvg. Each FL framework utilized the same training configuration, comprising $M=20$ workers for $R=20$ communication rounds. PEM was employed for local SYSID with two hyperparameters: model order $nx$ and local iteration $iter$.All experiments were conducted in MATLAB\textsuperscript{\textregistered} and repeated with 20 different seeds for statistical analysis. 

For each experiment, the overall SYSID performance is assessed by averaging the Best Fit Rate (BFR) of each local worker \( W^i \), \( i \in [M] \). For an output, BFR is computed as:  
\begin{equation}
    \text{BFR}_p = \frac{1}{M} \sum_{i \in [M]} \text{BFR}^{(i)}_p.
\end{equation}  
Here, the Best Fit Rate \( \text{BFR}^{(i)}_p \) for local worker \( W^i \) is defined by  
\begin{equation}
    \text{BFR}_{p}^{(i)} 
    = 100 \left(
    1 
    - 
    \sqrt{\frac{\displaystyle \sum_{k=1}^{K} \bigl(y_{p,k}^{(i)} - \tilde{y}_{p,k}^{(i)}\bigr)^{2}}
          {\displaystyle \sum_{k=1}^{K} \bigl(y_{p,k}^{(i)} - \bar{y}_{p}^{(i)}\bigr)^{2}}}
    \right),
\end{equation}
where $\bar{y}_{p}^{(i)} = (1\backslash K)\sum_{k=1}^K y_{p,k}^{(i)}$ and $p = 1, 2, \dots, ny$. We also recorded the total number of unstable global SSMs (\#UM) to assess the impact of direct averaging on global SSM stability. Additionally, we tracked the total number of global SSMs that failed to learn (\#F2L), defined as those with a BFR below zero. Experiments resulting in unstable or failed-to-learn global SSMs were excluded from BFR calculations.

The investigation consists of a five-fold analysis:  

\begin{itemize}  
    \item \textbf{Section \ref{ablation}:} To assess the impact of local SYSID performance on the efficiency of FL-SYSID, we conducted experiments on a synthetic SISO dataset using different values of $iter$ and $nx$.  
    \item \textbf{Section \ref{synthMIMO}:} We analyzed how the choice of $\mu_\ell$ affects the numerical stability of $T_i$ generated by FedAlign-A across two synthetic MIMO datasets with distinct dynamics. By exploring various $\mu_\ell$ settings, we evaluated their influence on the efficiency of FedAlign-A.  
\item \textbf{Sections \ref{sisocomp} and \ref{mimocomp}:} We conducted experiments on real-world SISO and MIMO datasets to compare the SYSID performance of FedAlign and FedAvg on both training and test data. 
\item \textbf{Section \ref{stat}:} We performed Wilcoxon tests using to statistically compare the SYSID performance of FedAlign and FedAvg on the test sets of SISO and MIMO datasets.
\end{itemize}

\subsection{Analyzing Alignment Challenges resulting from local SYSID performance} \label{ablation}

We created a synthetic dataset from a third-order system with zero dynamics (\( nu = ny = 1 \)). We sampled the initial states, inputs, and input noises as $\boldsymbol{x}_1^{(i)} \sim \mathcal{N}(0, 0.1^2 I_{nx})$, $\boldsymbol{u}_{1:K}^{(i)} \sim \mathcal{N}(0, 0.1^2 I_{nu})$, $\boldsymbol{w}_{1:K}^{(i)} \sim \mathcal{N}(0, 0.003^2 I_{nx})$, respectively. We attached a distinct dataset $D^i=\{\boldsymbol{u}^{(i)},\boldsymbol{y}^{(i)}\}$ to each $W^i$.

To analyze local SYSID performances on FL-SYSID, we set \( nx = \{2,3\} \) and \( iter = \{1,20\} \) for the local SSMs in the analysis. It should be pointed that, despite the actual system being third order ($nx=3$), we set $nx=2$ to assess reduced order modeling performance. We generated $\boldsymbol{\tilde{x}}^{(i)}_{pseudo}$ in FedAlign-O with  $\boldsymbol{u}_{1:K}^{(i)} \sim \mathcal{N}(0, 1^2 I_{nu})$. 

\begin{table}[t]
\centering
\footnotesize
\renewcommand{\arraystretch}{1.4}
\setlength{\tabcolsep}{2.5pt} 
\caption{Analysis of Local SYSID Performance in FL-SYSID Across 20 Experiments}\label{tab:nx23}
\begin{tabular}{@{}llcc|cc|cc@{}}
\toprule
\multicolumn{2}{c}{} 
& \multicolumn{2}{c}{\textbf{FedAvg}}  
& \multicolumn{2}{c}{\textbf{FedAlign-A}}  
& \multicolumn{2}{c}{\textbf{FedAlign-O}}  
\\ \midrule
\multicolumn{2}{c}{}  
& \textit{iter=1} & \textit{iter=20}  
& \textit{iter=1} & \textit{iter=20}  
& \textit{iter=1} & \textit{iter=20}  
\\ \midrule

\multirow{3}{*}{$nx=2$} 
& \textbf{BFR\(_1\)}  
& $55.47 (\pm 28.17)$ & $74.05 (\pm 0.48)$  
& $74.05 (\pm 0.44)$ & $74.05 (\pm 0.44)$  
& $74.05 (\pm 0.44)$ & $74.04 (\pm 0.44)$  
\\ 
& \textbf{\#UM}  
& 1 & 1  
& 0 & 0  
& 0 & 0  
\\  
& \textbf{\#F2L}  
& 0 & 0  
& 0 & 0  
& 0 & 0  
\\ \midrule

\multirow{3}{*}{$nx=3$} 
& \textbf{BFR\(_1\)}  
& $57.91 (\pm 29.34)$ & $83.53 (\pm 0.32)$  
& $83.50 (\pm 0.33)$ & $83.51 (\pm 0.32)$  
& $83.50 (\pm 0.32)$ & $83.51 (\pm 0.32)$  
\\ 
& \textbf{\#UM}  
& 3 & 2  
& 0 & 0  
& 0 & 0  
\\  
& \textbf{\#F2L}  
& 0 & 0  
& 0 & 0  
& 0 & 0  
\\ \bottomrule
\end{tabular}
\end{table}

\begin{figure}[t]
    \centering
    \subfloat[$iter=1$]{%
        \includegraphics[scale=0.55]{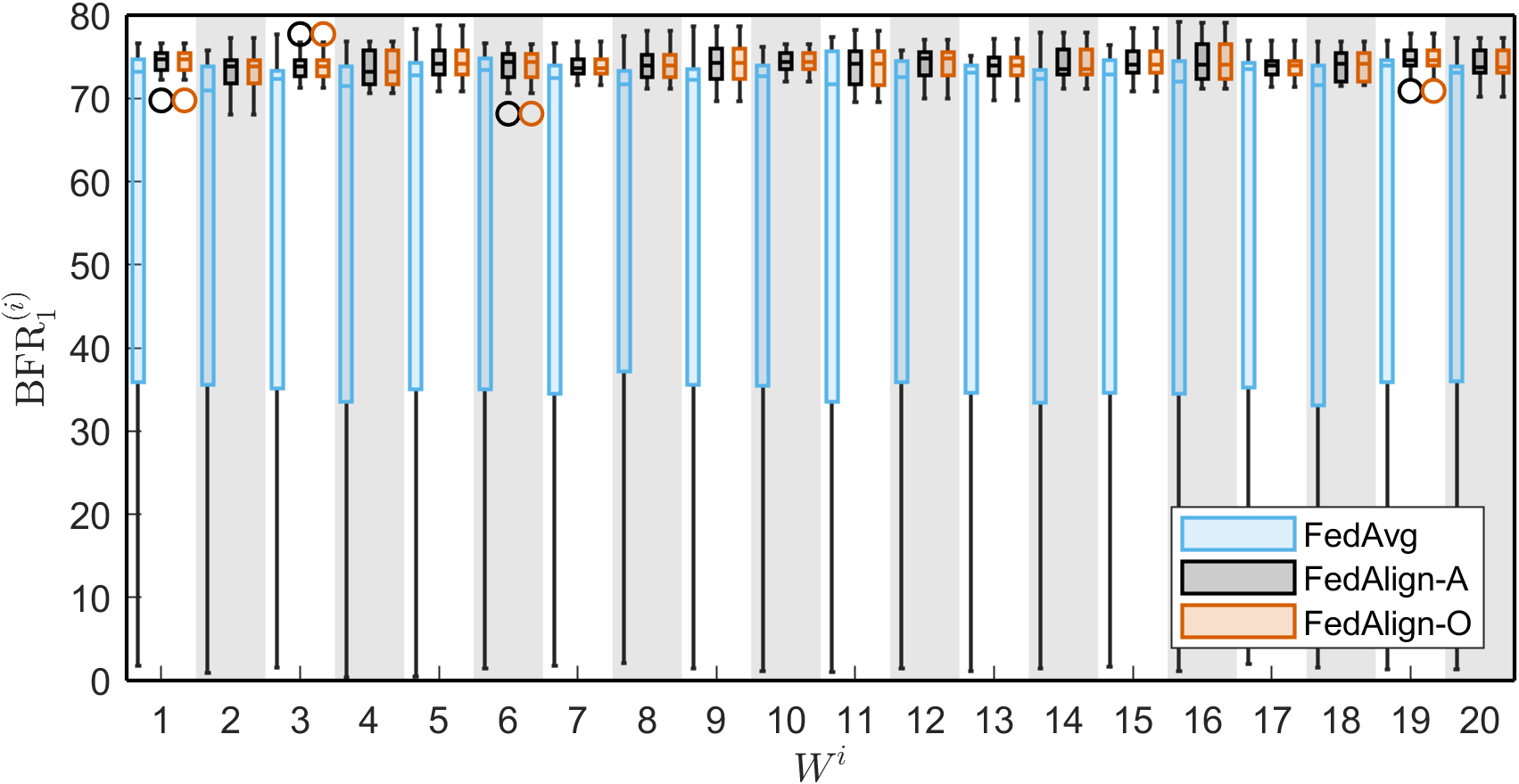}
    }
    \subfloat[$iter=20$]{%
        \includegraphics[scale=0.55]{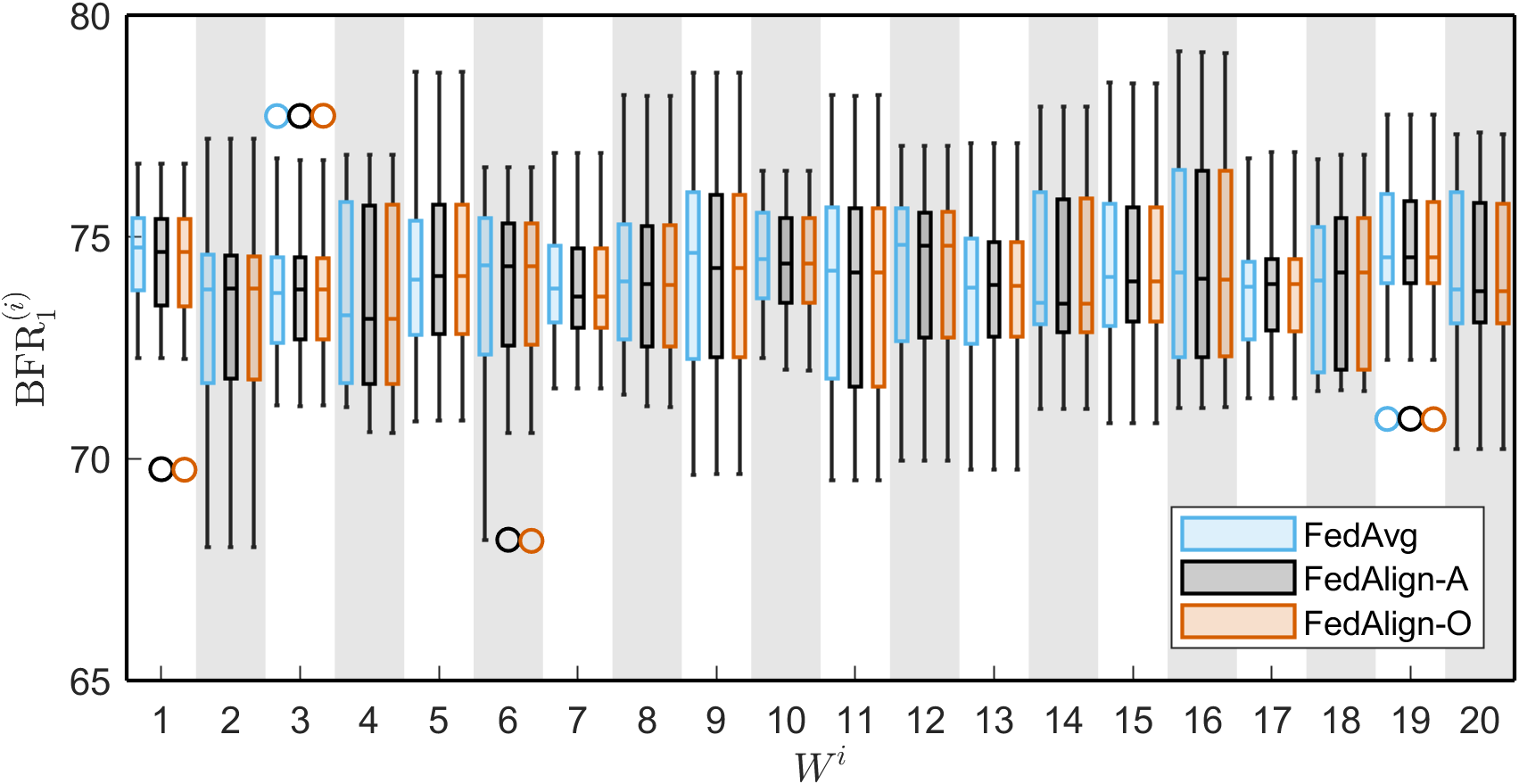}
    }
    \caption{Box plot comparison of FedAlign and FedAvg  for $nx=2$ on Synthetic Dataset}
    \label{fig:syntheticbox1}
\end{figure}

\begin{figure}[t]
    \centering
    \subfloat[$iter=1$]{%
        \includegraphics[scale=0.55]{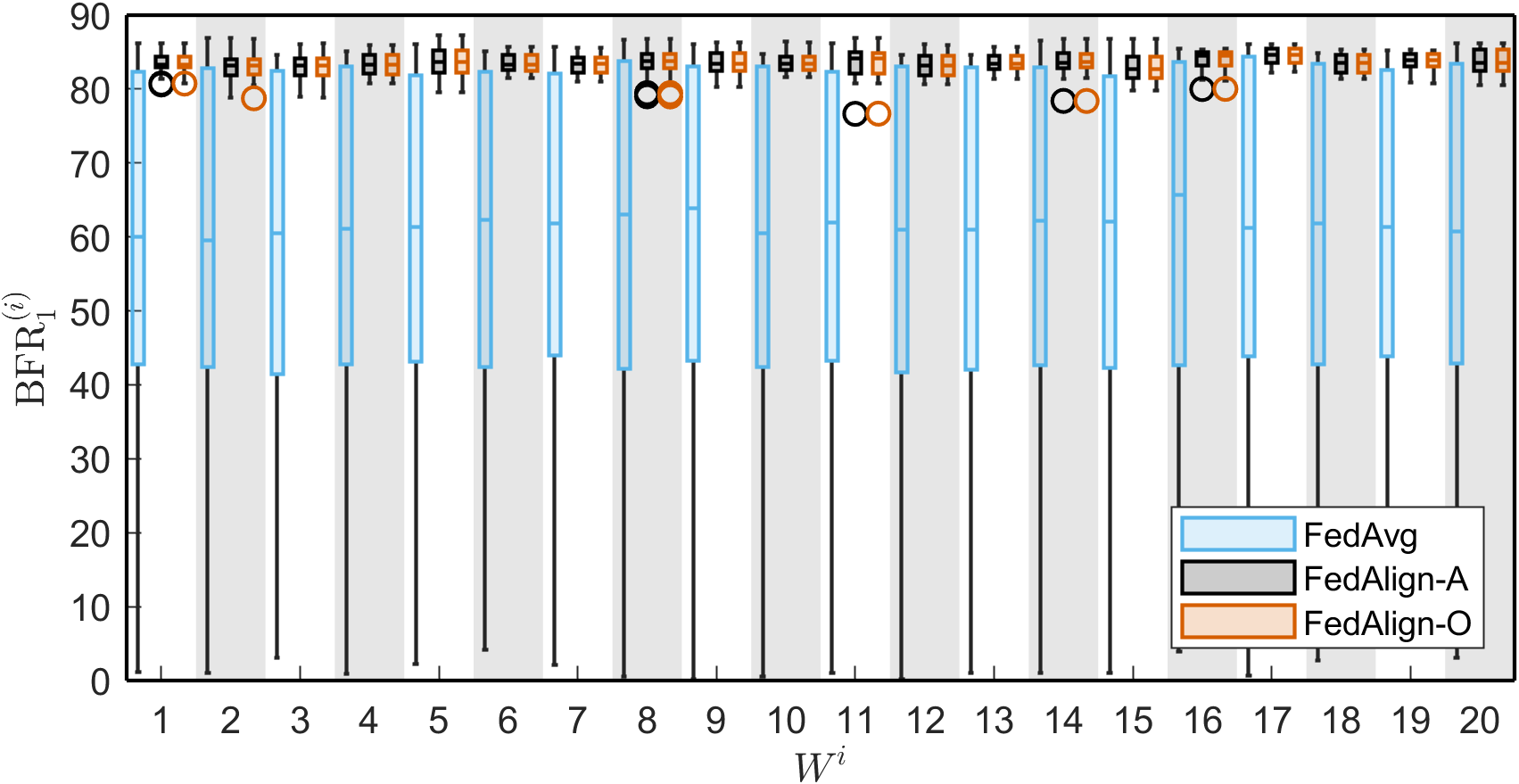}
    }
    \subfloat[$iter=20$]{%
        \includegraphics[scale=0.55]{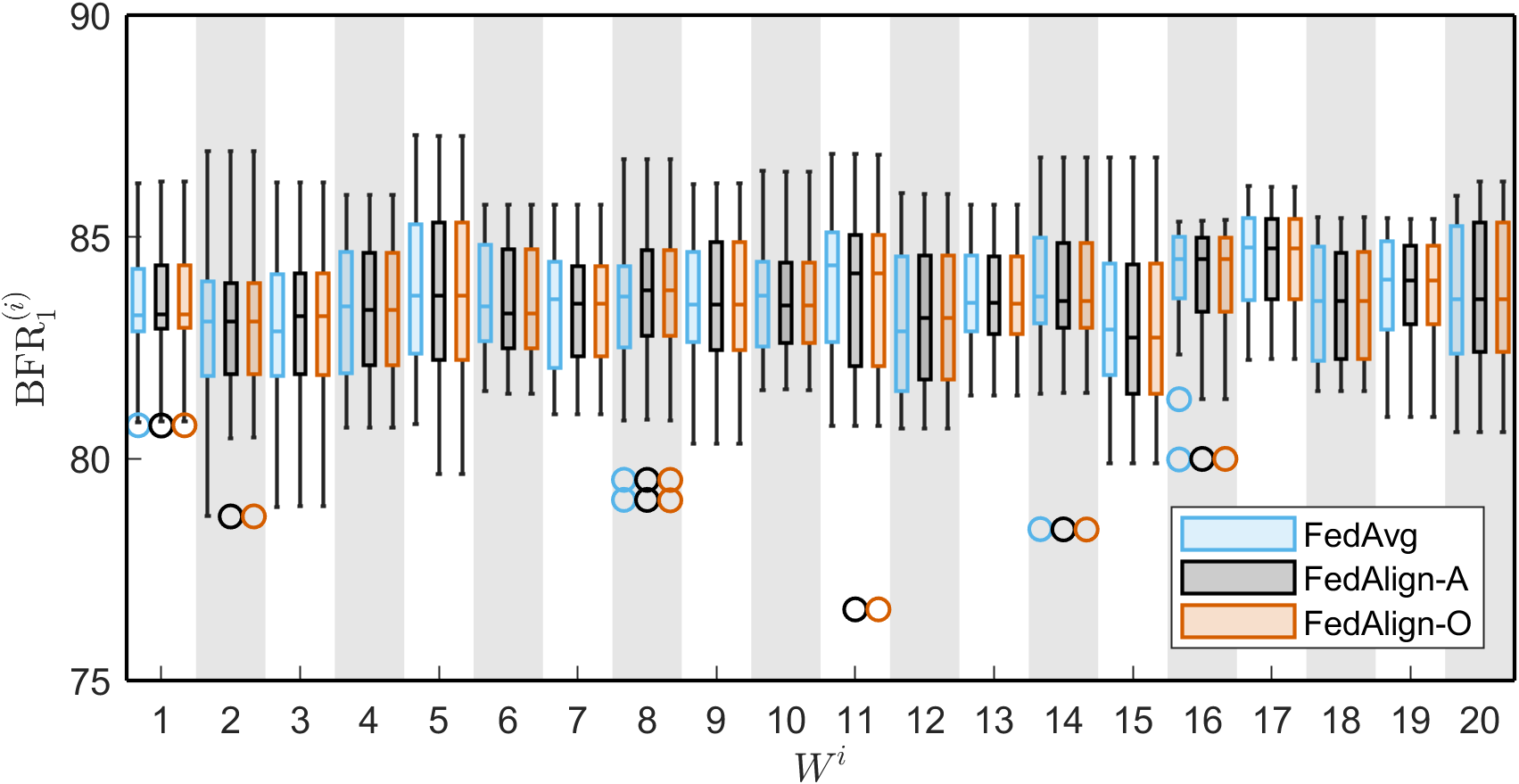}
    }
    \caption{Box plot comparison of FedAlign and FedAvg for $nx=3$ on Synthetic Dataset}
    \label{fig:syntheticbox2}
\end{figure}

\begin{figure}[t]
    \centering
    \subfloat[$nx=2$, $iter=1$]{\includegraphics[width=0.5\textwidth]{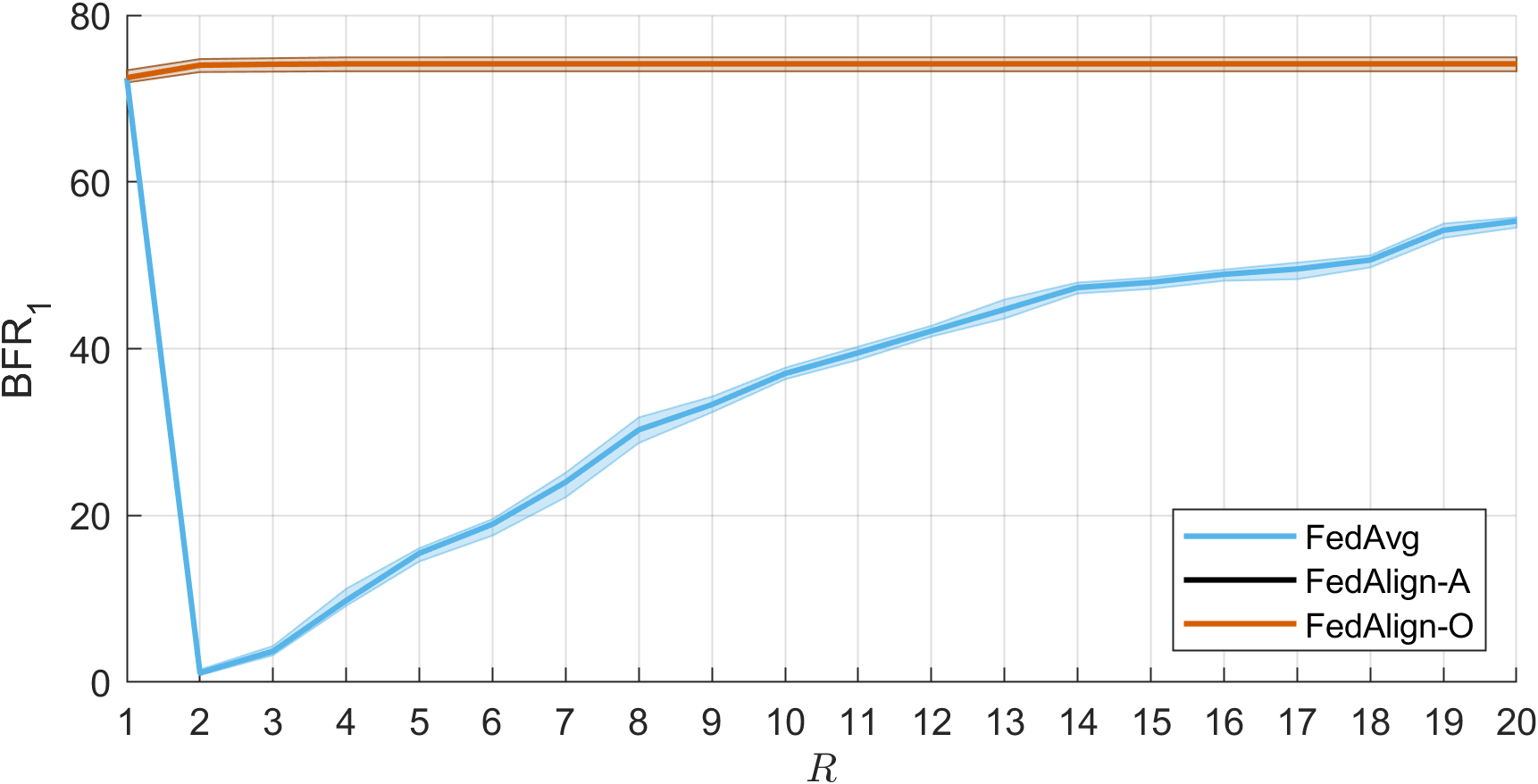}}
    \subfloat[$nx=2$, $iter=20$]{\includegraphics[width=0.5\textwidth]{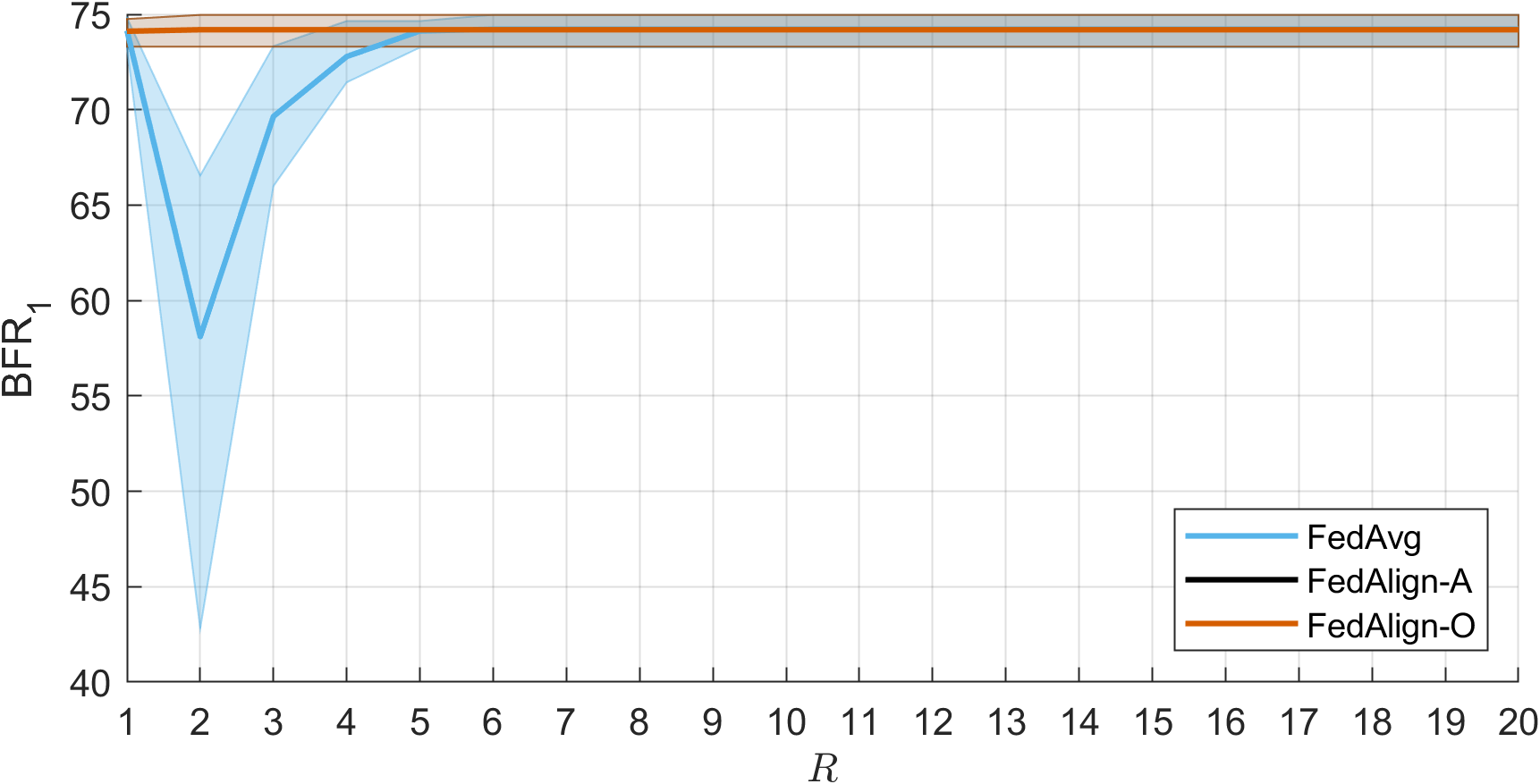}}
    \hfill
    \subfloat[$nx=3$, $iter=1$]{\includegraphics[width=0.5\textwidth]{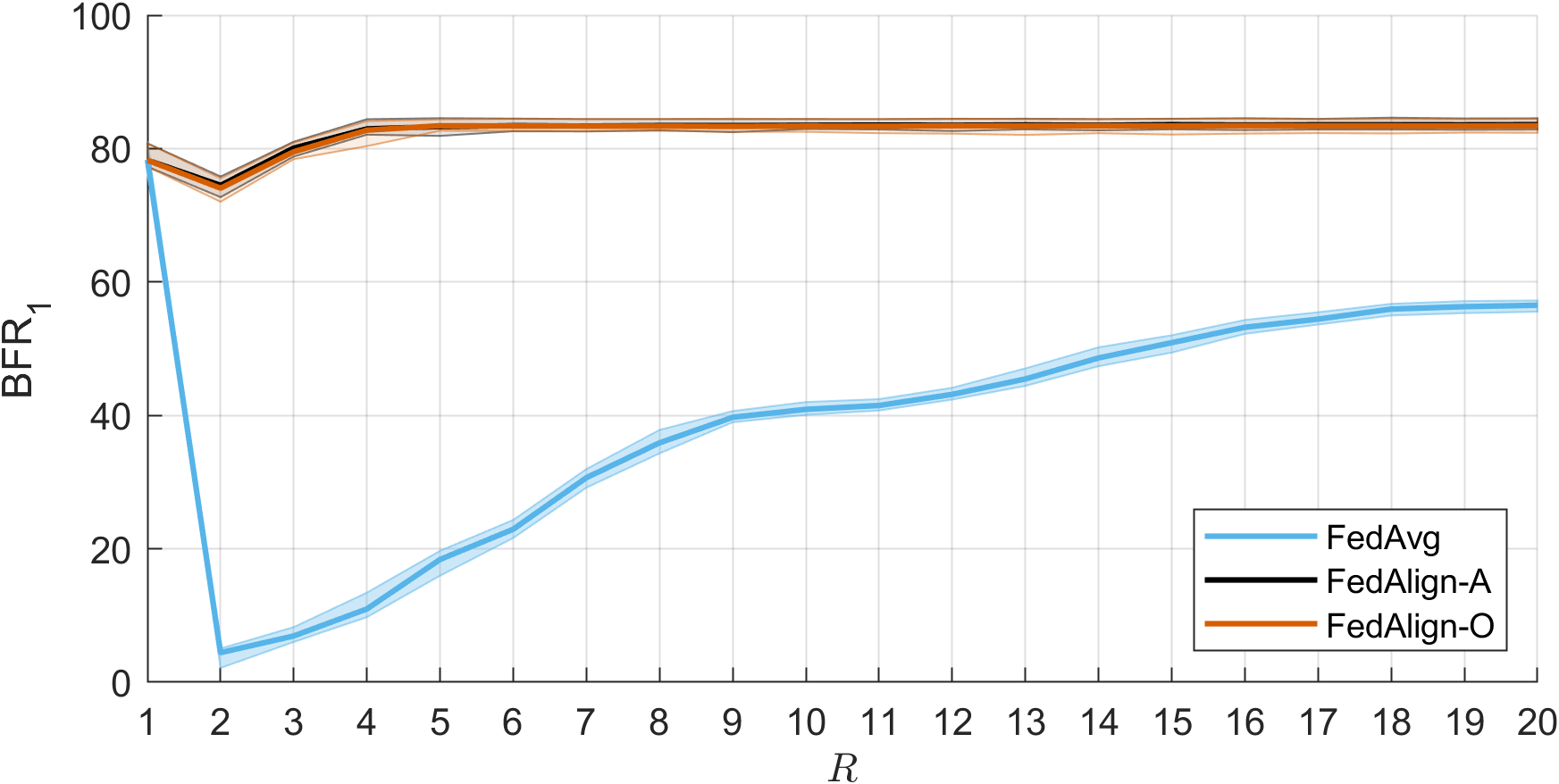}}
    \subfloat[$nx=3$, $iter=20$]{\includegraphics[width=0.5\textwidth]{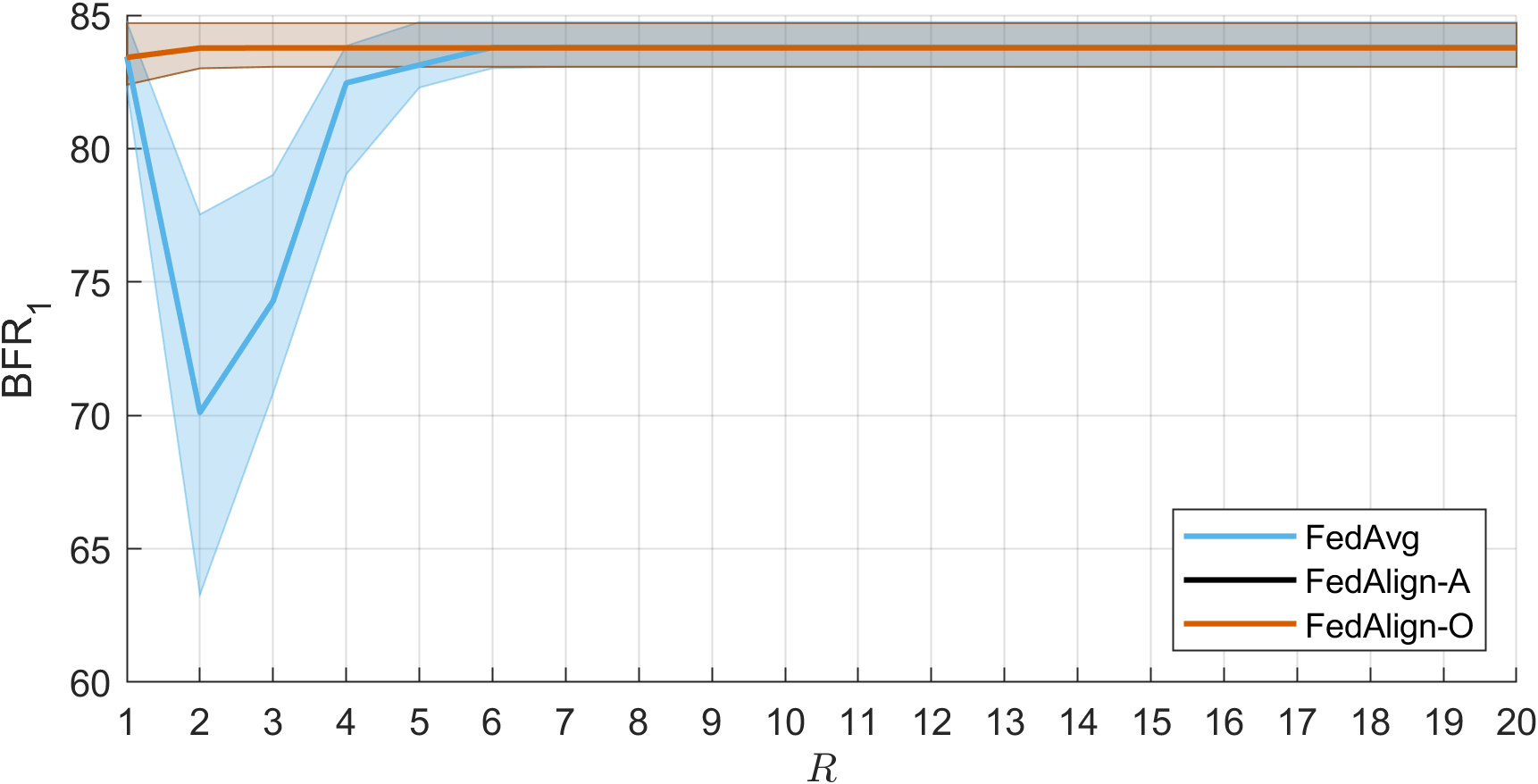}}
    \caption{Comparison of FedAlign and FedAvg Training on the Synthetic Dataset: Mean $\text{BFR}^{(i)}$ across local workers (solid lines) and the minimum and maximum $\text{BFR}^{(i)}$ across local workers (shaded areas).}
    \label{fig:syntheticcoverage}
\end{figure}

The mean BFR values with standard errors, \#UM, and \#F2L over 20 experiments are reported in Table \ref{tab:nx23}. The box plots of \( \text{BFR}^{(i)} \) are given in Fig. \ref{fig:syntheticbox1} and Fig. \ref{fig:syntheticbox2}, and the \( \text{BFR}^{(i)} \) during training are given in Fig. \ref{fig:syntheticcoverage}. Based on the results, we conclude that: 
\begin{itemize}
    \item FedAlign performs constantly better than FedAvg for the $nx=3$ and $iter=1$ settings as shown in Fig. \ref{fig:syntheticbox2}. FedAvg requires a higher number of local iterations ($iter=20$) to match the performance of FedAlign. 
    \item With the reduced-order learning model ($nx=2$), all methods experience a decrease in SYSID performance. As shown in Fig. \ref{fig:syntheticbox1}, FedAvg evaluates lower performance with higher variability. FedAvg could only achieve a performance similar to FedAlign only when the local iteration is increased to $iter=20$, similar to $nx=3$ case.  \item Fig. \ref{fig:syntheticcoverage} illustrates that FedAvg suffers from sudden performance drops in the earlier communication rounds. Due to the alignment of local parameter basins, FedAlign does not show such decreases during training, which results in faster convergence. 
    \item While FedAlign generates only stable global SSMs, thanks to local parameter basin alignment, FedAvg yields unstable global SSMs in all setups. Neither methods obtain F2L global SSM.
    \item Even though FedAlign-A does not demand pseudo-data generation to compute $T_i$, it achieves on-par SYSID performance with FedAlign-O. As shown in Fig. \ref{fig:syntheticcoverage}, it also matches the FedAlign-O's performance during training.
\end{itemize}

All in all, aligning the states of local SSMs provides several advantages to FedAlign, even with a single local iteration or reduced order modeling, including higher SYSID performance, quicker convergence, and improved global SSM stability. 

\subsection{Analyzing Alignment Challenges in FedAlign-A for MIMO SYSID}
\label{synthMIMO}

We generated two synthetic datasets, MIMO Synthetic dataset 1 (MIMO-1) and MIMO Synthetic dataset 2 (MIMO-2), to investigate how different $T_i$ constructed with various $\mu_\ell$ choices impact the SYSID performance of FedAlign-A. Both MIMO-1 and MIMO-2 are derived from fourth-order SSMs ($nx = 4) $ with two inputs and two outputs ($nu = ny = 2$) exhibiting different dynamics. We sampled the initial states as $\boldsymbol{x}_1^{(i)} \sim \mathcal{N}(0, 0.1^2 I_{nx})$, the inputs as $\boldsymbol{u}_{1:K}^{(i)} \sim \mathcal{N}(0, 0.1^2 I_{nu})$, and the output noises as $\boldsymbol{w}_{1:K}^{(i)} \sim \mathcal{N}(0, 0.01^2 I_{nx})$. We assigned a unique dataset $D^i=\{\boldsymbol{u}^{(i)},\boldsymbol{y}^{(i)}\}$ to each $W^i$.

In the analysis, we set $nx=4$ and $iter = \{1, 20\}$ for the local SSMs. We utilized $\boldsymbol{u}_{1:K}^{(i)} \sim \mathcal{N}(0, 0.1^2 I_{nu})$ to generate $\boldsymbol{\tilde{x}}^{(i)}_{pseudo}$ in FedAlign-O. In FedAlign-A, we constructed $M^{(i)}$ using three different $\mu_\ell$ settings:
\begin{itemize}
    \item FedAlign-A1: We used only the first input by setting 
    $\mu_1 = 4$ to construct $M^{(i)}$ as $M^{(i)} = \begin{bmatrix} \tilde{b}^{(i)}_1 & \tilde{A}^{(i)}\tilde{b}^{(i)}_1  & \left(\tilde{A}^{(i)}\right)^{2} \tilde{b}^{(i)}_1 & \left(\tilde{A}^{(i)}\right)^{3} \tilde{b}^{(i)}_1 \end{bmatrix}$
    \item FedAlign-A2: We used only the second input by setting $\mu_2 = 4$ to construct $M^{(i)}$ as $M^{(i)} = \begin{bmatrix} \tilde{b}^{(i)}_2 & \tilde{A}^{(i)}\tilde{b}^{(i)}_2  & \left(\tilde{A}^{(i)}\right)^{2} \tilde{b}^{(i)}_2 & \left(\tilde{A}^{(i)}\right)^{3} \tilde{b}^{(i)}_2 \end{bmatrix}$,
    \item FedAlign-A3: We used both first and second inputs by setting $\mu_1 = \mu_2 = 2$ to construct $M^{(i)}$ as $M^{(i)} = \begin{bmatrix} \tilde{b}^{(i)}_1 & \tilde{A}^{(i)}\tilde{b}^{(i)}_1  & \tilde{b}^{(i)}_2 & \tilde{A}^{(i)}\tilde{b}^{(i)}_2 \end{bmatrix}$.
\end{itemize}
For all settings, we examined the resulting condition number of $T_i$, $\kappa(T_i)$, throughout communication rounds to examine the impacts of numerical instabilities of $T_i$ on the efficiency of similarity transformation.

\begin{table}[t]
\centering
\footnotesize
\renewcommand{\arraystretch}{1.4}
\setlength{\tabcolsep}{3pt} 
\caption{Analysis of local SYSID performance in FL-SYSID with $iter=1$ across 20 experiments}\label{tab:iter1}
\begin{tabular}{@{}llccccc@{}}
\toprule
\multicolumn{2}{c}{} 
& \textbf{FedAvg} 
& \textbf{FedAlign-A1}
& \textbf{FedAlign-A2}
& \textbf{FedAlign-A3}
& \textbf{FedAlign-O}
\\ \midrule

\multirow{4}{*}{MIMO-1}
 & \textbf{BFR\(_1\)}
 & \(93.77 (\pm 14.56)\)
 & \(97.18 (\pm 0.05)\)
 & \(97.12 (\pm 0.08)\)
 & \(97.18 (\pm 0.05)\)
 & \(97.18 (\pm 0.05)\)
\\ 
 & \textbf{BFR\(_2\)}
 & \(93.81 (\pm 14.67)\)
 & \(97.17 (\pm 0.06)\)
 & \(97.14 (\pm 0.07)\)
 & \(97.17 (\pm 0.06)\)
 & \(97.17 (\pm 0.06)\)
\\ 
 & \textbf{\#UM}
 & 0
 & 0
 & 0
 & 0
 & 0
\\  
 & \textbf{\#F2L}
 & 0
 & 0
 & 0
 & 0
 & 0
\\ \midrule

\multirow{4}{*}{MIMO-2}
 & \textbf{BFR\(_1\)}
 & \(96.87 (\pm 0.09)\)
 & \(75.49 (\pm 18.71)\)
 & \(63.75 (\pm 41.67)\)
 & \(96.90 (\pm 0.03)\)
 & \(96.90 (\pm 0.03)\)
\\ 
 & \textbf{BFR\(_2\)}
 & \(96.80 (\pm 0.15)\)
 & \(82.06 (\pm 7.97)\)
 & \(71.19 (\pm 32.97)\)
 & \(96.85 (\pm 0.02)\)
 & \(96.85 (\pm 0.02)\)
\\ 
 & \textbf{\#UM}
 & 1
 & 8
 & 6
 & 0
 & 0
\\  
 & \textbf{\#F2L}
 & 1
 & 9
 & 11
 & 0
 & 0
\\ \bottomrule

\end{tabular}
\end{table}

\begin{table}[t]
\centering
\footnotesize
\renewcommand{\arraystretch}{1.4}
\setlength{\tabcolsep}{3pt} 
\caption{Analysis of local SYSID performance in FL-SYSID with $iter=20$ across 20 experiments}\label{tab:iter20}
\begin{tabular}{@{}llccccc@{}}
\toprule
\multicolumn{2}{c}{} 
& \textbf{FedAvg} 
& \textbf{FedAlign-A1}
& \textbf{FedAlign-A2}
& \textbf{FedAlign-A3}
& \textbf{FedAlign-O}
\\ \midrule

\multirow{4}{*}{MIMO-1}
 & \textbf{BFR\(_1\)}
 & \(97.18 (\pm 0.05)\)
 & \(97.18 (\pm 0.05)\)
 & \(97.13 (\pm 0.08)\)
 & \(97.18 (\pm 0.05)\)
 & \(97.18 (\pm 0.05)\)
\\ 
 & \textbf{BFR\(_2\)}
 & \(97.17 (\pm 0.06)\)
 & \(97.17 (\pm 0.06)\)
 & \(97.15 (\pm 0.07)\)
 & \(97.17 (\pm 0.06)\)
 & \(97.17 (\pm 0.06)\)
\\ 
 & \textbf{\#UM}
 & 0
 & 0
 & 0
 & 0
 & 0
\\  
 & \textbf{\#F2L}
 & 0
 & 0
 & 0
 & 0
 & 0
\\ \midrule

\multirow{4}{*}{MIMO-2}
 & \textbf{BFR\(_1\)}
 & \(96.90 (\pm 0.03)\)
 & \(-\)
 & \(90.85 (\pm 6.47)\)
 & \(96.90 (\pm 0.03)\)
 & \(96.90 (\pm 0.03)\)
\\ 
 & \textbf{BFR\(_2\)}
 & \(96.85 (\pm 0.02)\)
 & \(-\)
 & \(91.70 (\pm 6.99)\)
 & \(96.85 (\pm 0.02)\)
 & \(96.85 (\pm 0.02)\)
\\ 
 & \textbf{\#UM}
 & 0
 & 20
 & 9
 & 0
 & 0
\\  
 & \textbf{\#F2L}
 & 0
 & 0
 & 2
 & 0
 & 0
\\ \bottomrule

\end{tabular}
\end{table}

\begin{figure}[t]
    \centering
    \subfloat[MIMO-1 for $iter=1$ \label{sythcond11}]{\includegraphics[width=0.5\textwidth]{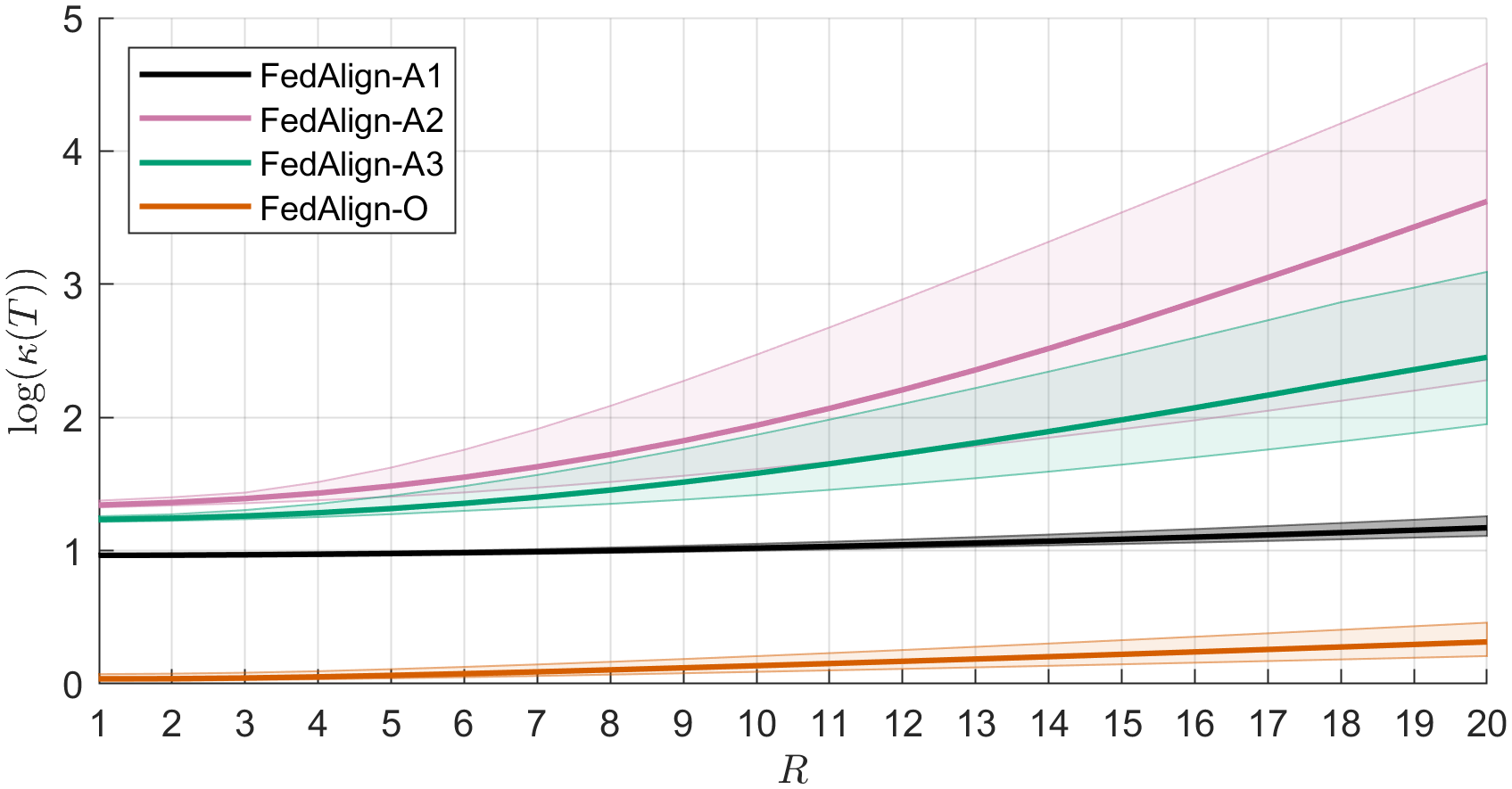}}
    \subfloat[MIMO-1 for $iter=20$ \label{sythcond120} ]{\includegraphics[width=0.5\textwidth]{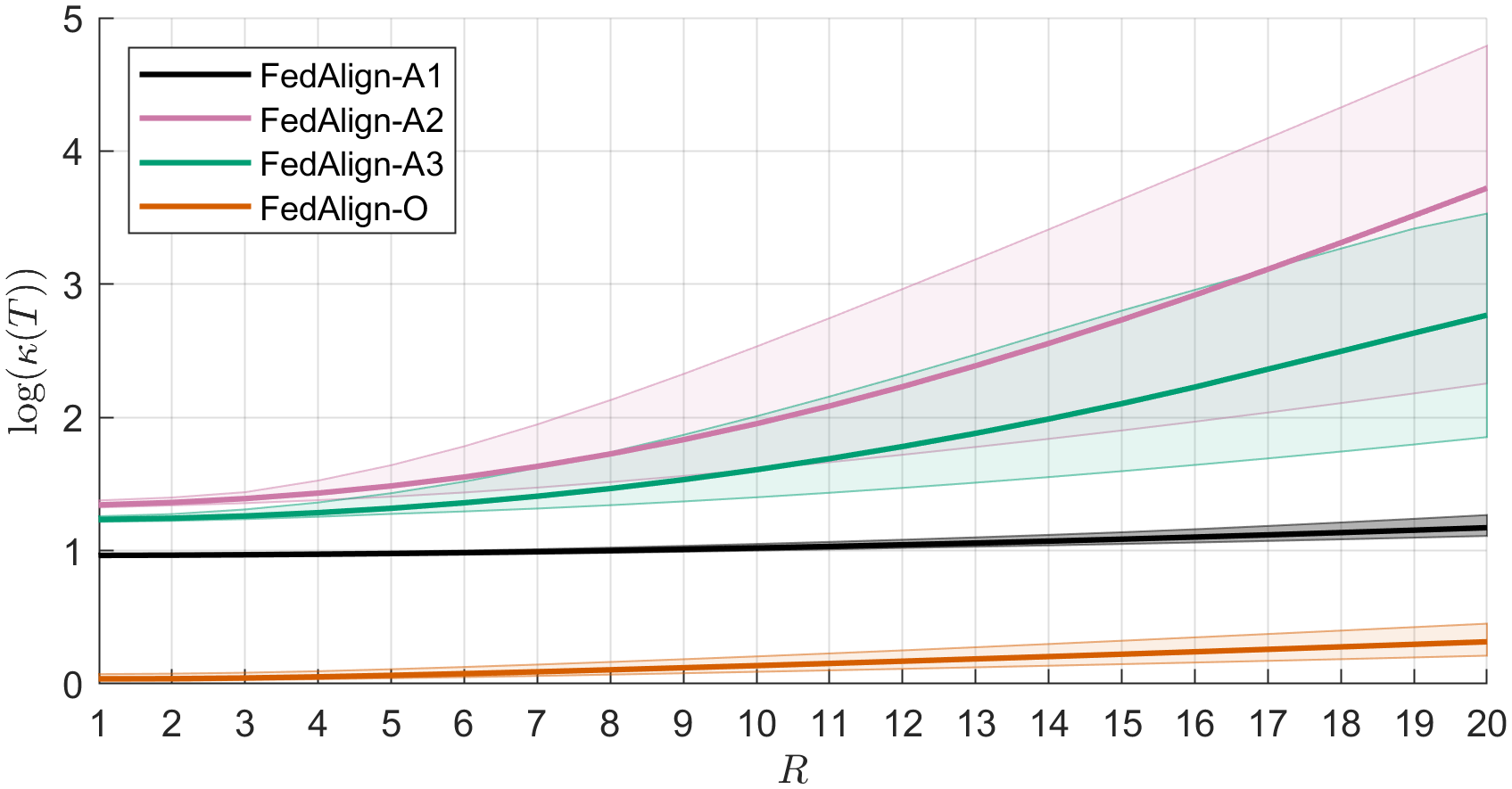}}
    \hfill
    \subfloat[MIMO-2 for $iter=1$ \label{sythcond21}]{\includegraphics[width=0.5\textwidth]{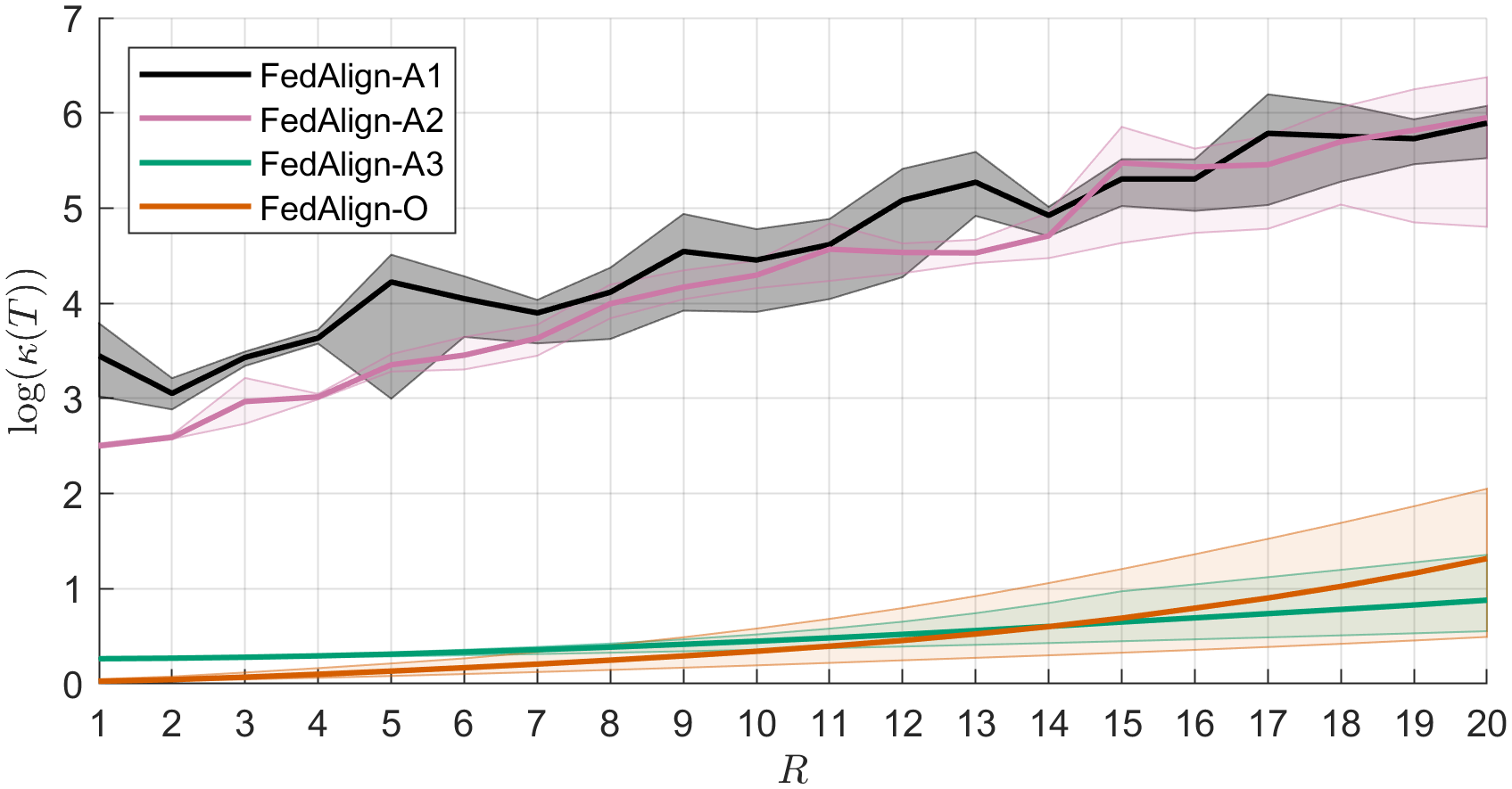}}
    \subfloat[MIMO-2 for $iter=20$ \label{sythcond220}]{\includegraphics[width=0.5\textwidth]{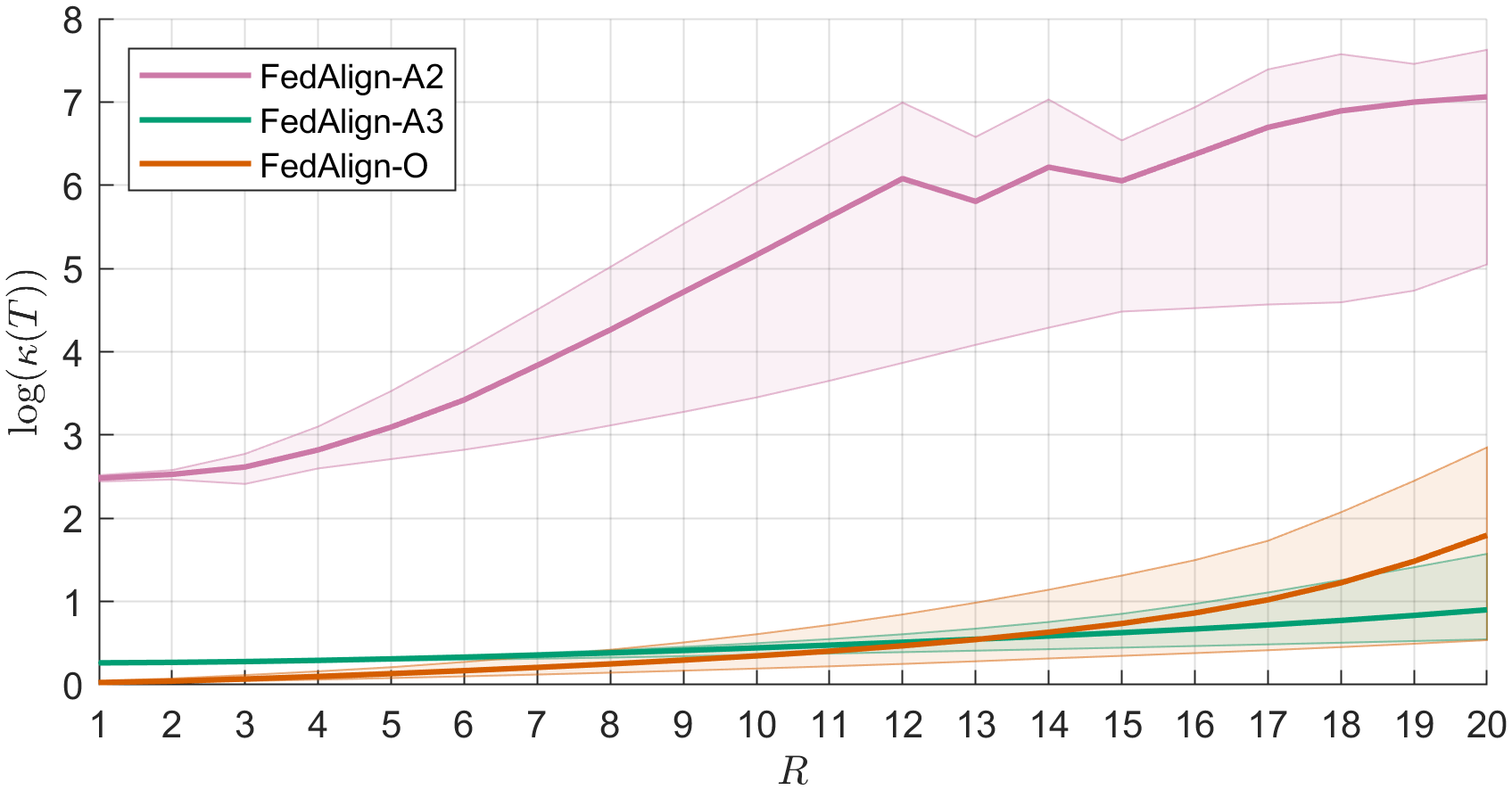}}
    \caption{Sensitivity comparison of FedAlign-A and FedAlign-O during training on the synthetic datasets: Mean $log(\kappa(T_i))$ across local workers (solid lines) and the minimum and maximum $log(\kappa(T_i))$ across local workers (shaded areas).}
    \label{fig:syntheticcond}
\end{figure}

\begin{figure}[t]
    \centering
    \subfloat{%
        \includegraphics[scale=0.52]{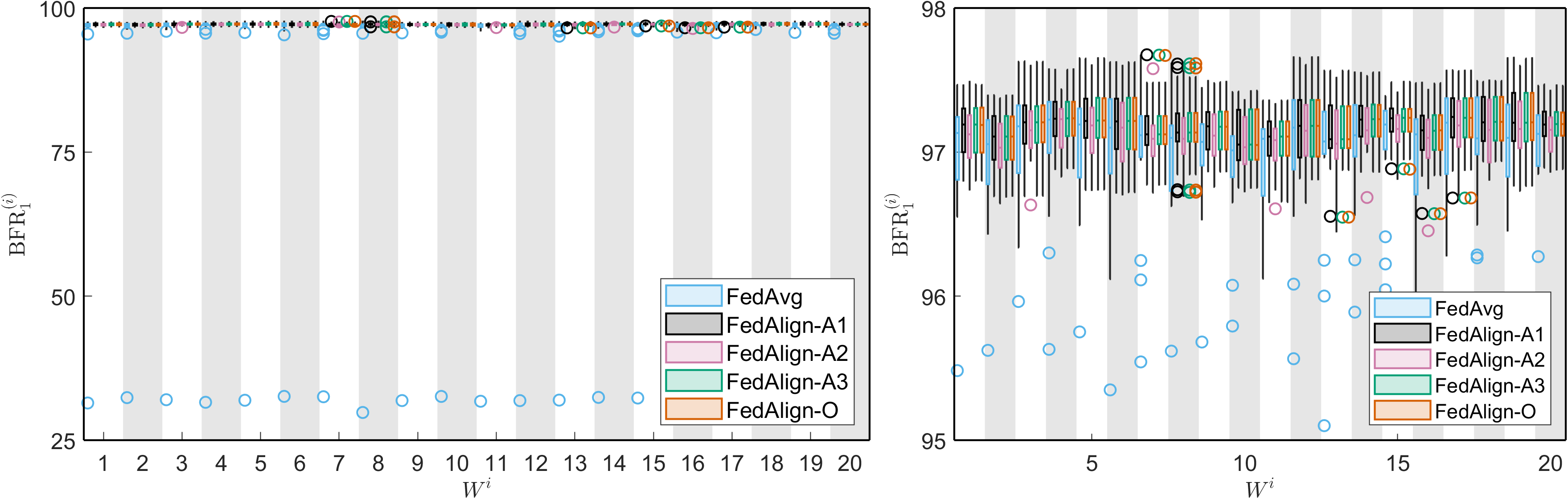}
    }\\
    \subfloat{%
        \includegraphics[scale=0.52]{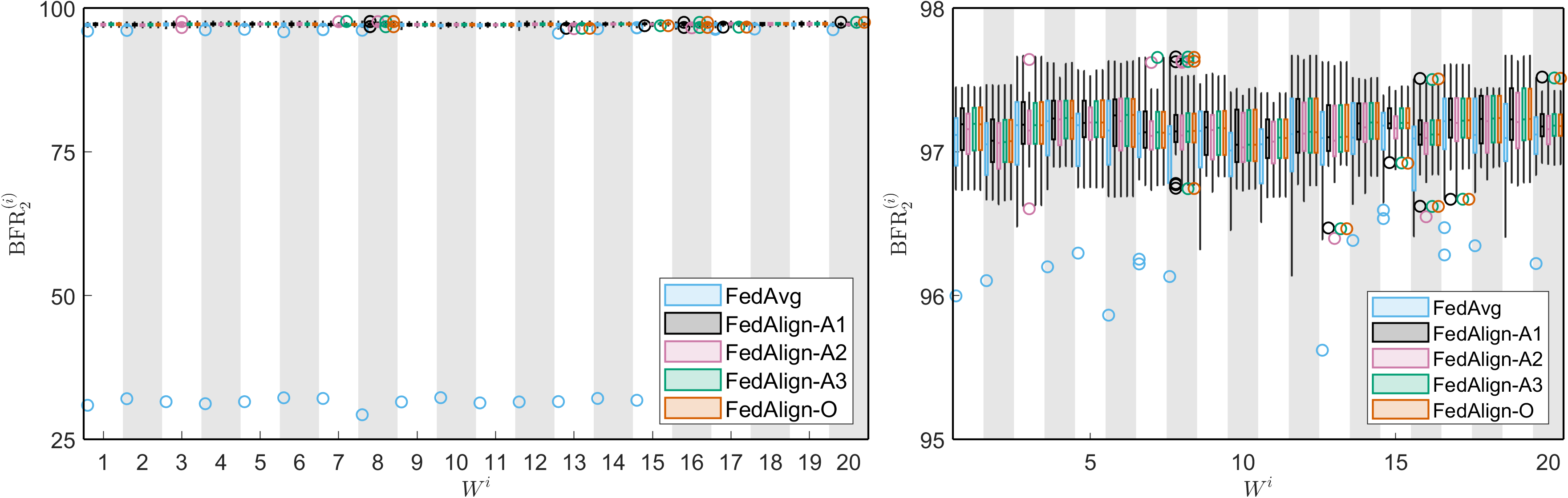}
    }
    \caption{Box plot comparison of FedAlign and FedAvg with $iter=1$ on MIMO-1 dataset for two outputs \(y_1\) (top row) and \(y_2\) (bottom row). Full-scale plot on the left, zoomed-in plot on the right.}
    \label{fig:syntheticMIMO1-1}
\end{figure}

\begin{figure}[t]
    \centering
    \subfloat{%
        \includegraphics[scale=0.55]{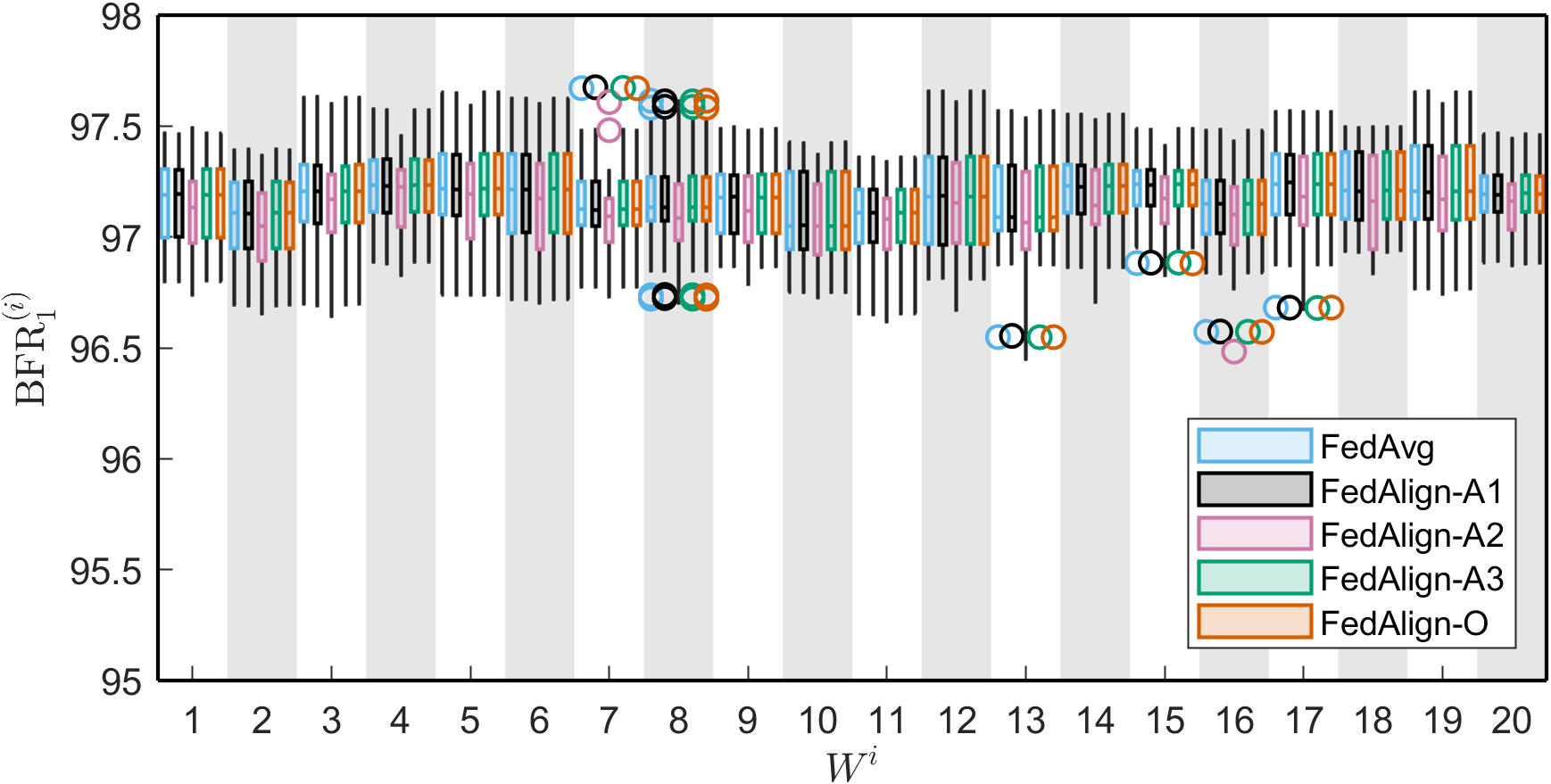}
    }
    \subfloat{%
        \includegraphics[scale=0.55]{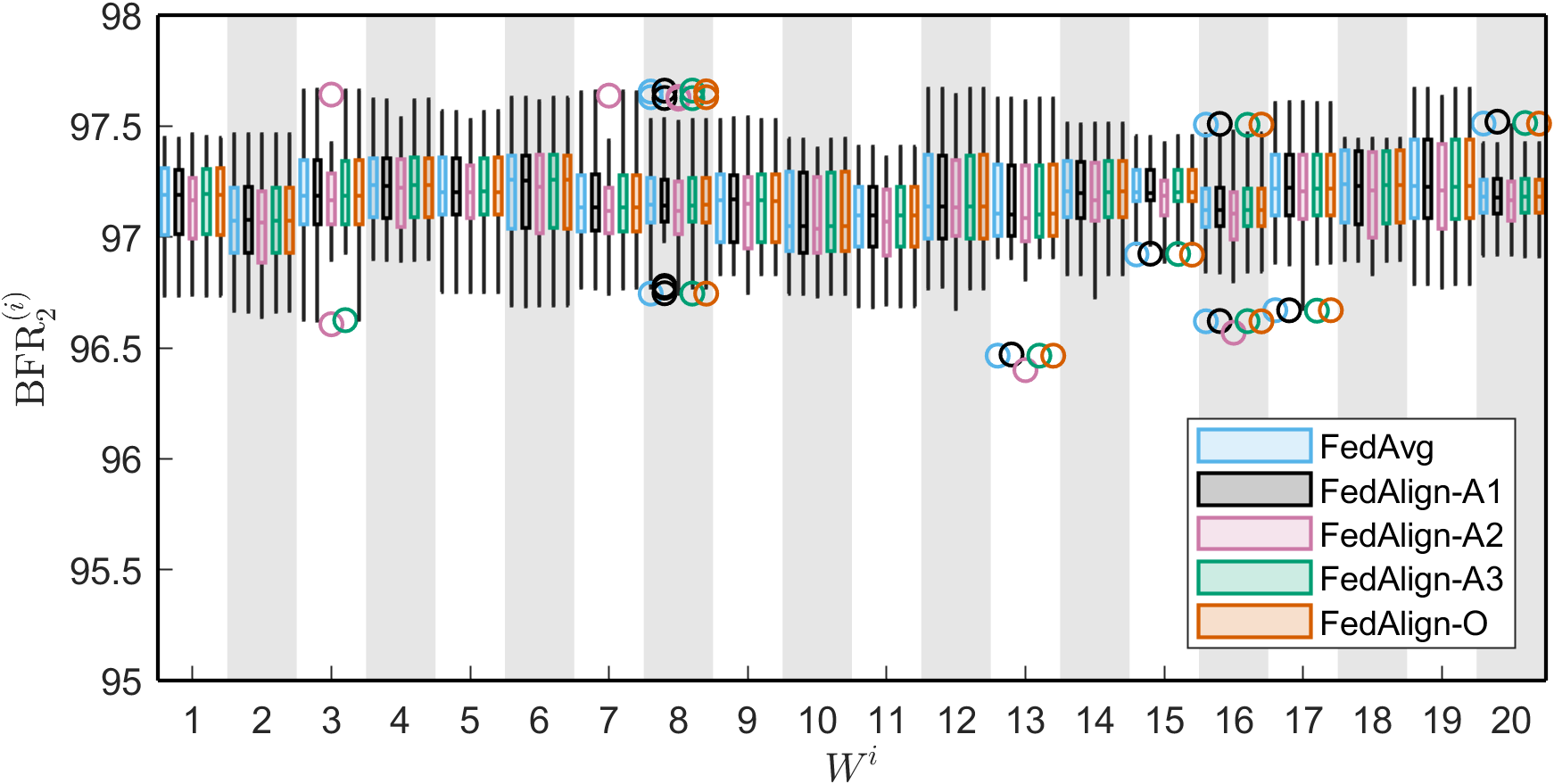}
    }
    \caption{Box plot comparison of FedAlign and FedAvg with $iter=20$ on MIMO-1 dataset for two outputs \(y_1\) (left) and \(y_2\) (right) }
    \label{fig:syntheticMIMO1-2}
\end{figure}

\begin{figure}[t]
    \centering
    \subfloat{%
        \includegraphics[scale=0.5]{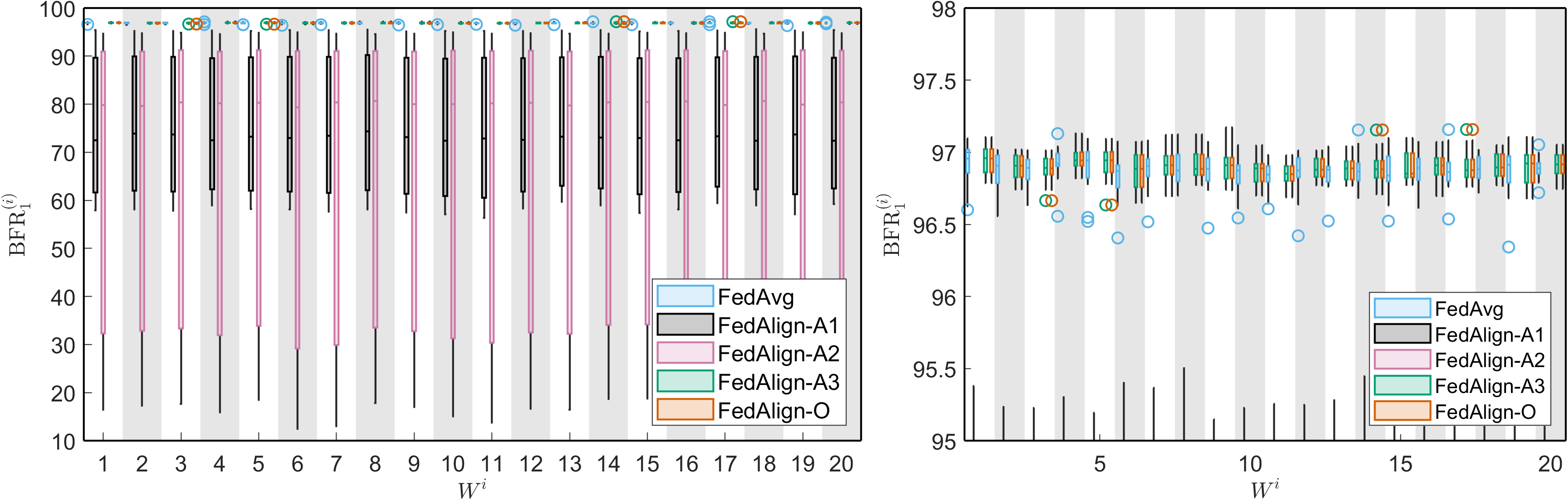}
    }\\
    \subfloat{%
        \includegraphics[scale=0.5]{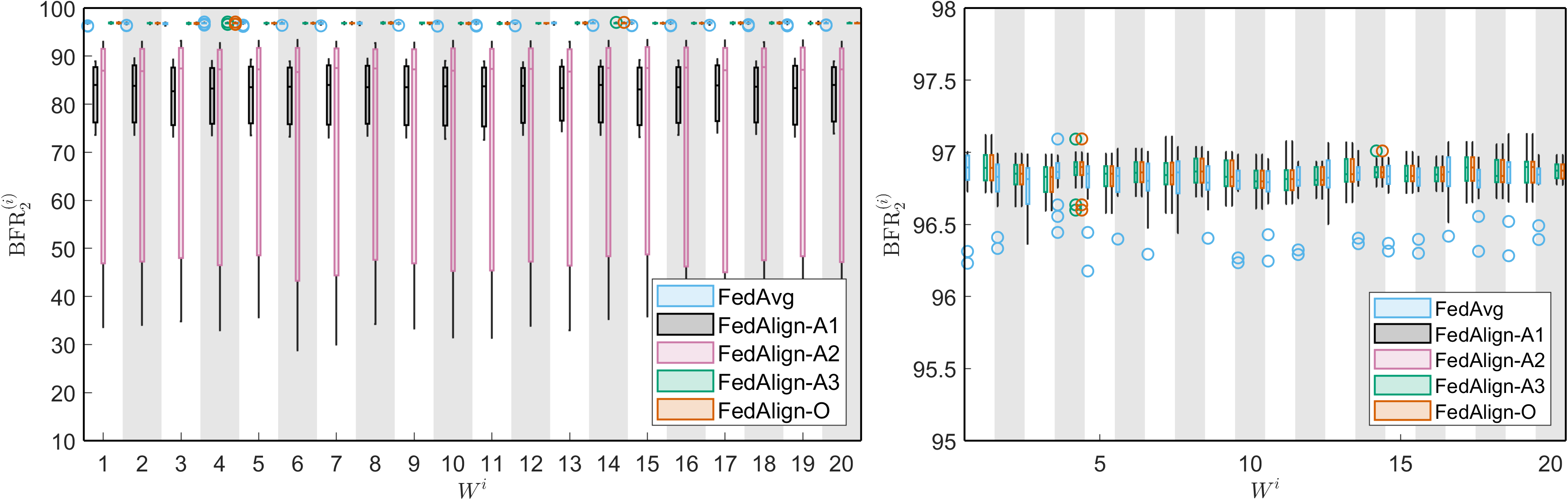}
    }
    \caption{Box plot comparison of FedAlign and FedAvg with $iter=1$ on MIMO-2 dataset for two outputs \(y_1\) (top row) and \(y_2\) (bottom row). Full-scale plot on the left, zoomed-in plot on the right.}
    \label{fig:syntheticMIMO2-1}
\end{figure}

\begin{figure}[t]
    \centering
    \subfloat{%
        \includegraphics[scale=0.5]{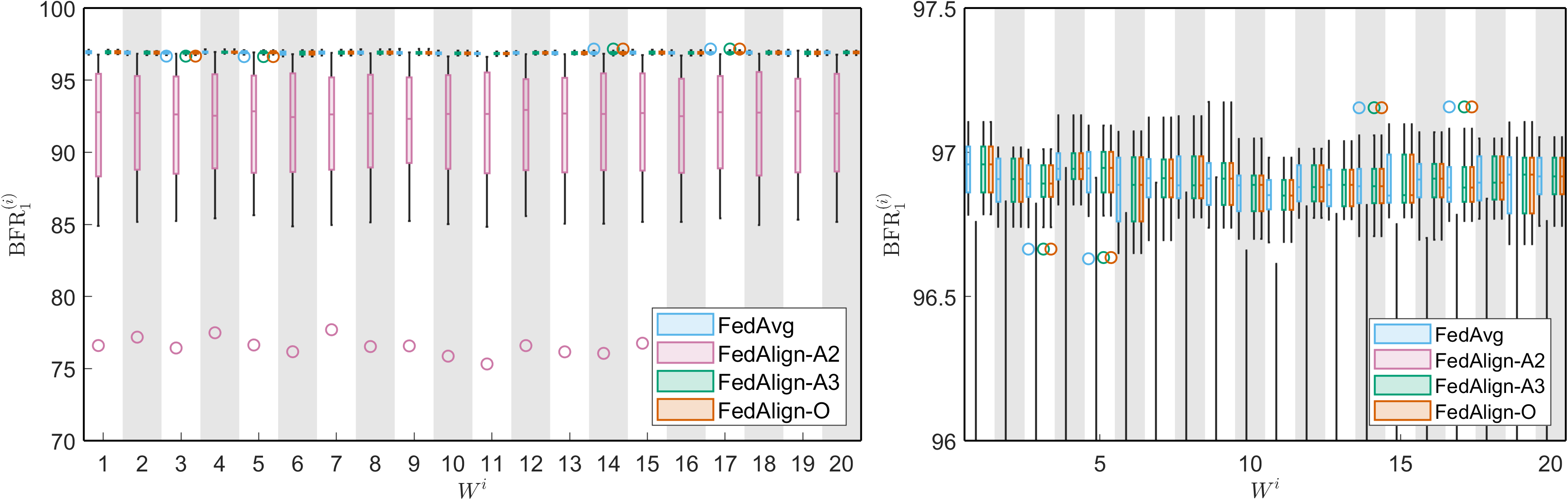}
    }\\
    \subfloat{%
        \includegraphics[scale=0.5]{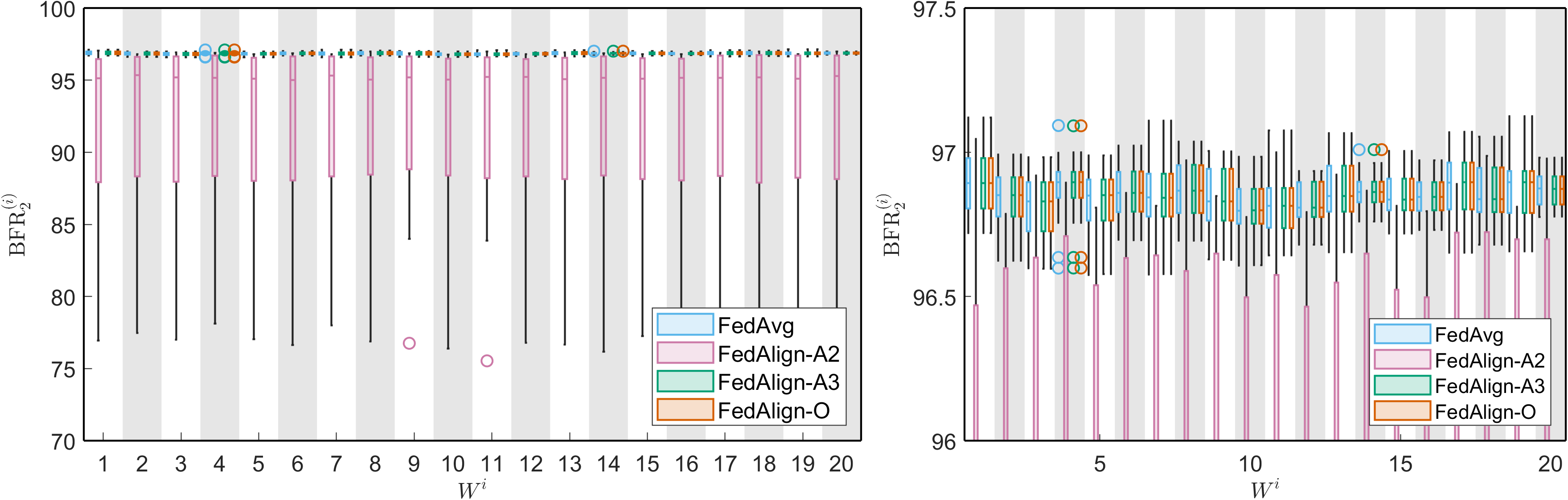}
    }
    \caption{Box plot comparison of FedAlign and FedAvg with $iter=20$ on MIMO-2 dataset for two outputs \(y_1\) (top row) and \(y_2\) (bottom row). Full-scale plot on the left, zoomed-in plot on the right.}
    \label{fig:syntheticMIMO2-2}
\end{figure}

Table \ref{tab:iter1} and Table \ref{tab:iter20} present the mean $\text{BFR}$ values with standard errors alongside \#UM and \#F2L for $iter=1$ and $iter=20$, respectively. Fig. \ref{fig:syntheticcond} illustrates how $log(\kappa(T_i))$ changes during training in FedAlign whereas Fig. \ref{fig:syntheticMIMO1-1} - Fig. \ref{fig:syntheticMIMO1-2} and Fig. \ref{fig:syntheticMIMO2-1} - Fig. \ref{fig:syntheticMIMO2-2} display the box plots of $\text{BFR}^{(i)}$ for MIMO-1 and MIMO-2, respectively. Based on these findings, we observe that:
\begin{itemize}
    \item On the MIMO-1 dataset for $iter= \{1,20\}$, all methods ensure efficient similarity transformation by computing small $\kappa(T_i)$ as seen from Fig. \ref{sythcond11} and Fig. \ref{sythcond120}, thus providing high SYSID performance.
    \item On the MIMO-2 dataset for $iter=1$, $\kappa(T_i)$ becomes higher in FedAlign-A1 and FedAlign-A2 while FedAlign-A3 and FedAlign-O obtain significantly lower $\kappa(T_i)$. As a result, inefficient similarity transformations in FedAlign-A1 and FedAlign-A2 lead to poor SYSID performance and the generation of unstable or F2L global SSMs. On the other hand, for $iter=20$, FedAlign-A1 generates only unstable global SSMs whereas $\kappa(T_i)$ remains higher in FedAlign-A2, as shown in Fig. \ref{sythcond220}.
    \item FedAvg evaluates fewer BFR values with higher deviation than FedAlign on the MIMO-1 dataset for $iter=1$, and requires higher local iterations ($iter=20$) to match FedAlign's performance, as shown in Fig. \ref{fig:syntheticMIMO1-1} and Fig. \ref{fig:syntheticMIMO1-2}. However, on the MIMO-2 dataset for $iter= \{1,20\}$, FedAvg gives higher performance than FedAlign-A1 and FedAlign-A2, and achieves performance on par with FedAlign-A3 and FedAlign-O, as depicted in Fig. \ref{fig:syntheticMIMO2-1} and Fig. \ref{fig:syntheticMIMO2-2}.
    \item  The performance of FedAlign-A depends strongly on the choice of $\mu_\ell$, which affects the stability of the similarity transform $T_i$. While this sensitivity is negligible on MIMO-1, it leads to instability on MIMO-2 with poorly chosen $\mu_\ell$. With proper selection (e.g., FedAlign-A3), FedAlign-A matches FedAlign-O, confirming the importance of this structural hyperparameter.

\end{itemize}

In summary, FedAlign-A may yield high $\kappa(T_i)$ depending on the choice of $\mu_\ell$, which considerably hinders the effectiveness of local parameter basin alignment, thus implying that $\mu_\ell$ must be set carefully. On the other hand, FedAlign-O obtains small $\kappa(T_i)$ as it does not obligate CCF representation for the global SSM.

\subsection{Performance comparison study on real-world SISO datasets}
\label{sisocomp}

We assessed the SYSID performance using the real-world SISO datasets. The datasets are pre-split into training and test subsets $D = \{D_{\text{train}}, D_{\text{test}}\}$ and listed as follows: 

\begin{itemize}
    \item MR Damper dataset \cite{wang2009identification} includes 3499 samples where velocity is the input and damper force is the output. The first 3000 samples comprise $D_{train}$ while $D_{test}$ contains the remaining samples.
    We added Gaussian noise, $\boldsymbol{v}_{1:K}^{(i)} \sim \mathcal{N}(0, 5^2)$ into $D_{train}$ for each $W^i$ to obtain distinct datasets. 
    \item The Hair Dryer dataset \cite{aguilar2020fractional} includes 1000 samples where heater voltage is the input and thermocouple voltage is the output.  The first 300 samples constitute \( D_{\text{train}} \) and \( D_{\text{test}} \) contains samples 801-900. To gather zero-mean data, we detrended the dataset. Moreover, we added \( \boldsymbol{v}_{1:K}^{(i)} \sim \mathcal{N}(0, 0.05^2) \) into \( D_{\text{train}} \) for each $W^i$.
    \item The Piezoelectric dataset \cite{piezoelectric} includes 10,000 samples where actuator voltage is the input and displacement is the output. The first 5,000 samples form $D_{train}$ while the last 5,000 samples establish $D_{test}$. We added $\boldsymbol{v}_{1:K}^{(i)} \sim \mathcal{N}(0, 5^2)$ into $D_{train}$ for each \( W^i \).
\end{itemize}

As depicted in Section \ref{ablation}, FedAlign achieves higher BFR values with fewer local iterations. Therefore, we set $iter = 1$ for FedAlign, while for FedAvg, we set $iter = \{1, 20\}$. For the MR Damper and Hair Dryer datasets, we set $nx=3$. For the Piezoelectric dataset, we set $nx=4$. We generated $\boldsymbol{\tilde{x}}^{(i)}_{pseudo}$ by using $\boldsymbol{u}_{1:K}^{(i)} \in  D_{\text{test}}^i$ in FedAlign-O.

\begin{figure}[t]
    \centering
    \subfloat[FedAlign \& FedAvg for $iter=1$]{%
        \includegraphics[scale=0.55]{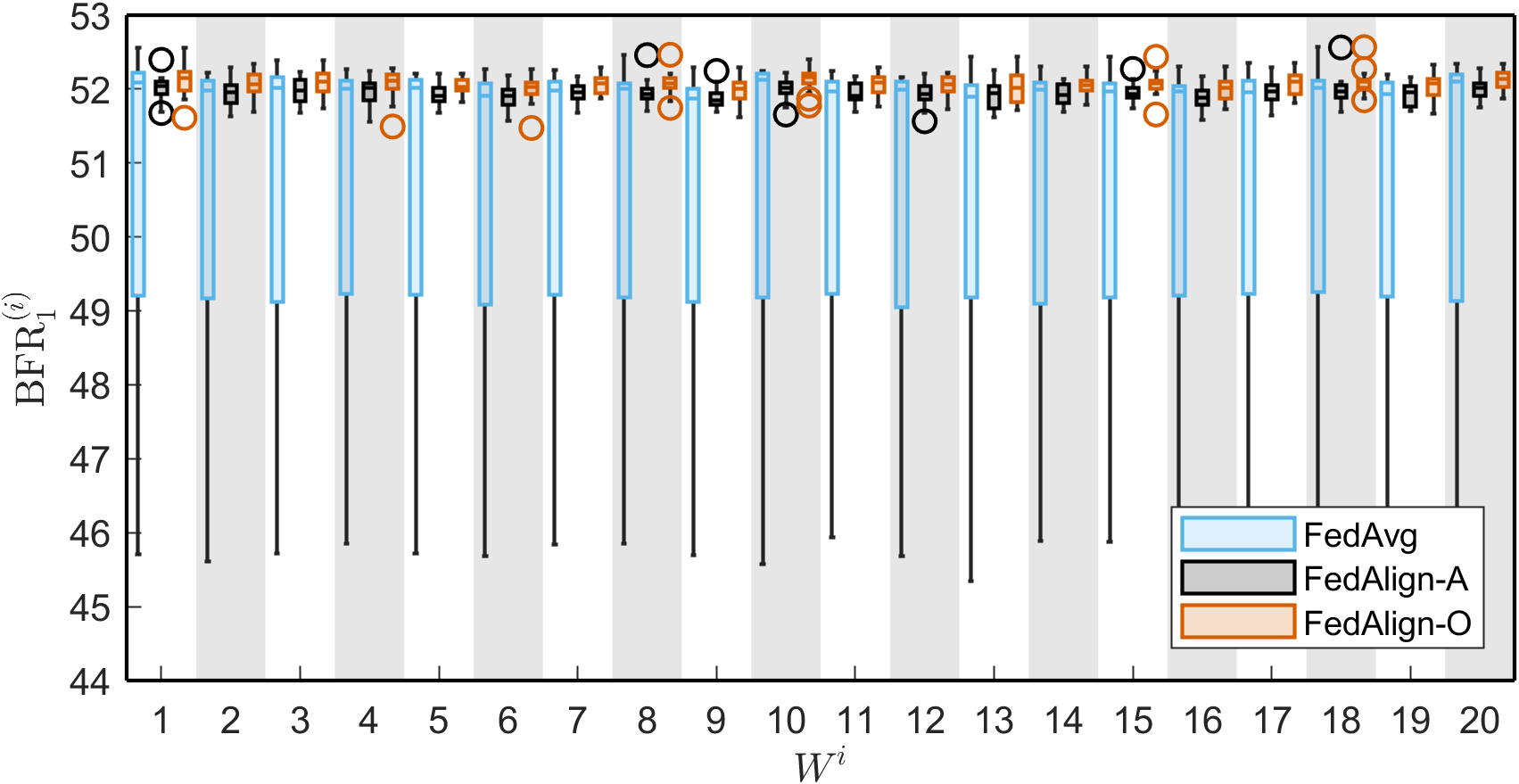}
    }
    \subfloat[FedAlign for $iter=1$, FedAvg for $iter=20$ \label{fig:mrdamperbox201}]{%
        \includegraphics[scale=0.55]{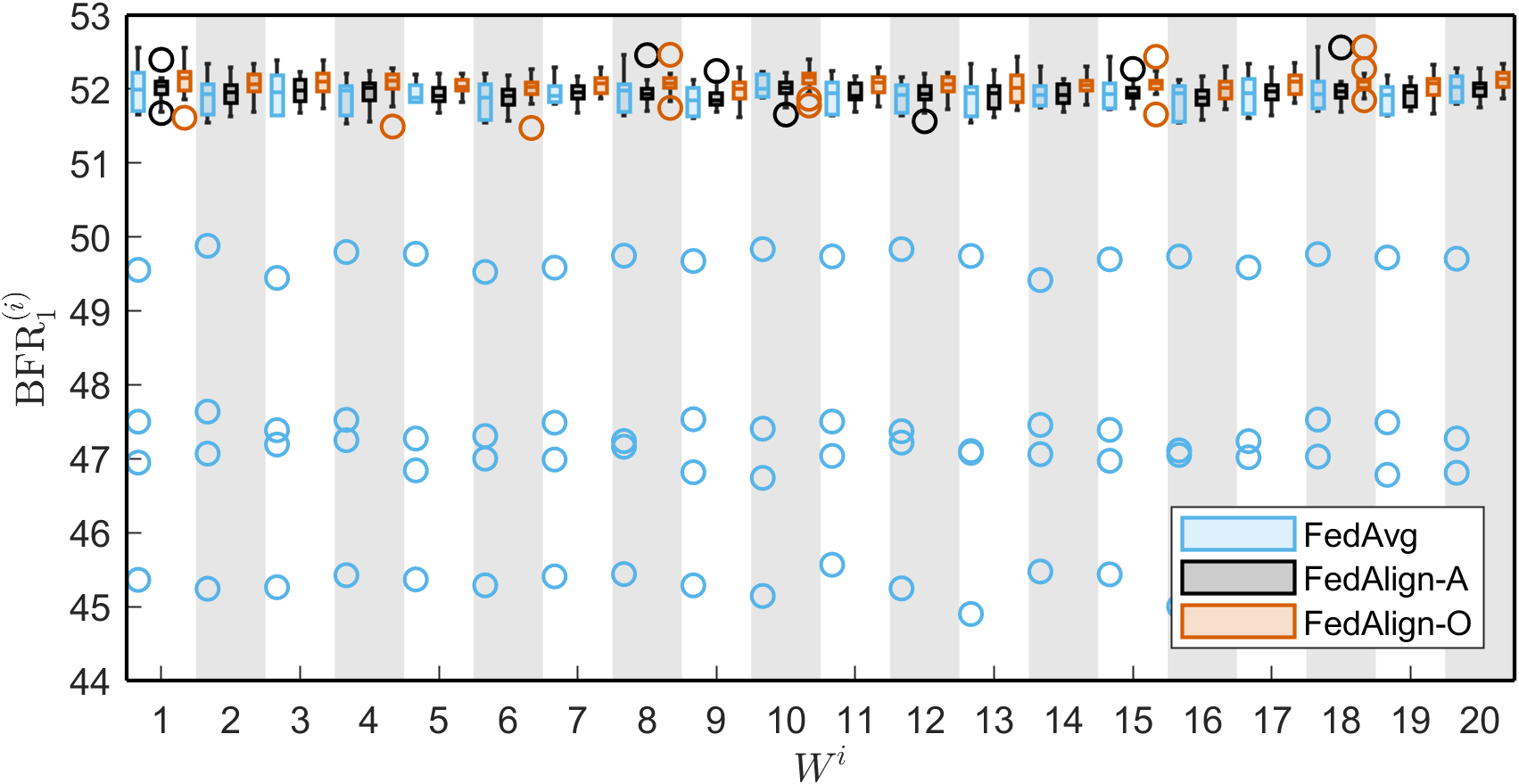}
    }
    \caption{Box plot comparison of FedAlign and FedAvg on MR Damper Dataset}
    \label{fig:mrdamperbox}
\end{figure}

\begin{figure}[t]
    \centering
    \subfloat[FedAlign \& FedAvg for $iter=1$]{%
        \includegraphics[scale=0.5]{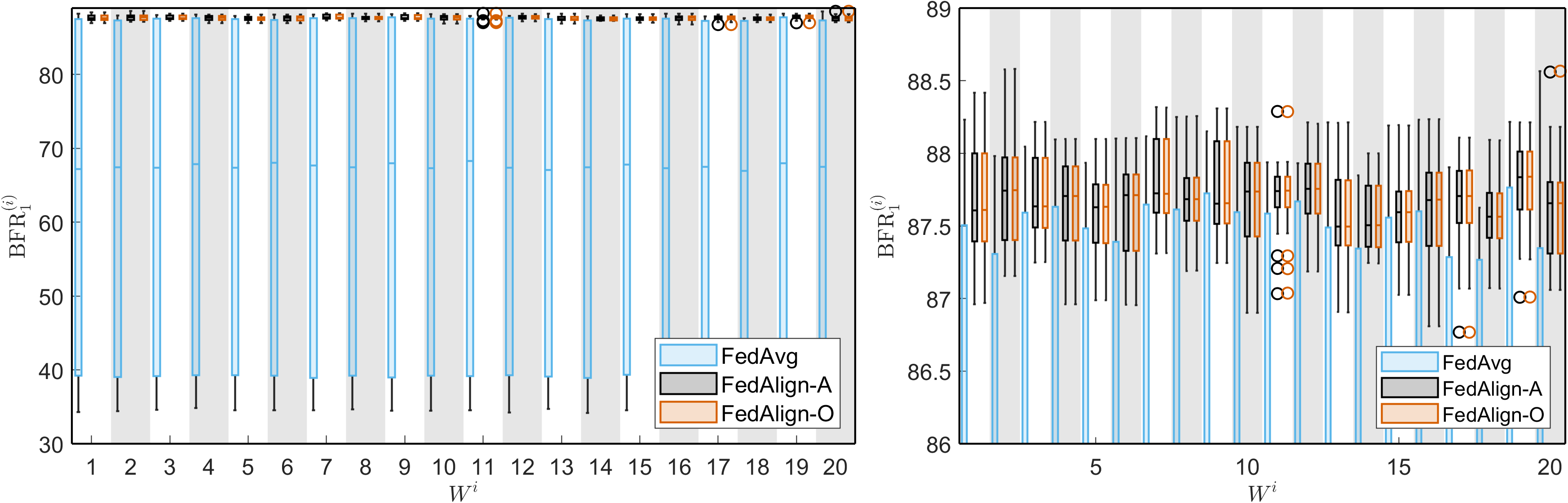}
    }\\
    \subfloat[FedAlign for $iter=1$, FedAvg for $iter=20$ \label{fig:hairdryerbox201}]{%
        \includegraphics[scale=0.5]{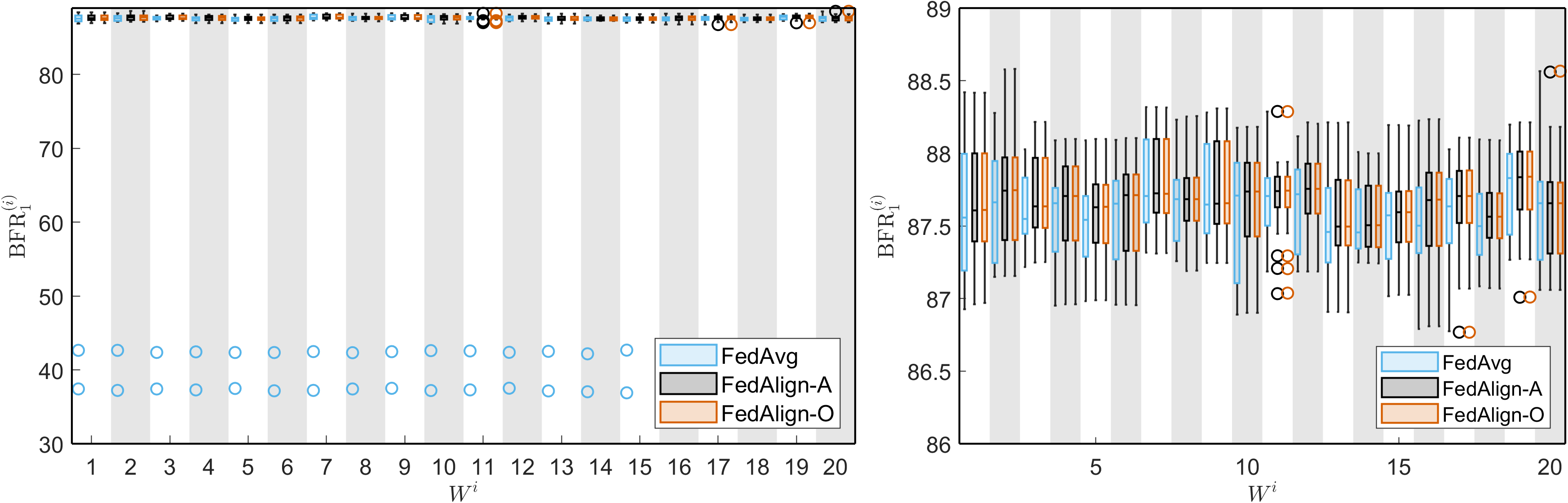}
    }
    \caption{Box plot comparison of FedAlign and FedAvg on Hair Dryer Dataset. Full-scale plot on the left, zoomed-in plot on the right.}
    \label{fig:hairdryerbox}
\end{figure}

\begin{figure}[t]
    \centering
    \subfloat[FedAlign \& FedAvg for $iter=1$]{%
        \includegraphics[scale=0.55]{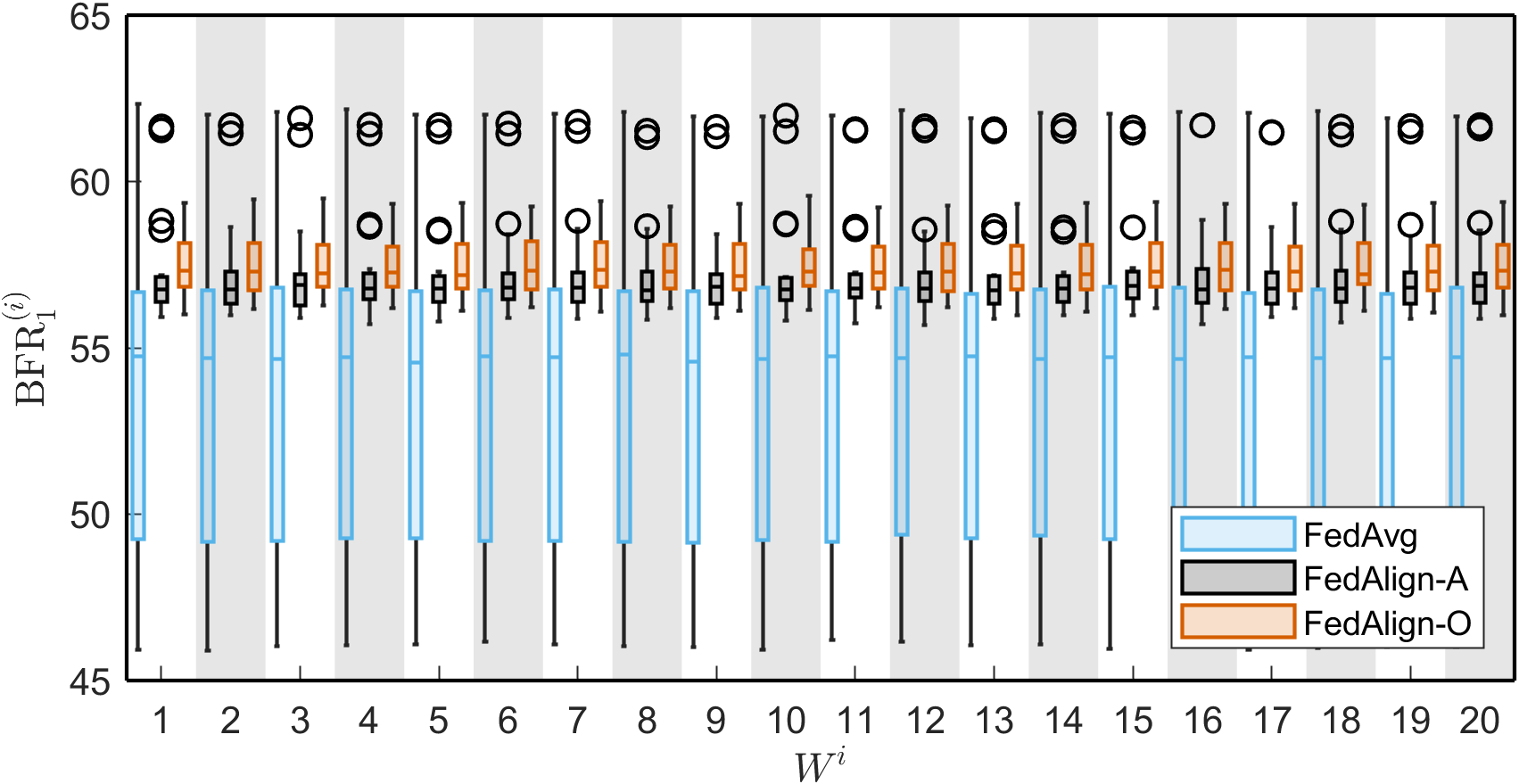}
    }
    \subfloat[FedAlign for $iter=1$, FedAvg for $iter=20$]{%
        \includegraphics[scale=0.55]{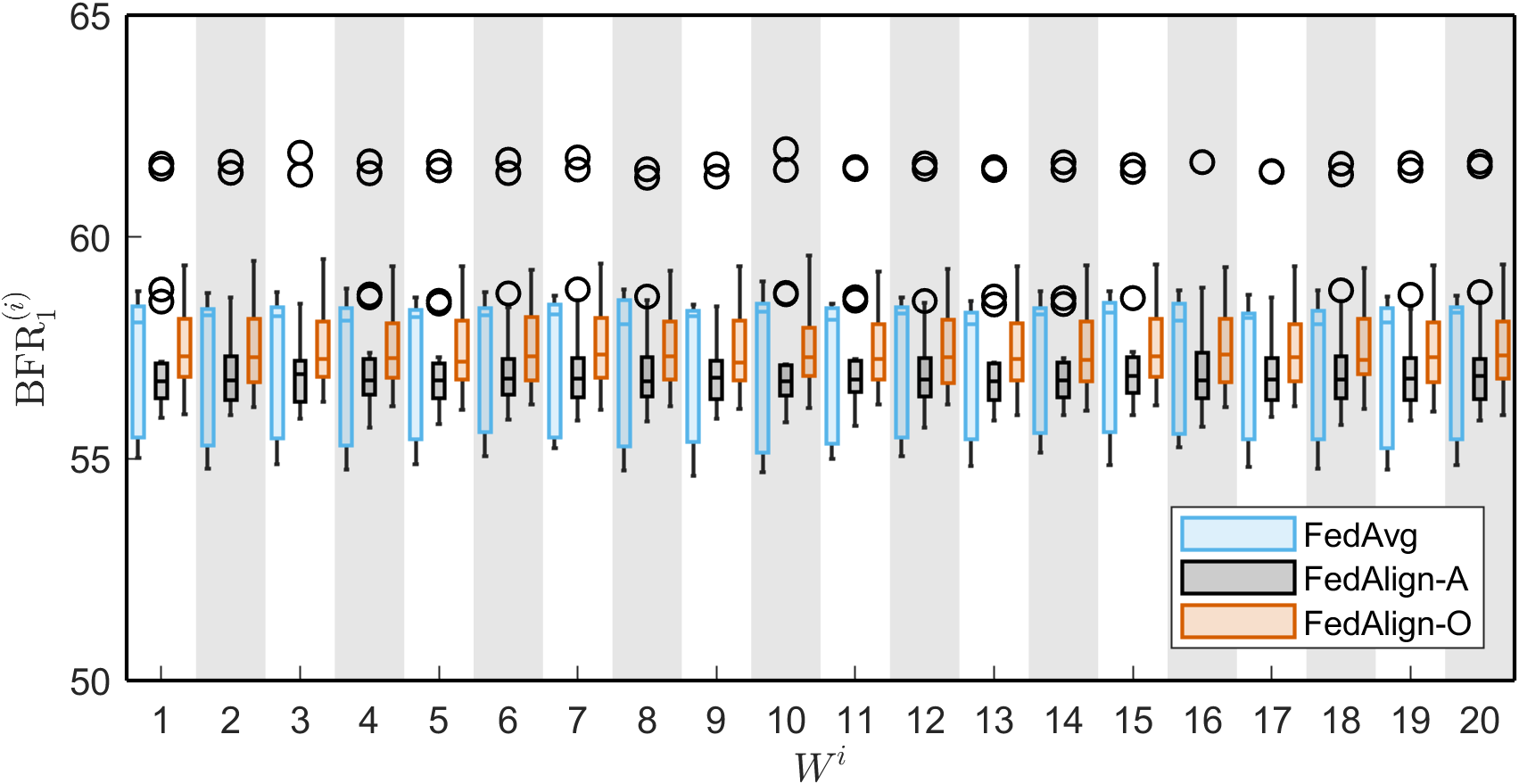}
    }
    \caption{Box plot comparison of FedAlign and FedAvg on Piezoelectric Dataset}
    \label{fig:piezobox}
\end{figure}

\begin{figure}[t]
    \centering
    \subfloat[MR Damper: FedAlign \& FedAvg for $iter=1$]{
        \includegraphics[width=0.49\textwidth]{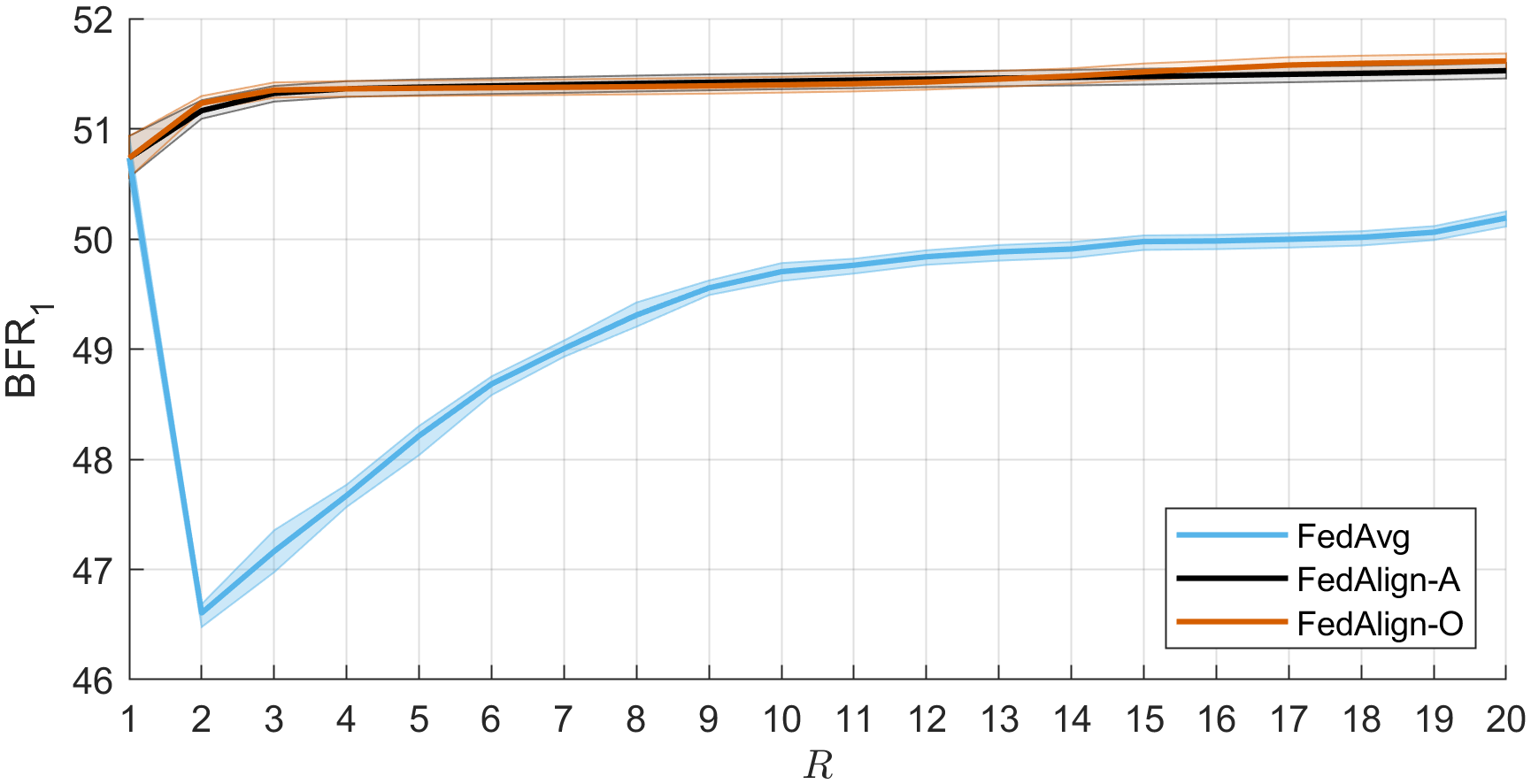}
    }
    \subfloat[MR Damper: FedAlign for $iter=1$, FedAvg for $iter=20$]{
        \includegraphics[width=0.49\textwidth]{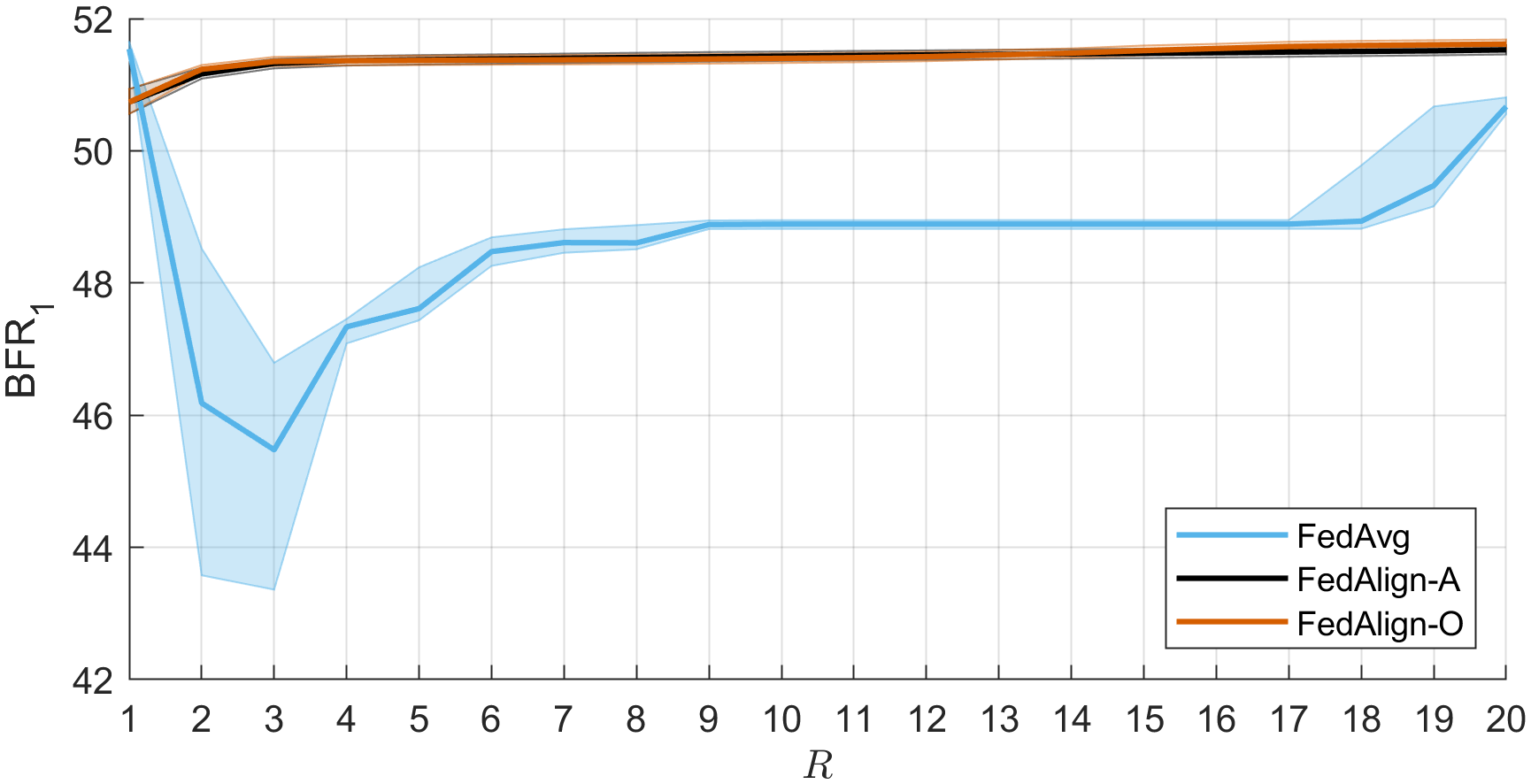}
    }
    \hfill
    \subfloat[Hair Dryer: FedAlign \& FedAvg for $iter=1$]{
        \includegraphics[width=0.49\textwidth]{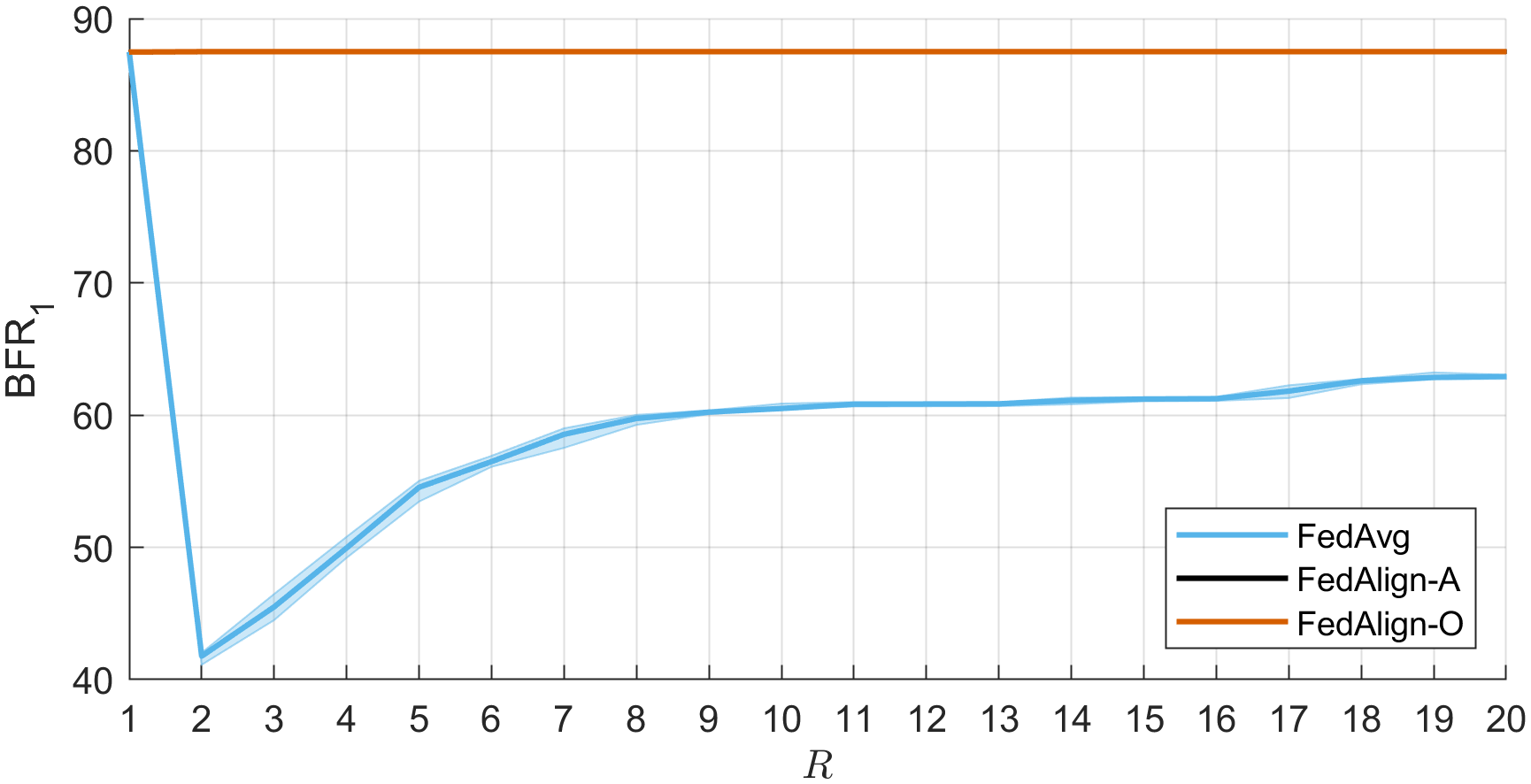}
    }
    \subfloat[Hair Dryer: FedAlign for $iter=1$, FedAvg for $iter=20$]{
        \includegraphics[width=0.49\textwidth]{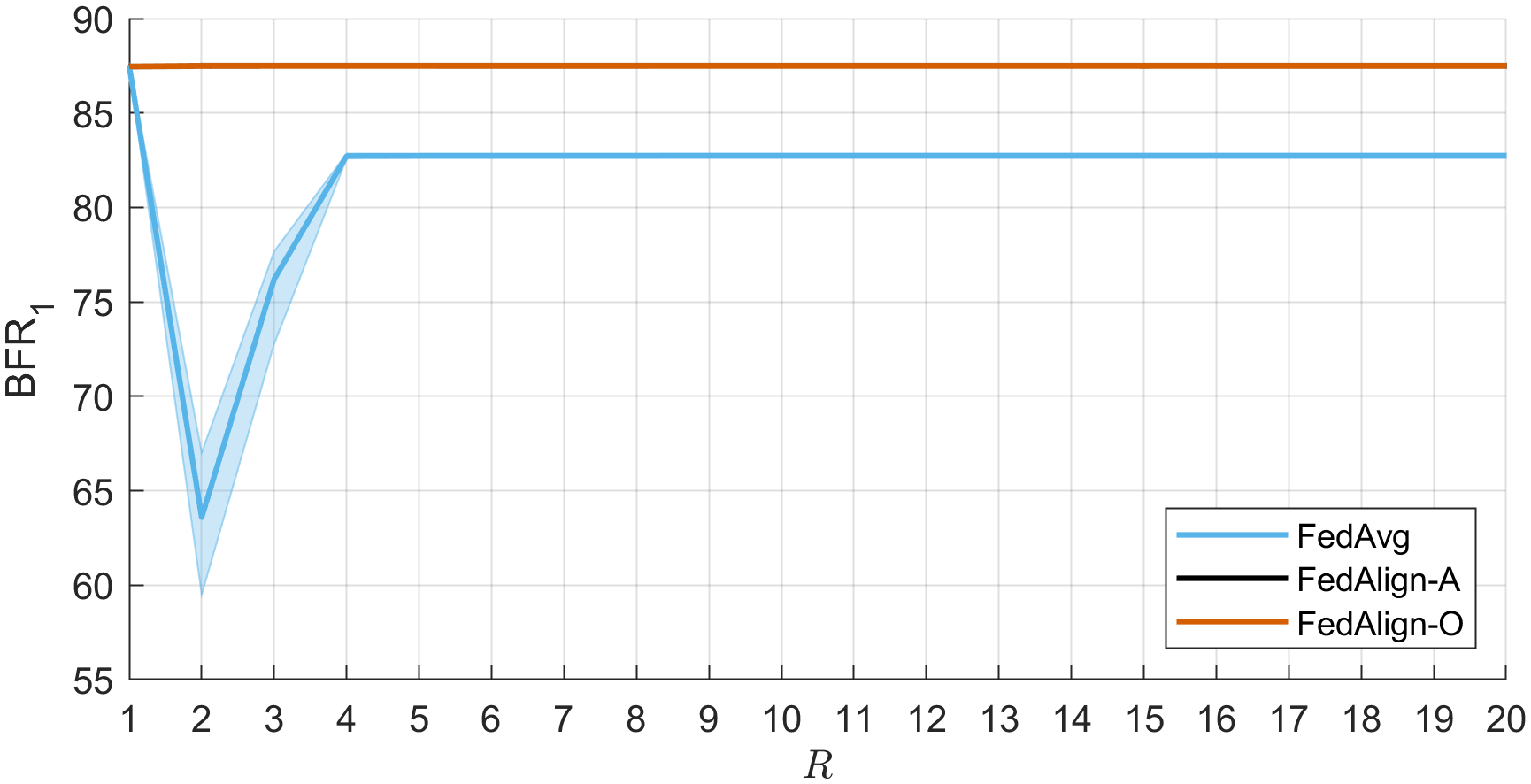}
    }
    \hfill
    \subfloat[Piezoelectric: FedAlign \& FedAvg for $iter=1$]{
        \includegraphics[width=0.49\textwidth]{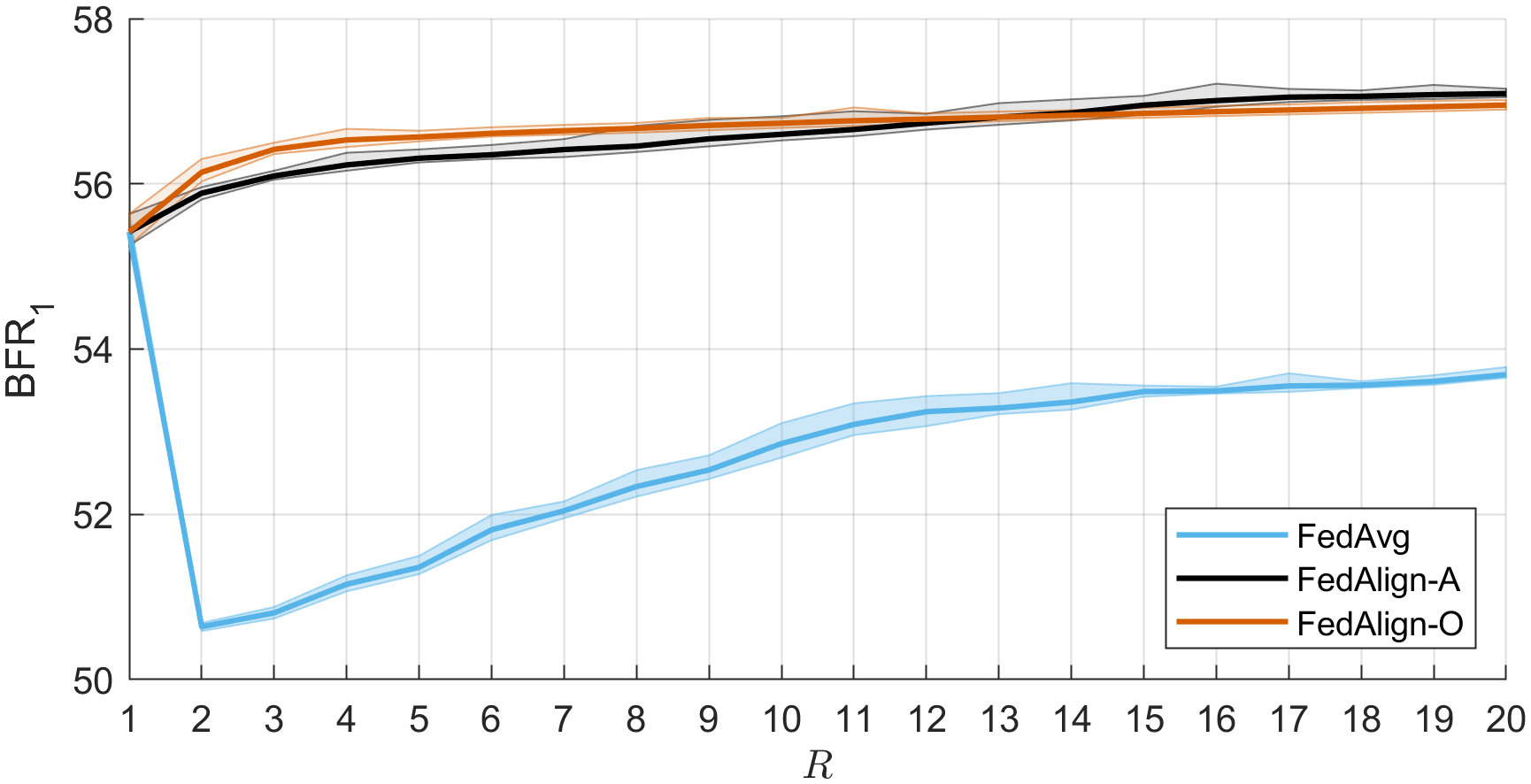}
    }
    \subfloat[Piezoelectric: FedAlign for $iter=1$, FedAvg for $iter=20$]{
        \includegraphics[width=0.49\textwidth]{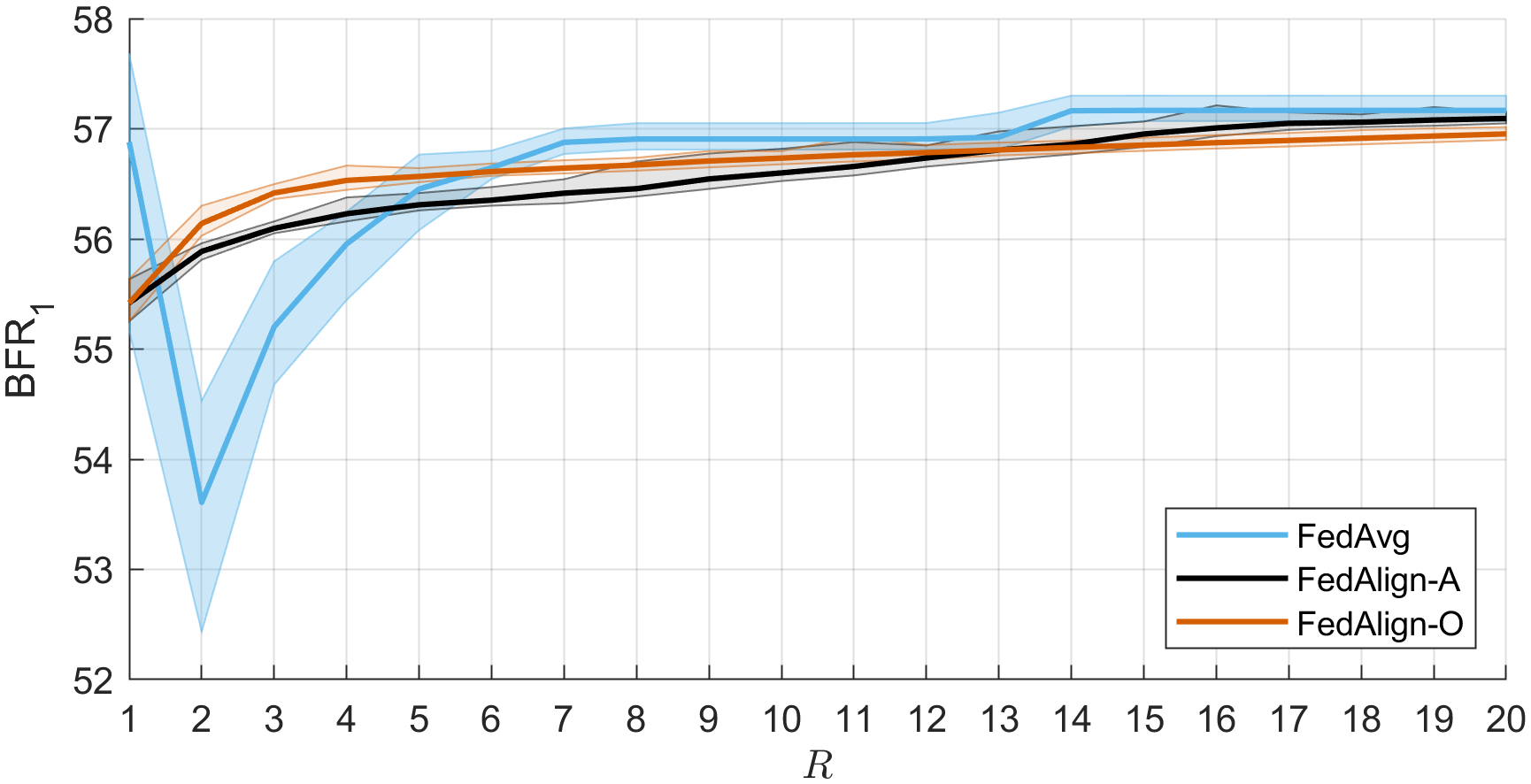}
    }
    \caption{Comparison of FedAlign and FedAvg during training on real-world datasets: mean $\text{BFR}^{(i)}$ across local workers with solid lines, while minimum, and maximum $\text{BFR}^{(i)}$ across local workers with shaded areas.}
    \label{fig:realworldlfitSISO}
\end{figure}

We reported the mean BFR values with standard errors alongside \#UM and \#F2L in Table \ref{tab:benchmark}. We also evaluated $\text{BFR}$ using $D_{\text{test}}$ to analyze the global SSM's performance against test data. We gave box plots of $\text{BFR}^{(i)}$ for FedAvg and FedAlign in Fig. \ref{fig:mrdamperbox} - Fig. \ref{fig:piezobox}. Furthermore, we illustrated the $\text{BFR}^{(i)}$ progression across communication rounds in Fig. \ref{fig:realworldlfitSISO}. From the results, we infer that: 
\begin{itemize}
    \item For $iter=1$, FedAlign constantly shows superior performances on all datasets as seen in Fig. \ref{fig:mrdamperbox} - Fig. \ref{fig:piezobox}. Although using higher local iterations ($iter=20$) improves the performance of FedAvg, it matches FedAlign only in the Piezoelectric dataset. Even with $iter=20$ setup, FedAvg exhibits lower performance with higher deviation in MR Damper and Hair Dryer datasets as shown in Fig. \ref{fig:mrdamperbox201} and Fig. \ref{fig:hairdryerbox201}. 
    \item As demonstrated in Fig. \ref{fig:realworldlfitSISO}, FedAlign converges faster with steady performance, whereas FedAvg displays significant performance decreases during early communication rounds. 
    \item FedAvg obtains unstable global SSMs in the Hair Dryer and Piezoelectric datasets. On the other hand, FedAlign generates only stable global SSMs, thanks to local parameter basin alignment. Neither of the methods yields any F2L global SSM.
    \item FedAlign evaluates higher test BFR in all datasets for $iter=1$. FedAvg with $iter=20$ outperforms FedAlign against unseen data in the Piezoelectric dataset, whereas it still obtains fewer test BFR in the MR Damper and Hair Dryer datasets. 
    \item FedAlign-A and FedAlign-O exhibit similar performances against training and test data. However, it is worth noting that FedAlign-O deviates less in the Piezoelectric dataset.
\end{itemize}

To sum up, the efficient alignment of local parameter basins enables FedAlign to excel in both training and testing, alongside faster convergence and enhanced stability, even with fewer local iterations.

\begin{table}[t]
\centering
\footnotesize
\renewcommand{\arraystretch}{1.4}
\setlength{\tabcolsep}{3pt} 
\caption{Performance analysis over 20 experiments}\label{tab:benchmark}
\begin{tabular}{@{}lllccccc@{}}
\toprule
\multicolumn{3}{c}{} 
& \multicolumn{2}{c}{\textbf{FedAvg}} 
& \textbf{FedAlign-A} 
& \textbf{FedAlign-O} 
\\ \cmidrule(lr){4-5} \cmidrule(lr){6-7}
\multicolumn{3}{c}{}  
& \textit{iter=1} & \textit{iter=20}  
& \textit{iter=1}  
& \textit{iter=1}  
\\ \midrule

\multirow{4}{*}{MR Damper}    
 & \multirow{2}{*}{\textbf{BFR}} & Train  
 & \(50.58 (\pm 2.54)\) & \(51.06 (\pm 2.04)\)  
 & \(51.95 (\pm 0.08)\) & \(52.06 (\pm 0.08)\)  
\\  
 &  & Test  
 & \(58.30 (\pm 2.98)\)  & \(58.40 (\pm 2.86)\)  
 & \(59.82 (\pm 0.18)\) & \(60.04 (\pm 0.07)\)  
\\ 
 & \multicolumn{2}{l}{\textbf{\#UM}}  
 & 0 & 0  
 & 0 & 0  
\\  
 & \multicolumn{2}{l}{\textbf{\#F2L}}  
 & 0 & 0  
 & 0 & 0  
\\ \midrule

\multirow{4}{*}{Hair Dryer}    
 & \multirow{2}{*}{\textbf{BFR}} & Train  
 & \(63.05 (\pm 22.63)\) & \(82.88 (\pm 14.73)\)  
 & \(87.66 (\pm 0.07)\) & \(87.66 (\pm 0.07)\)  
\\  
 &  & Test  
 & \(61.17 (\pm 24.23)\) & \(82.85 (\pm 15.97)\)  
 & \(87.91 (\pm 0.05)\) & \(87.89 (\pm 0.05)\)  
\\  
 & \multicolumn{2}{l}{\textbf{\#UM}}  
 & 1 & 0  
 & 0 & 0  
\\  
 & \multicolumn{2}{l}{\textbf{\#F2L}}  
 & 0 & 0  
 & 0 & 0  
\\ \midrule

\multirow{4}{*}{Piezoelectric}    
 & \multirow{2}{*}{\textbf{BFR}} & Train  
 & \(53.80 (\pm 5.04)\) & \(57.31 (\pm 1.44)\)  
 & \(57.30 (\pm 1.62)\) & \(57.46 (\pm 0.92)\)  
\\  
 &  & Test  
 & \(53.66 (\pm 5.26)\) & \(60.94 (\pm 1.81)\)  
 & \(57.57 (\pm 1.57)\) & \(57.82 (\pm 0.98)\)  
\\ 
 & \multicolumn{2}{l}{\textbf{\#UM}}  
 & 0 & 1  
 & 0 & 0  
\\  
 & \multicolumn{2}{l}{\textbf{\#F2L}}  
 & 0 & 0  
 & 0 & 0  
\\ \bottomrule

\end{tabular}

\end{table}

\subsection{Performance comparison study on real-world MIMO datasets}
\label{mimocomp}

We conducted our performance evaluation on the following presplit datasets, each of which is already divided into training and test subsets, denoted by $D = \{D_{\text{train}}, D_{\text{test}}\} $.

\begin{itemize}
    \item Steam Engine dataset \cite{steameng} contains 451 samples with steam pressure and magnetization voltage as inputs ($nu = 2$) and generated voltage and rotational speed as the outputs ($ny = 2$). $D_{train}$ includes first 250 samples and $D_{test}$ includes the remaining samples. To create diverse datasets $\forall W^i$, $\boldsymbol{v}_{1:K}^{(i)} \sim \mathcal{N}(0, 0.001^2 I_{2})$ is sampled and added into $D_{train}$.
    \item The CD player data set \cite{cdplayer} consists of 2,048 samples, with mechanical actuator forces ($nu = 2$) as input and arm tracking accuracy as output ($ny = 2$). \( D_{\text{train}} \) includes the first 1200 samples, and \( D_{\text{test}} \) consists of remaining samples. The dataset is normalized to achieve zero mean and unit variance. Noise \( \boldsymbol{v}_{1:K}^{(i)} \sim \mathcal{N}(0, 0.05^2I_{2}) \) is added to \( D_{\text{train}} \) for each \( W^i \).
    \item The Evaporator dataset contains \cite{evaporator} 6305 samples with feed and vapor flows to the first evaporator stage and cooling water flow as inputs ($nu = 3$) and dry matter content, flow, and temperature of the outcoming product as outputs ($ny = 3$). The dataset is normalized to achieve zero mean and unit variance. $D_{train}$ consists of the first 3000 samples, and $D_{test}$ includes samples 3001-6000. Gaussian noise, $\boldsymbol{v}_{1:K}^{(i)} \sim \mathcal{N}(0, 0.1^2I_{3})$, is added to $D_{train}$ for each \( W^i \).
\end{itemize}

In this analysis, we set $iter = 1$ in FedAlign and $iter = \{1,20\}$ in FedAvg as in real-world SISO datasets. We set $nx = 4$ for the Steam Engine and the Evaporator, and $nx = 2$ for the CD Player datasets. We utilized $\boldsymbol{u}_{1:K}^{(i)} \in  D_{\text{test}}^i$ to generate $\boldsymbol{\tilde{x}}^{(i)}_{pseudo}$ in FedAlign-O. We set $\mu_1 = nx$ and constructed $M^{(i)} = P^{(i)}_1$ in FedAlign-A.

\begin{table}[t]
\centering
\footnotesize

\renewcommand{\arraystretch}{1.4}
\setlength{\tabcolsep}{3pt} 
\caption{Performance analysis over 20 experiments}\label{tab:benchmarkMIMO}
\begin{tabular}{@{}lllccccc@{}}
\toprule
\multicolumn{3}{c}{} 
& \multicolumn{2}{c}{\textbf{FedAvg}} 
& \textbf{FedAlign-A} 
& \textbf{FedAlign-O} 
\\ \cmidrule(lr){4-5} \cmidrule(lr){6-7}
\multicolumn{3}{c}{}  
& \textit{iter=1} & \textit{iter=20}  
& \textit{iter=1}  
& \textit{iter=1}  
\\ \midrule

\multirow{6}{*}{Steam Engine}    
 & \multirow{2}{*}{\textbf{BFR\(_1\)}} & Train  
 & \(91.09 (\pm 2.46)\) & \(91.64 (\pm 0.26)\)  
 & \(91.48 (\pm 0.002)\) & \(91.49 (\pm 0.002)\)  
\\  
 &  & Test  
 & \(89.31 (\pm 1.20)\) & \(89.68 (\pm 0.23)\)  
 & \(89.81 (\pm 0.001)\) & \(89.81 (\pm 0.001)\)  
\\  
 & \multirow{2}{*}{\textbf{BFR\(_2\)}} & Train  
 & \(68.53 (\pm 9.74)\)  & \(70.59 (\pm 1.70)\)  
 & \(69.68 (\pm 0.008)\) & \(69.68 (\pm 0.008)\)  
\\  
 &  & Test  
 & \(46.25 (\pm 6.78)\) & \(47.93 (\pm 1.93)\)  
 & \(49.13 (\pm 0.005)\) & \(49.14 (\pm 0.005)\)  
\\ 
 & \multicolumn{2}{l}{\textbf{\#UM}}  
 & 0 & 0  
 & 0 & 0  
\\  
 & \multicolumn{2}{l}{\textbf{\#F2L}}  
 & 0 & 0  
 & 0 & 0  
\\ \midrule

\multirow{6}{*}{CD Player}    
 & \multirow{2}{*}{\textbf{BFR\(_1\)}} & Train  
 & \(65.27 (\pm 16.17)\) & -  
 & \(70.71 (\pm 0.04)\) & \(70.71 (\pm 0.04)\)  
\\  
 &  & Test  
 & \(62.43 (\pm 15.73)\) & -  
 & \(66.91 (\pm 0.02)\) & \(66.91 (\pm 0.02)\)  
\\  
 & \multirow{2}{*}{\textbf{BFR\(_2\)}} & Train  
 & \(72.90 (\pm 20.42)\) & -  
 & \(78.90 (\pm 0.03)\) & \(78.90 (\pm 0.03)\)  
\\  
 &  & Test  
 & \(73.20 (\pm 19.48)\) & -  
 & \(79.10 (\pm 0.01)\) & \(79.10 (\pm 0.01)\)  
\\ 
 & \multicolumn{2}{l}{\textbf{\#UM}}  
 & 6 & 20  
 & 0 & 0  
\\  
 & \multicolumn{2}{l}{\textbf{\#F2L}}  
 & 1 & 0  
 & 0 & 0  
\\ \midrule

\multirow{8}{*}{Evaporator}    
 & \multirow{2}{*}{\textbf{BFR\(_1\)}} & Train  
 & \(53.16 (\pm 0.58)\) & \(54.01 (\pm 0.05)\)  
 & \(53.43 (\pm 0.31)\) & \(53.49 (\pm 0.07)\)  
\\  
 &  & Test  
 & \(46.54 (\pm 0.38)\) & \(46.96 (\pm 0.05)\)  
 & \(46.76 (\pm 0.16)\) & \(46.79 (\pm 0.06)\)  
\\ 
 & \multirow{2}{*}{\textbf{BFR\(_2\)}} & Train  
 & \(50.11 (\pm 0.50)\) & \(49.26 (\pm 0.04)\)  
 & \(49.86 (\pm 0.25)\) & \(49.81 (\pm 0.06)\)  
\\  
 &  & Test  
 & \(45.42 (\pm 1.04)\) & \(43.80 (\pm 0.05)\)  
 & \(44.72 (\pm 0.47)\) & \(44.63 (\pm 0.08)\)  
\\ 
 & \multirow{2}{*}{\textbf{BFR\(_3\)}} & Train  
 & \(51.11 (\pm 0.03)\) & \(51.13 (\pm 0.03)\)  
 & \(51.11 (\pm 0.03)\) & \(51.12 (\pm 0.03)\)  
\\  
 &  & Test  
 & \(54.40 (\pm 0.06)\) & \(54.49 (\pm 0.02)\)  
 & \(54.45 (\pm 0.03)\) & \(54.45 (\pm 0.03)\)  
\\ 
 & \multicolumn{2}{l}{\textbf{\#UM}}  
 & 1 & 0  
 & 0 & 0  
\\  
 & \multicolumn{2}{l}{\textbf{\#F2L}}  
 & 0 & 0  
 & 0 & 0  
\\ \bottomrule

\end{tabular}

\end{table}

\begin{figure}[t]
    \centering
    \subfloat{%
        \includegraphics[scale=0.5]{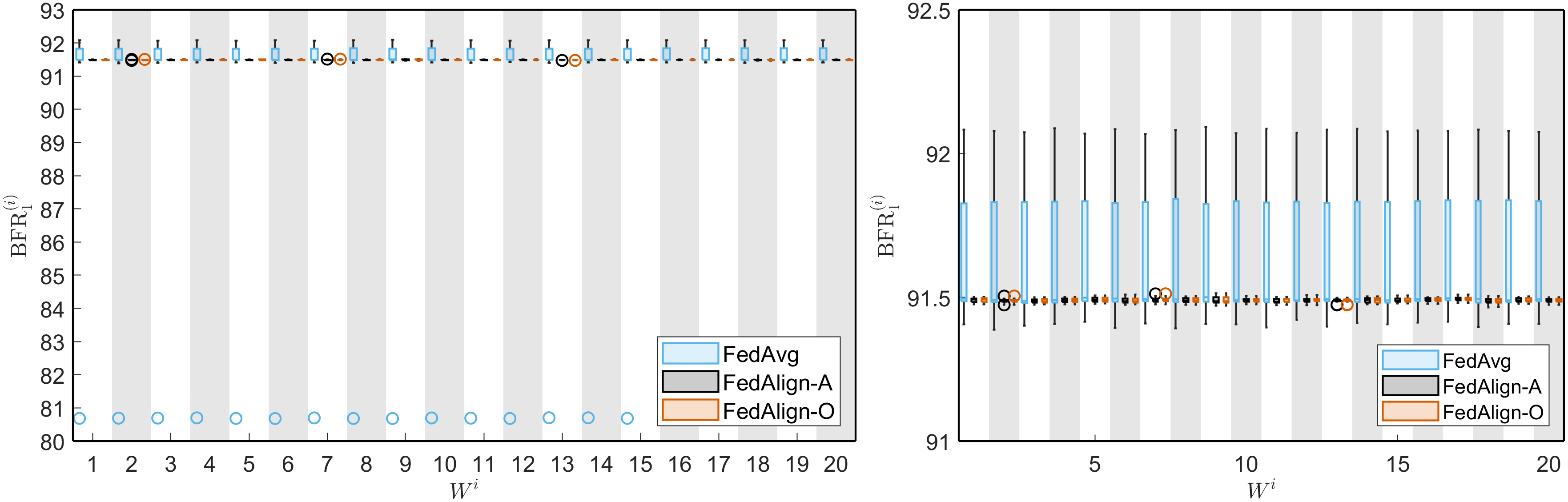}
    }\\
    \subfloat{%
        \includegraphics[scale=0.5]{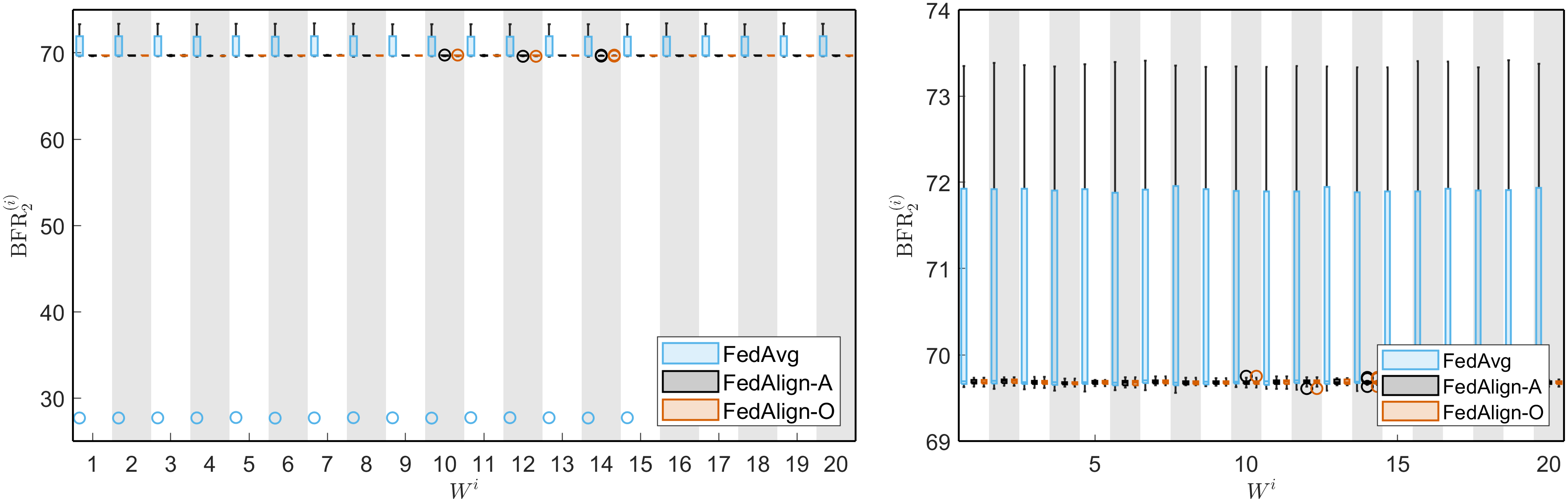}
    }
    \caption{Box plot comparison of FedAlign and FedAvg with $iter=1$ on the Steam Engine dataset for two outputs \(y_1\) (top row) and \(y_2\) (bottom row). Full-scale plot on the left, zoomed-in plot on the right.}
    \label{fig:steamEngbox1}
\end{figure}

\begin{figure}[t]
    \centering
    \subfloat{%
        \includegraphics[scale=0.5]{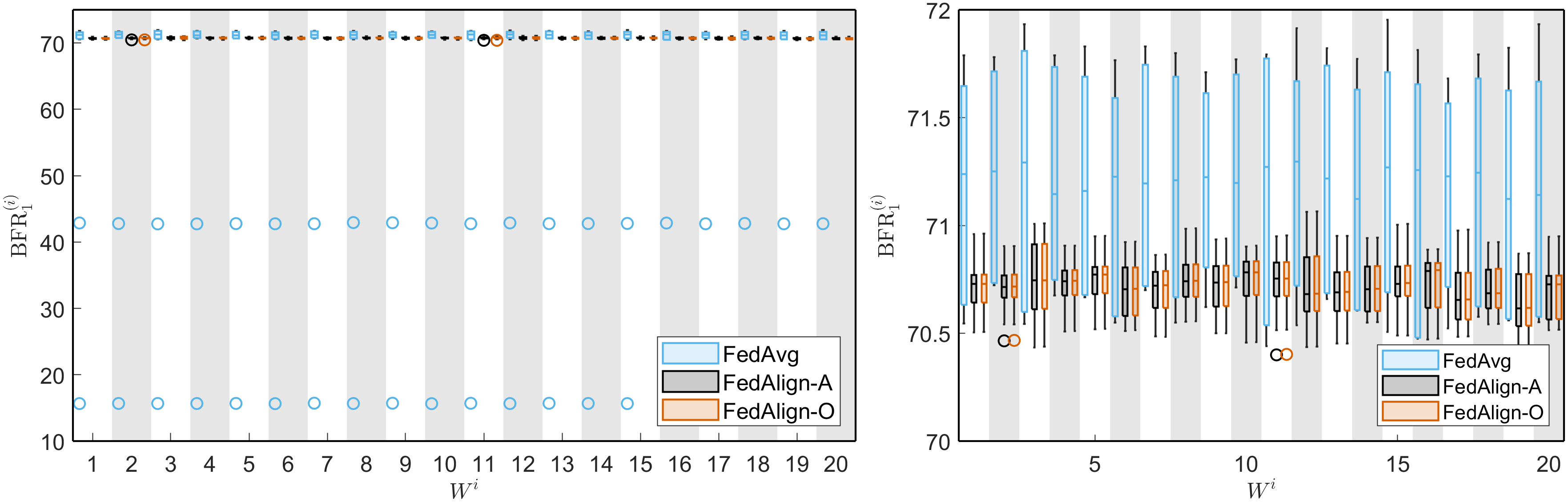}
    }\\
    \subfloat{%
        \includegraphics[scale=0.5]{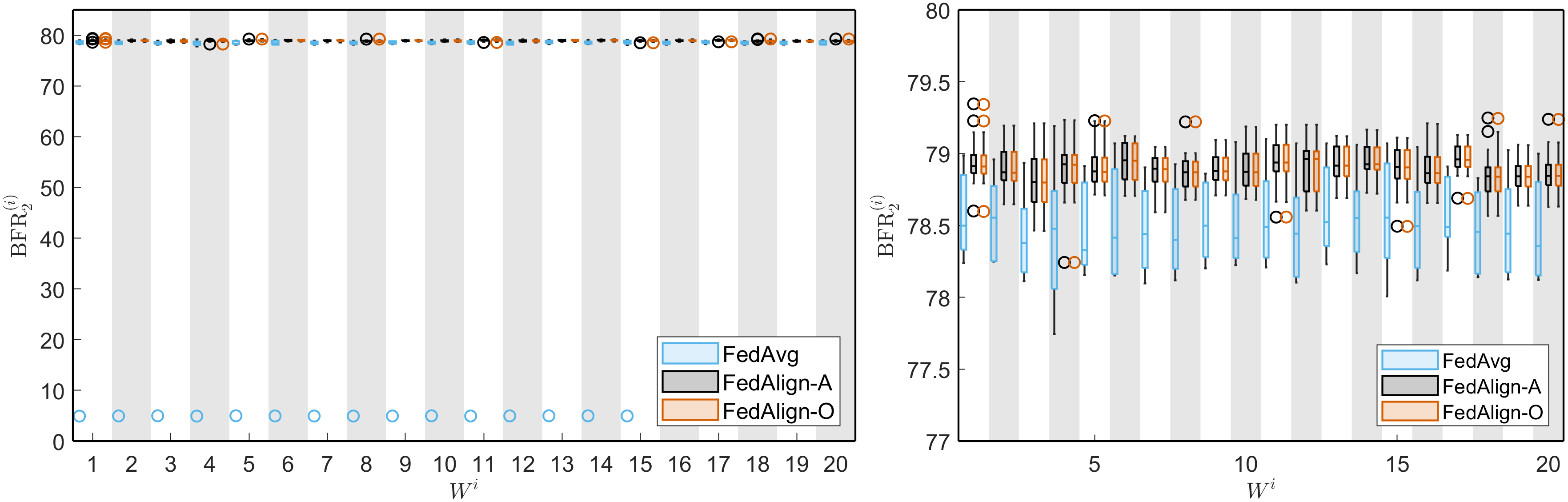}
    }
    \caption{Box plot comparison of FedAlign and FedAvg with $iter=1$ on the CD Player dataset for two outputs \(y_1\) (top row) and \(y_2\) (bottom row). Full-scale plot on the left, zoomed-in plot on the right.}
    \label{fig:cdPlayerbox}
\end{figure}

\begin{figure}[t]
    \centering
    \subfloat[FedAlign \& FedAvg for $iter=1$\label{fig:evaporatorbox-left}]{%
        \begin{minipage}{0.48\textwidth}
            \centering
            \includegraphics[width=\textwidth]{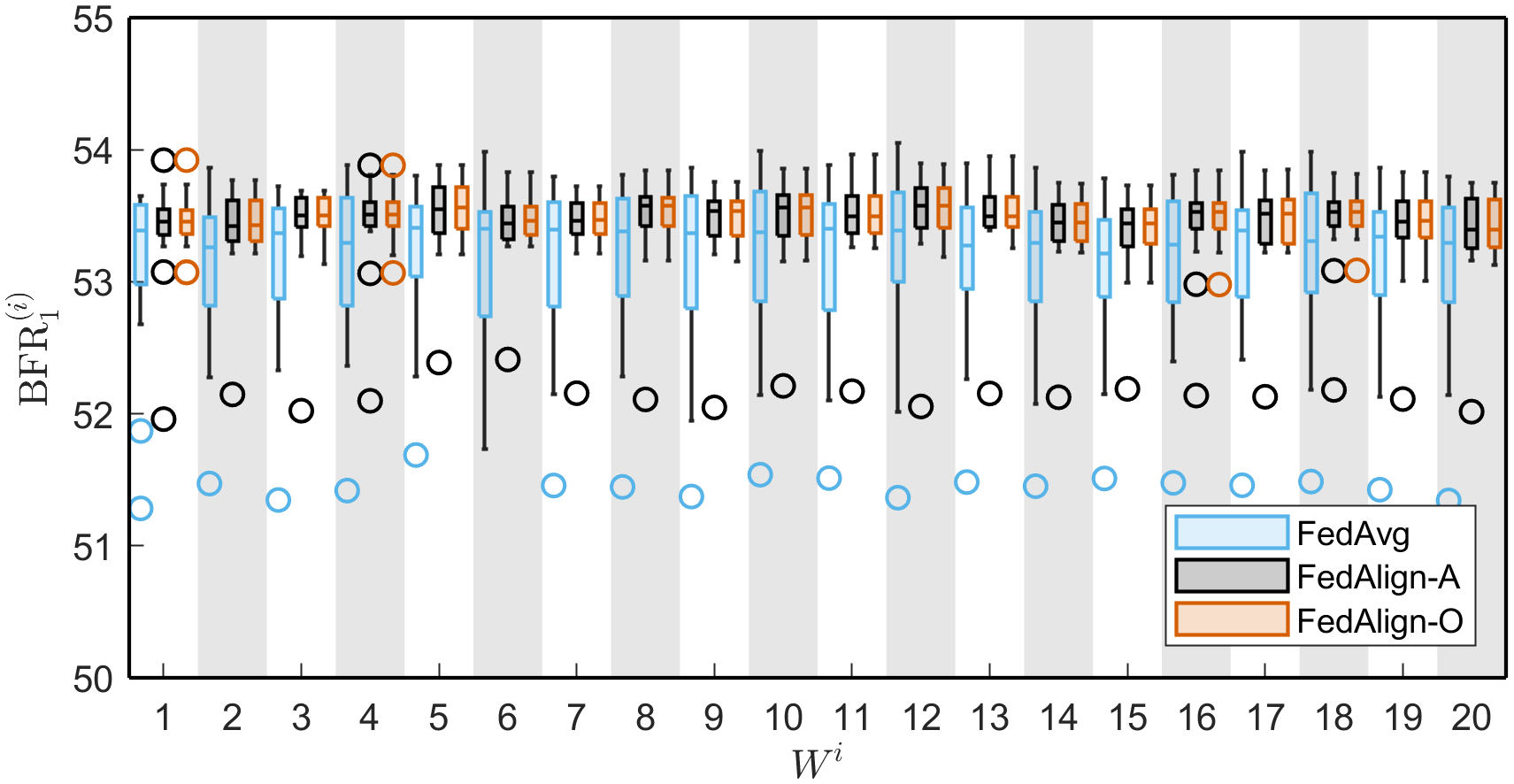}\\
            \vspace{1mm}  
            \includegraphics[width=\textwidth]{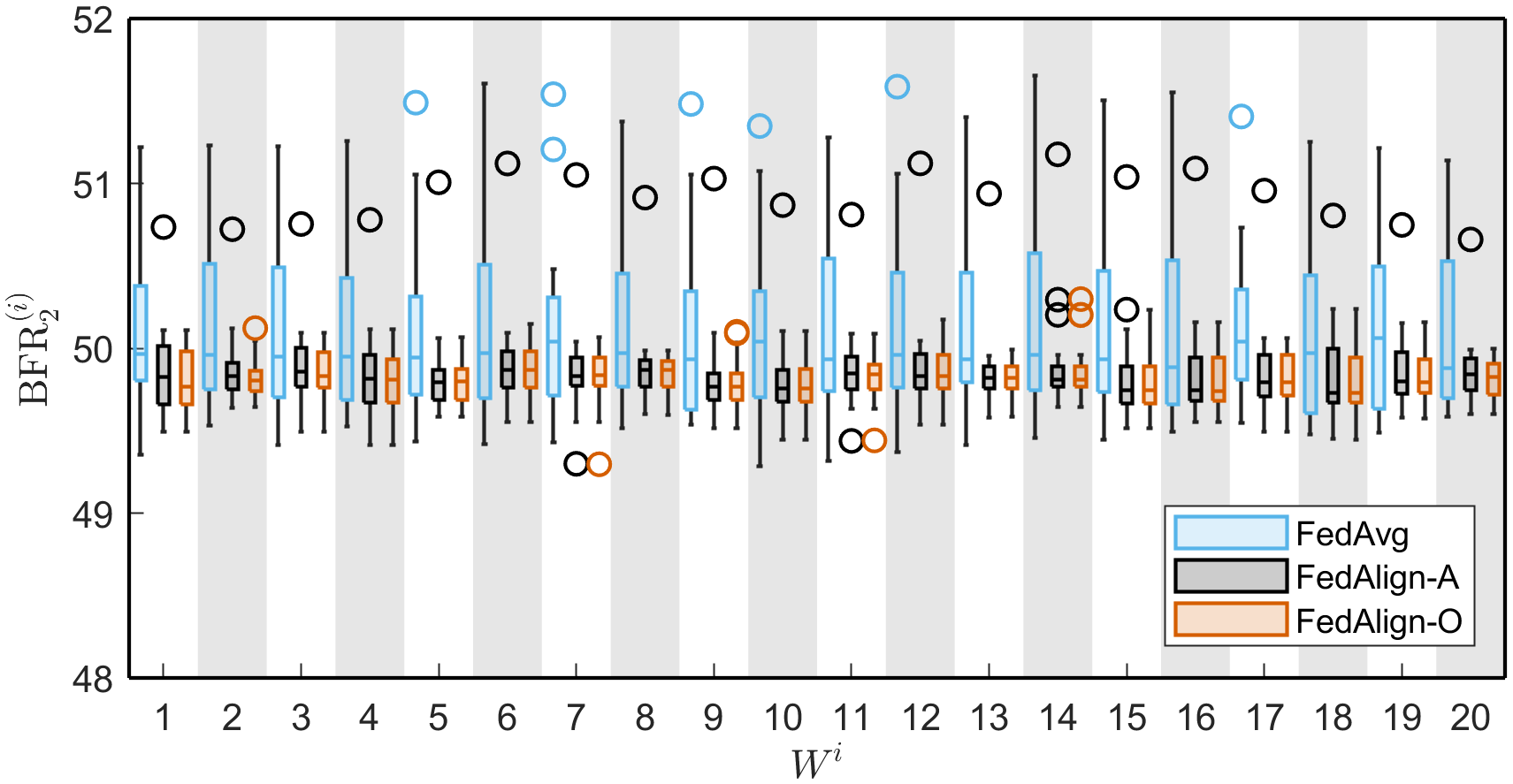}\\
            \vspace{1mm}
            \includegraphics[width=\textwidth]{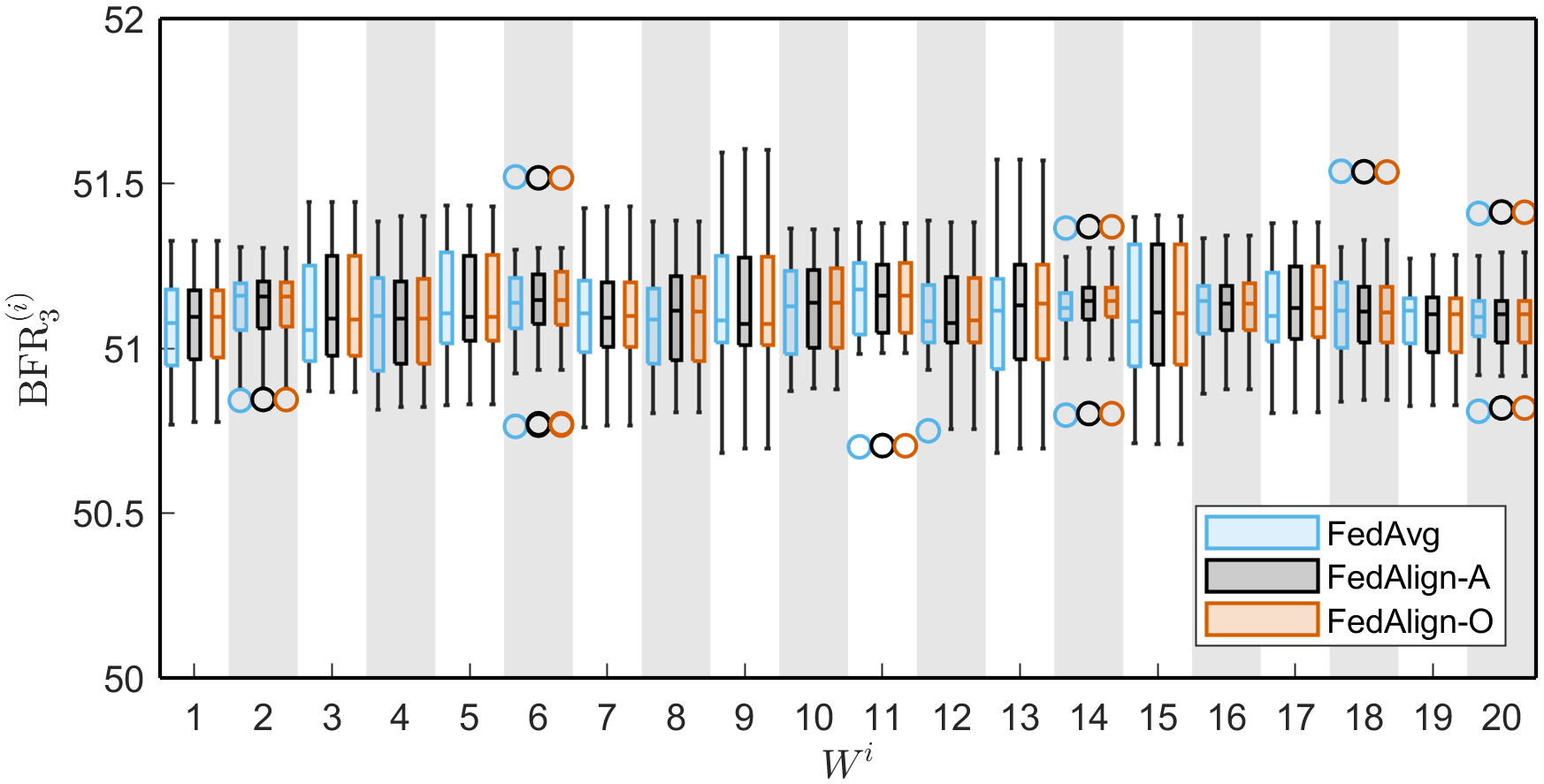}
        \end{minipage}
    }
    \hfill
    \subfloat[FedAlign for $iter=1$, FedAvg for $iter=20$\label{fig:evaporatorbox-right}]{%
        \begin{minipage}{0.48\textwidth}
            \centering
            \includegraphics[width=\textwidth]{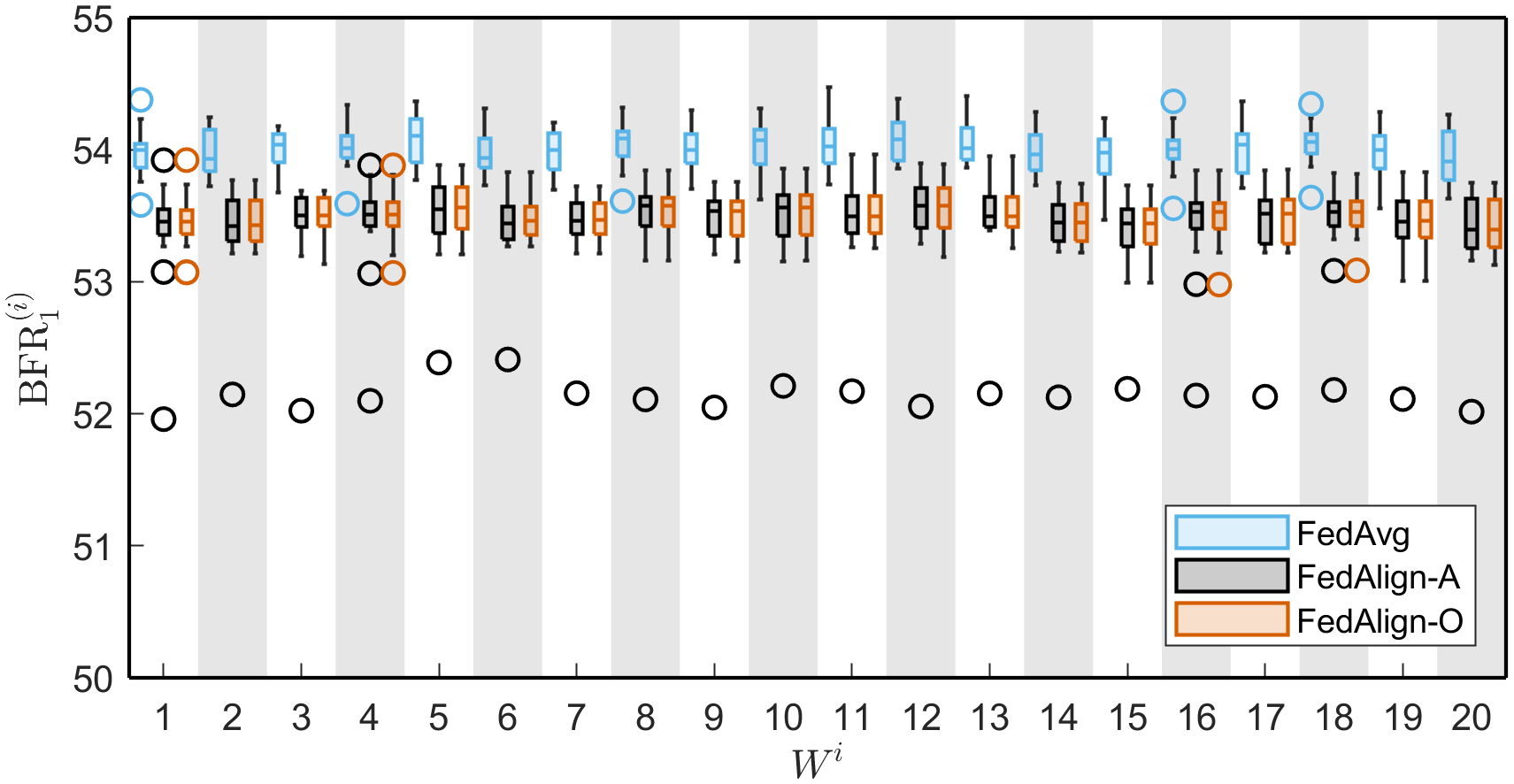}\\
            \vspace{1mm}
            \includegraphics[width=\textwidth]{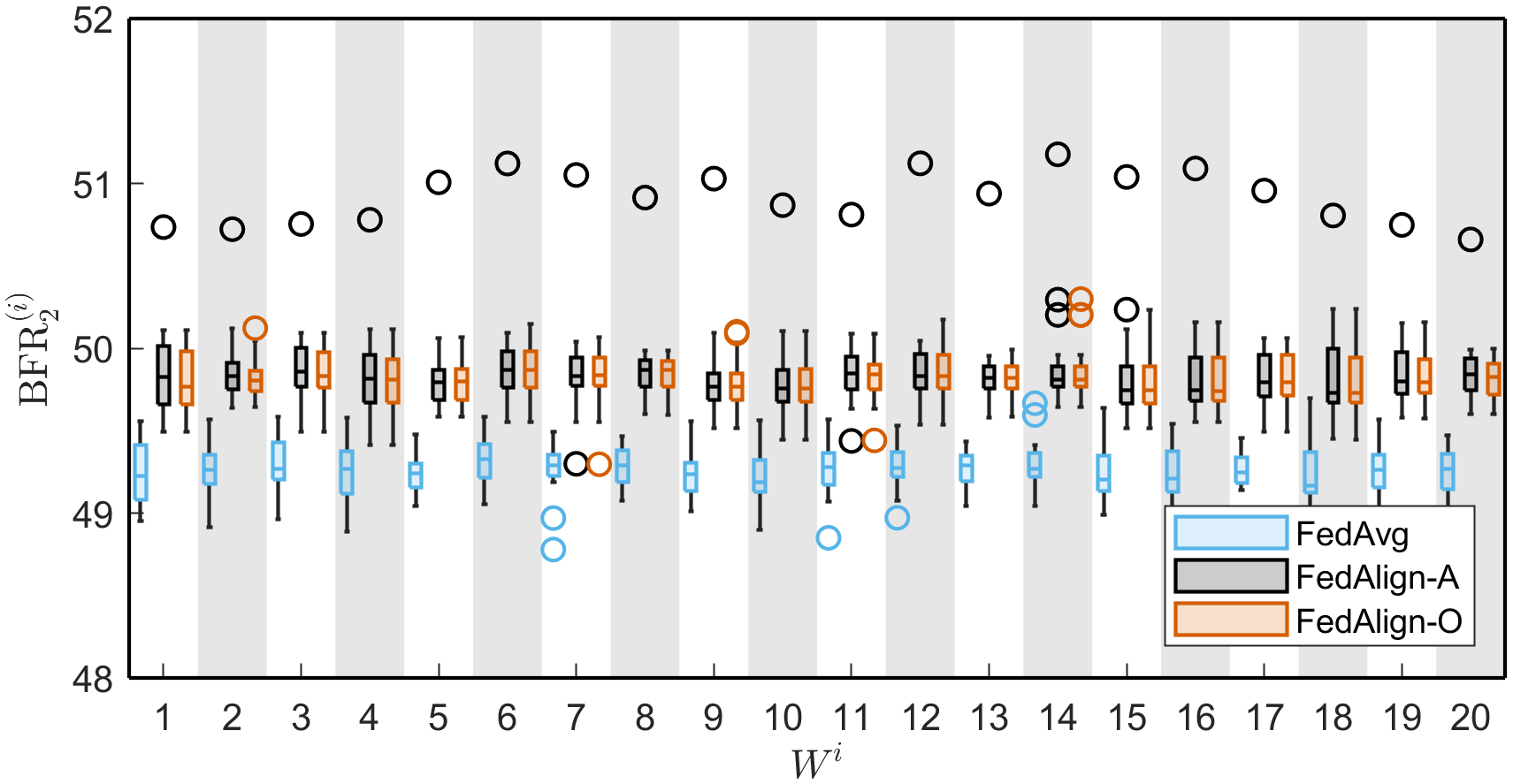}\\
            \vspace{1mm}
            \includegraphics[width=\textwidth]{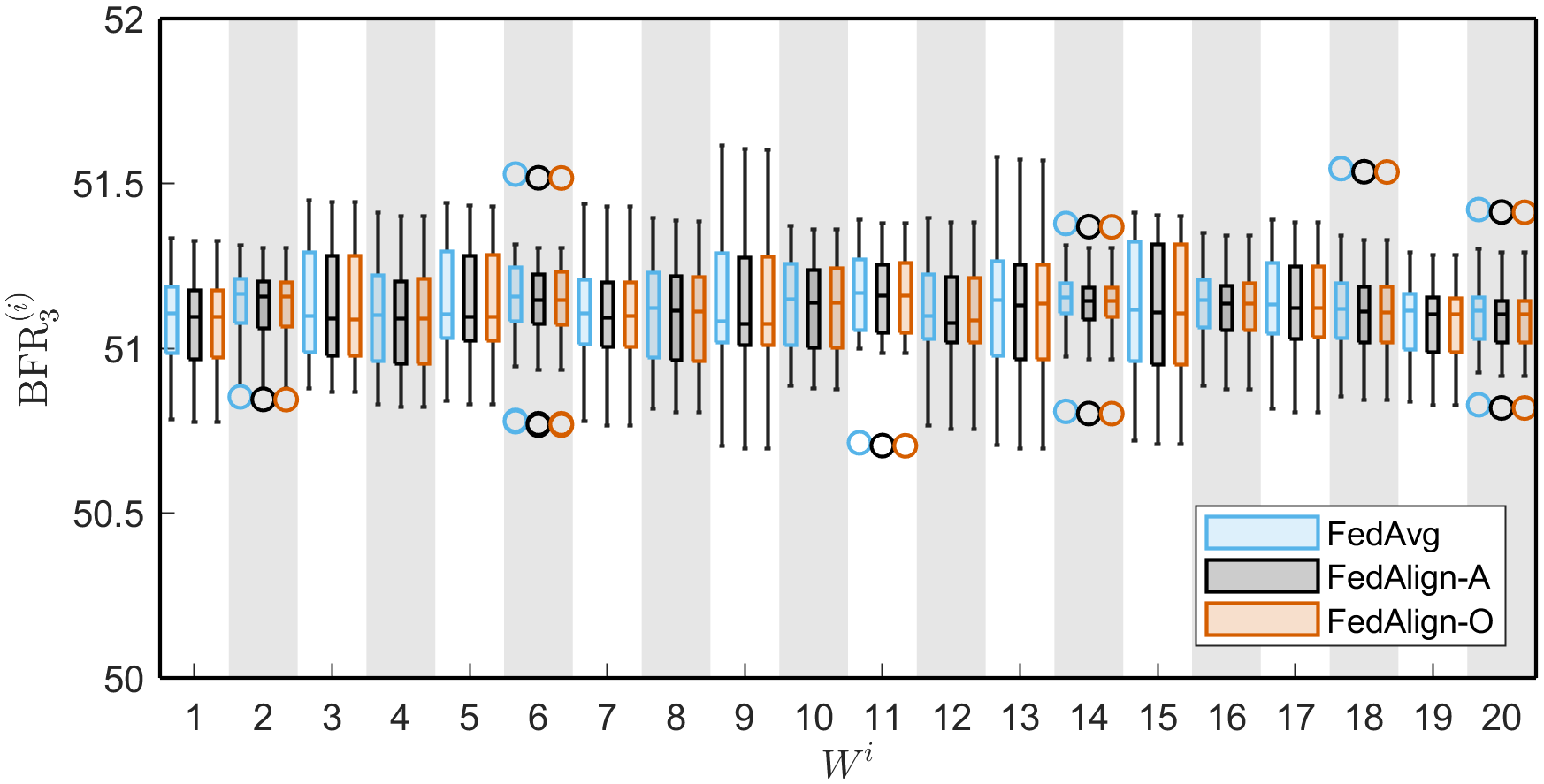}
        \end{minipage}
    }
    \caption{Box plot comparison of FedAlign and FedAvg on the Evaporator dataset for three outputs \(y_1\), \(y_2\), and \(y_3\). Each row shows a different output: \(y_1\) (top row), \(y_2\) (middle row), and \(y_3\) (bottom row).}
    \label{fig:evaporatorbox}
\end{figure}

\begin{figure}[t]
    \centering
    \subfloat{%
        \includegraphics[scale=0.55]{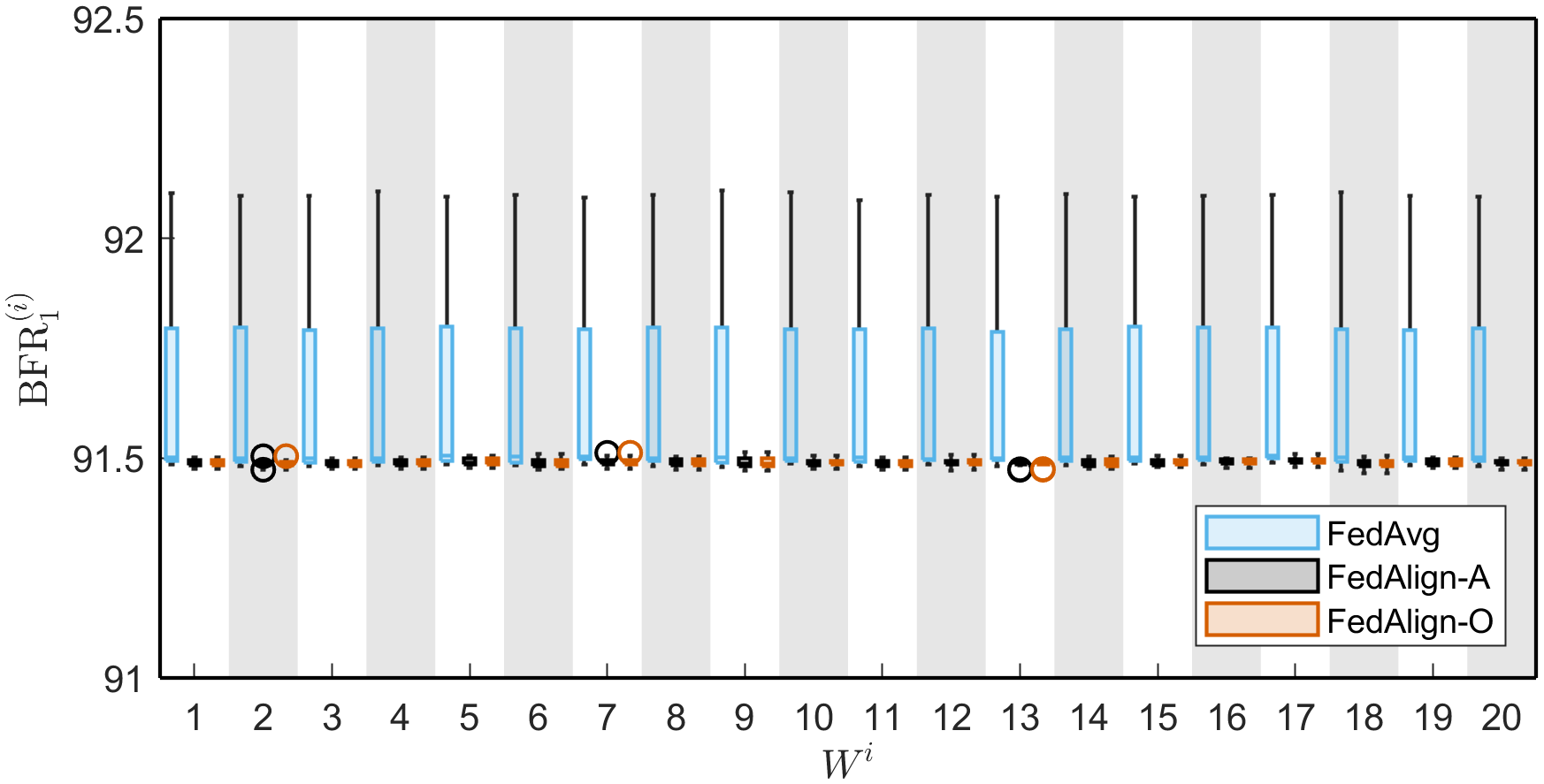}
    }
    \subfloat{%
        \includegraphics[scale=0.55]{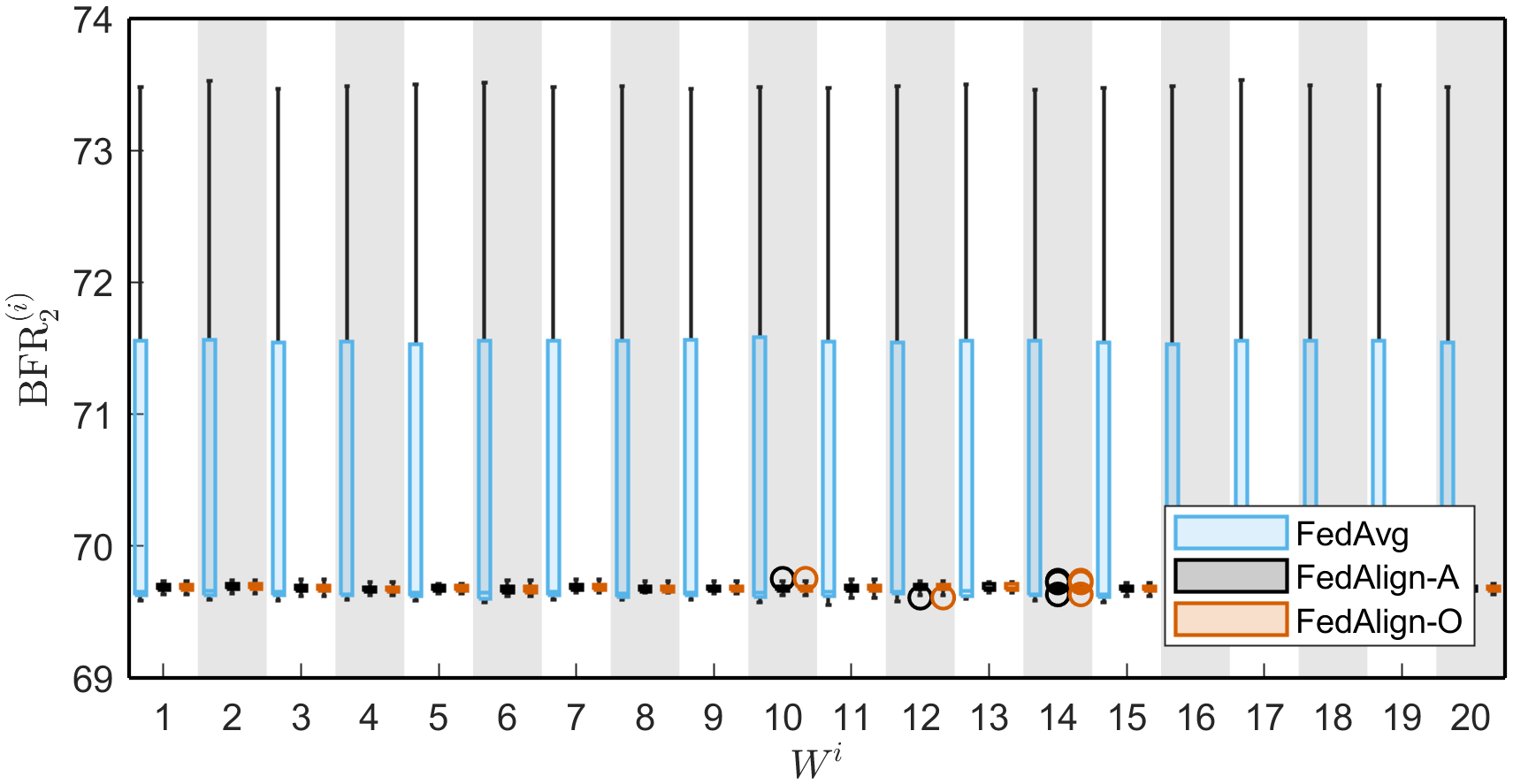}
    }
    \caption{Box plot comparison of FedAvg with $iter=20$ and FedAlign with $iter=1$ on the Steam Engine dataset for two outputs \(y_1\) (left) and \(y_2\) (right). }
    \label{fig:steamEngbox2}
\end{figure}

\begin{figure}[t]
    \centering
    %
    \subfloat[FedAlign \& FedAvg for $iter=1$]{%
        \begin{minipage}{0.48\textwidth}
            \centering
            \includegraphics[width=\textwidth]{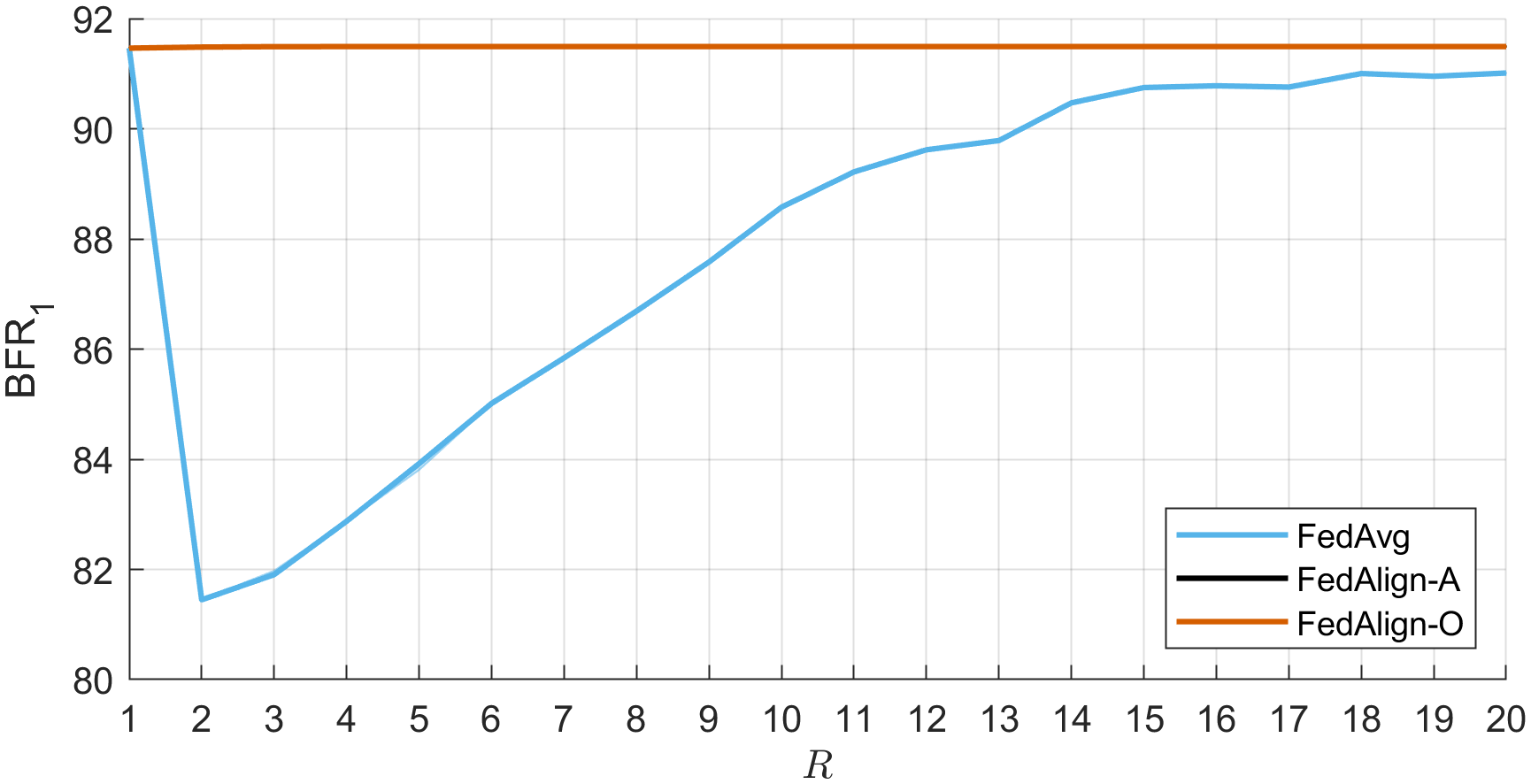}\\[1em]
            \includegraphics[width=\textwidth]{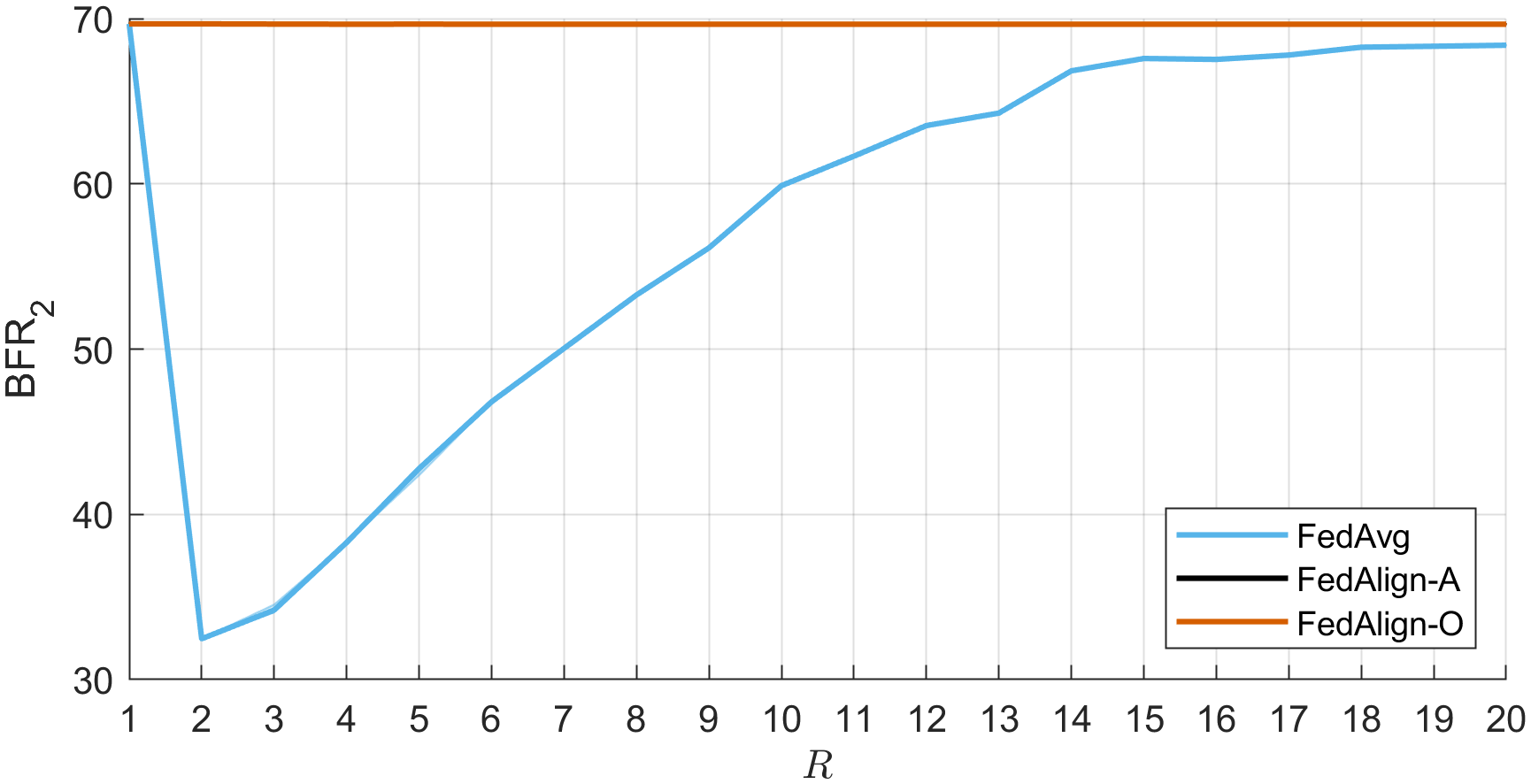}
        \end{minipage}
    }
    \hfill
    %
    \subfloat[FedAlign for $iter=1$, FedAvg for $iter=20$]{%
        \begin{minipage}{0.48\textwidth}
            \centering
            \includegraphics[width=\textwidth]{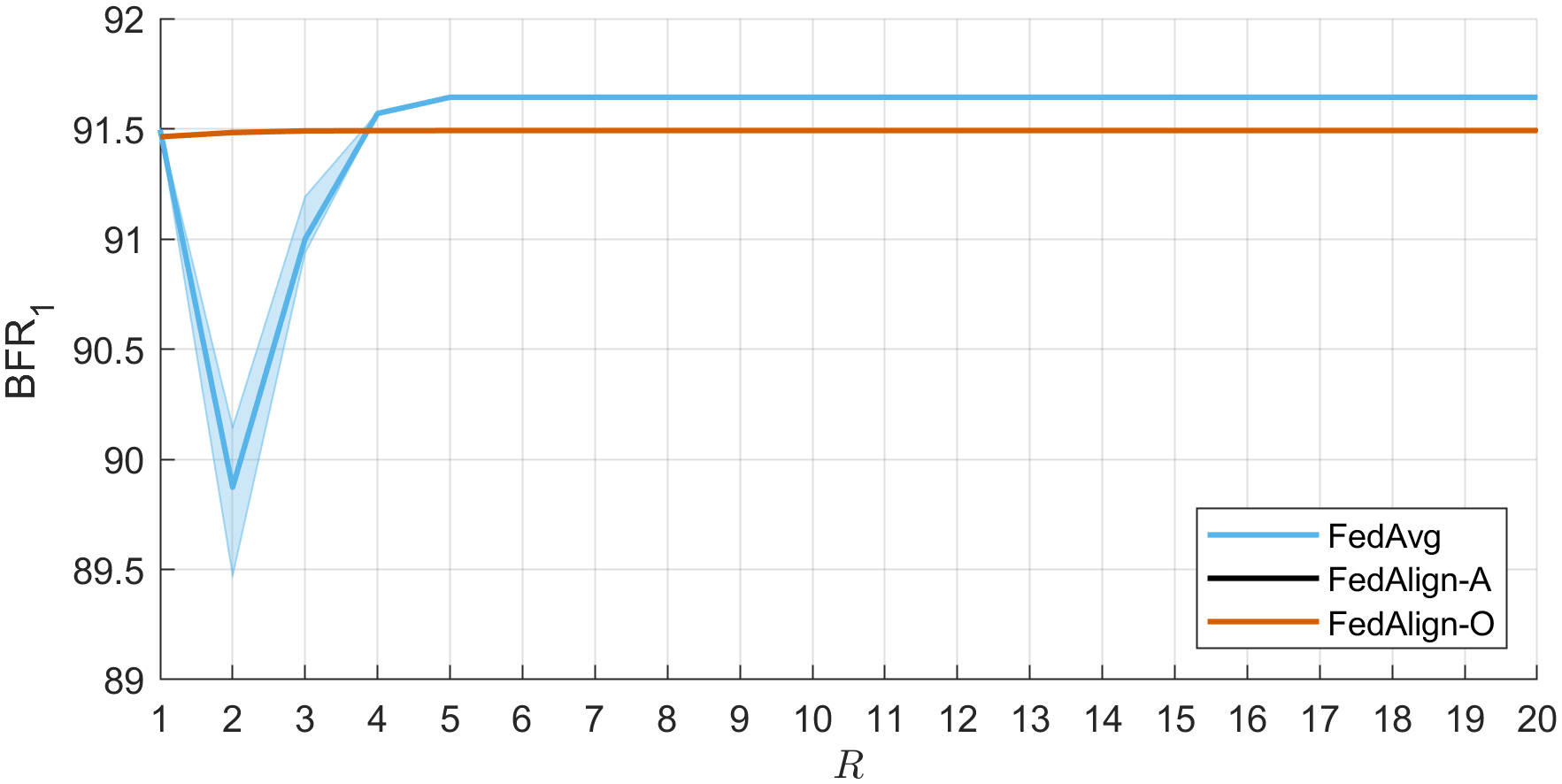}\\[1em]
            \includegraphics[width=\textwidth]{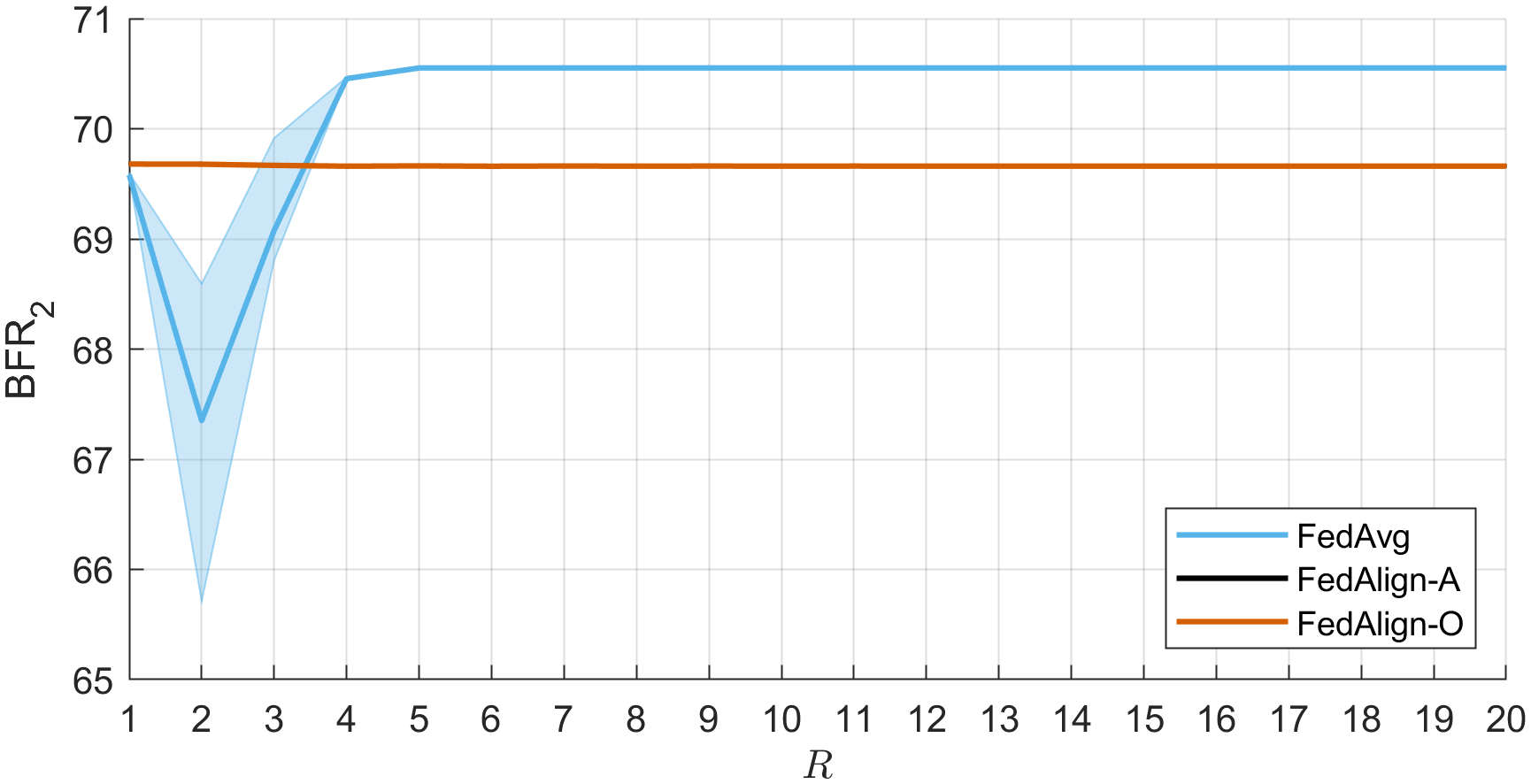}
        \end{minipage}
    }
    \caption{Comparison of FedAlign and FedAvg training on Steam Engine dataset for two outputs \(y_1\) (top row) and \(y_2\) (bottom row): Mean $\text{BFR}^{(i)}$ across local workers (solid lines) and the minimum and maximum $\text{BFR}^{(i)}$ across local workers (shaded areas).}
    \label{fig:steamEnglfit}
\end{figure}

\begin{figure}[t]
    \centering
    \subfloat[]{\includegraphics[width=0.5\textwidth]{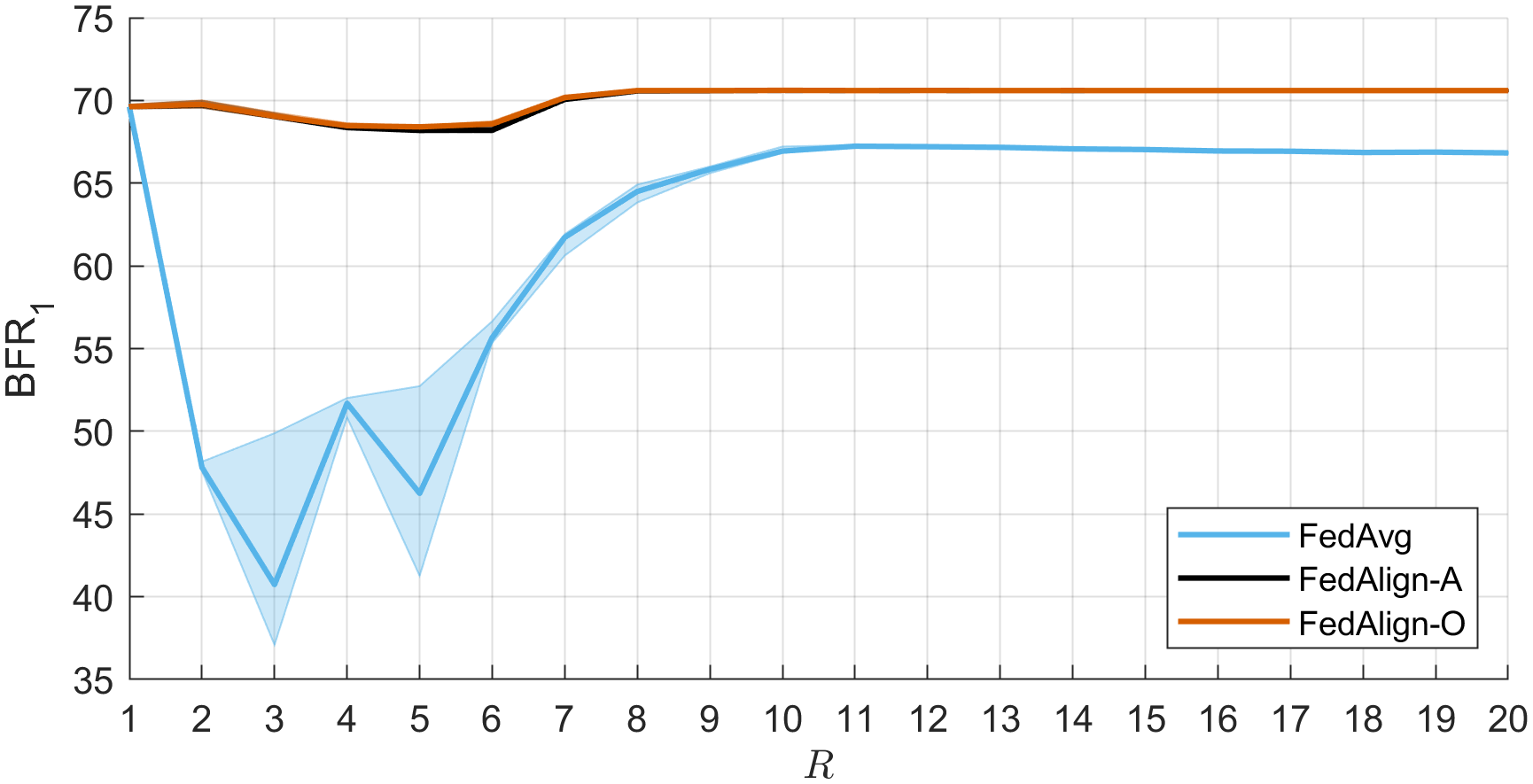}}
    \subfloat[]{\includegraphics[width=0.5\textwidth]{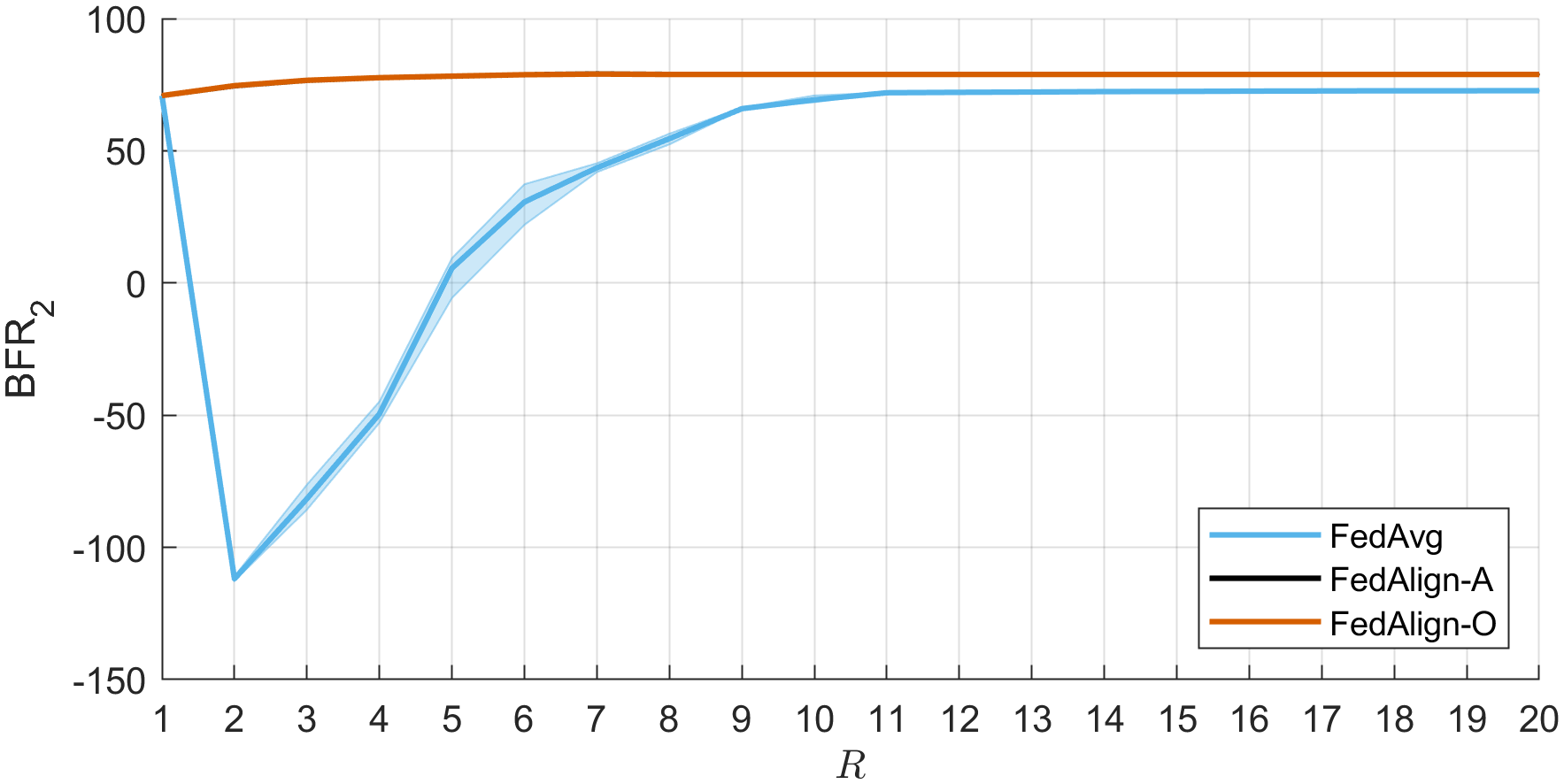}}
    \caption{Comparison of FedAlign and FedAvg with $iter=1$ Training on CD Player dataset: Mean $\text{BFR}^{(i)}$ across local workers (solid lines) and the minimum and maximum $\text{BFR}^{(i)}$ across local workers (shaded areas). (a) shows \(y_1\), 
    and (b) shows \(y_2\).}
    \label{fig:cdPlayerlfit}
\end{figure}

\begin{figure}[t]
    \centering
    \subfloat[FedAlign \& FedAvg for $iter=1$\label{fig:evaporatorlfit-left}]{%
        \begin{minipage}{0.48\textwidth}
            \centering
            \includegraphics[width=\textwidth]{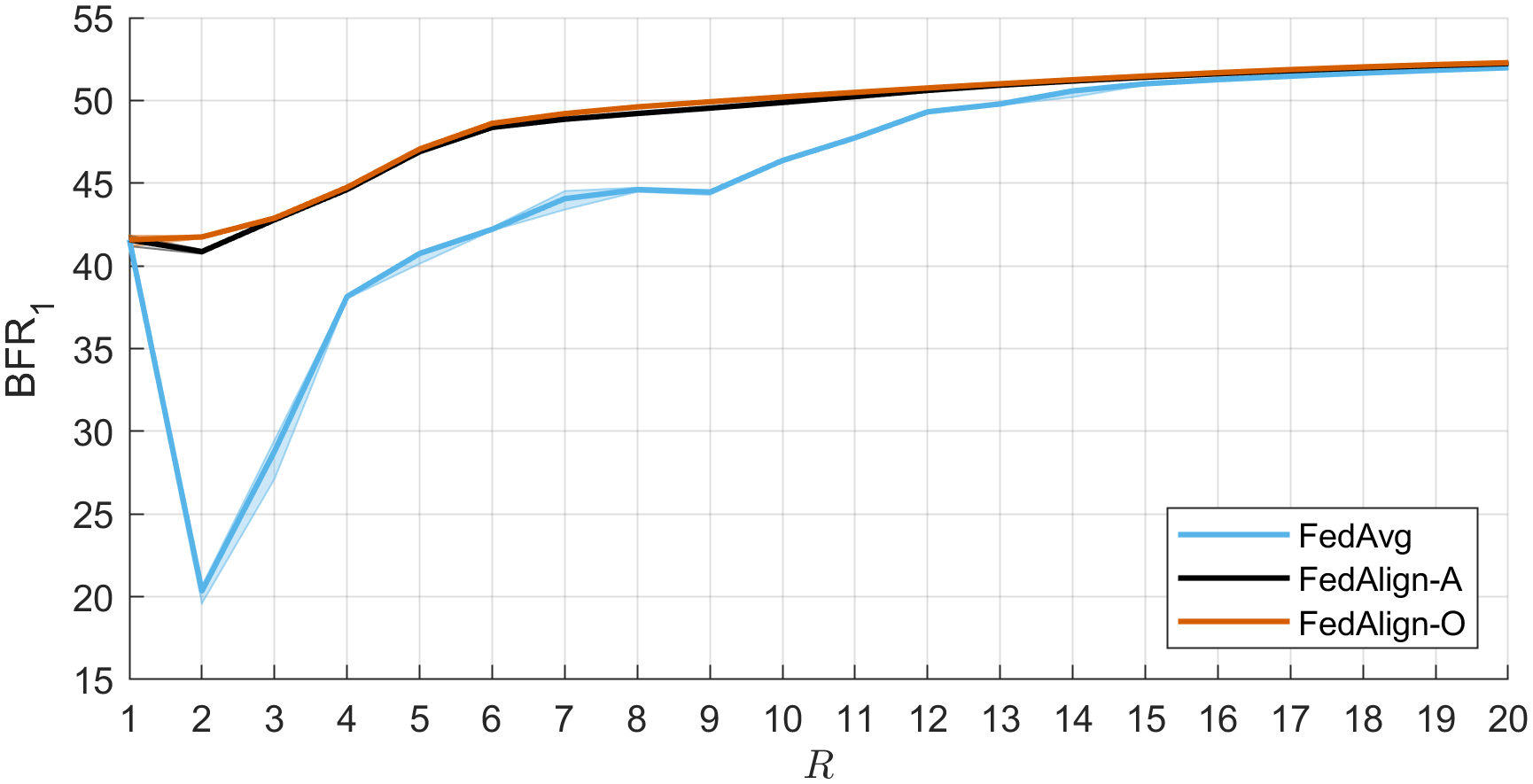}\\
            \vspace{1mm}  
            \includegraphics[width=\textwidth]{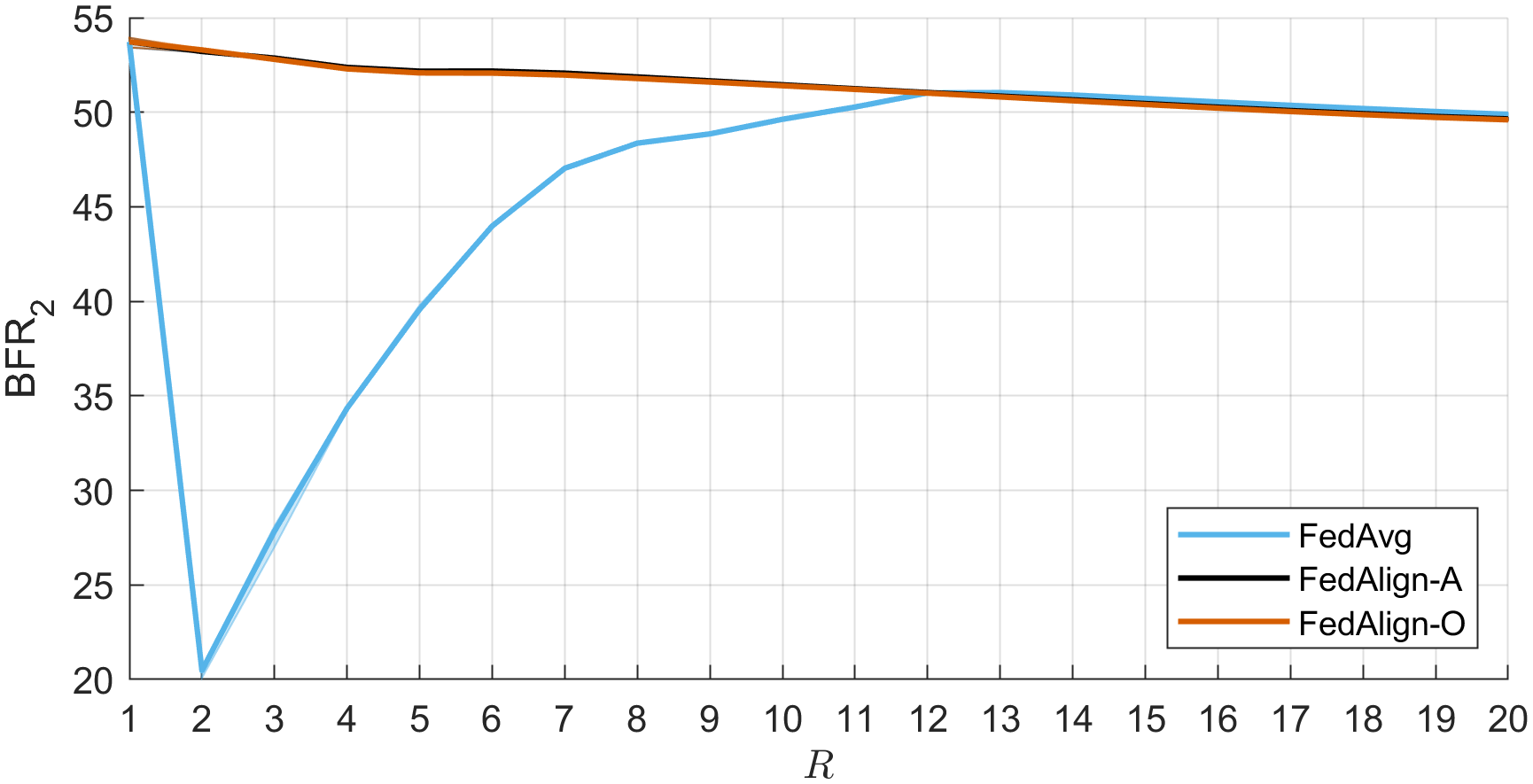}\\
            \vspace{1mm}
            \includegraphics[width=\textwidth]{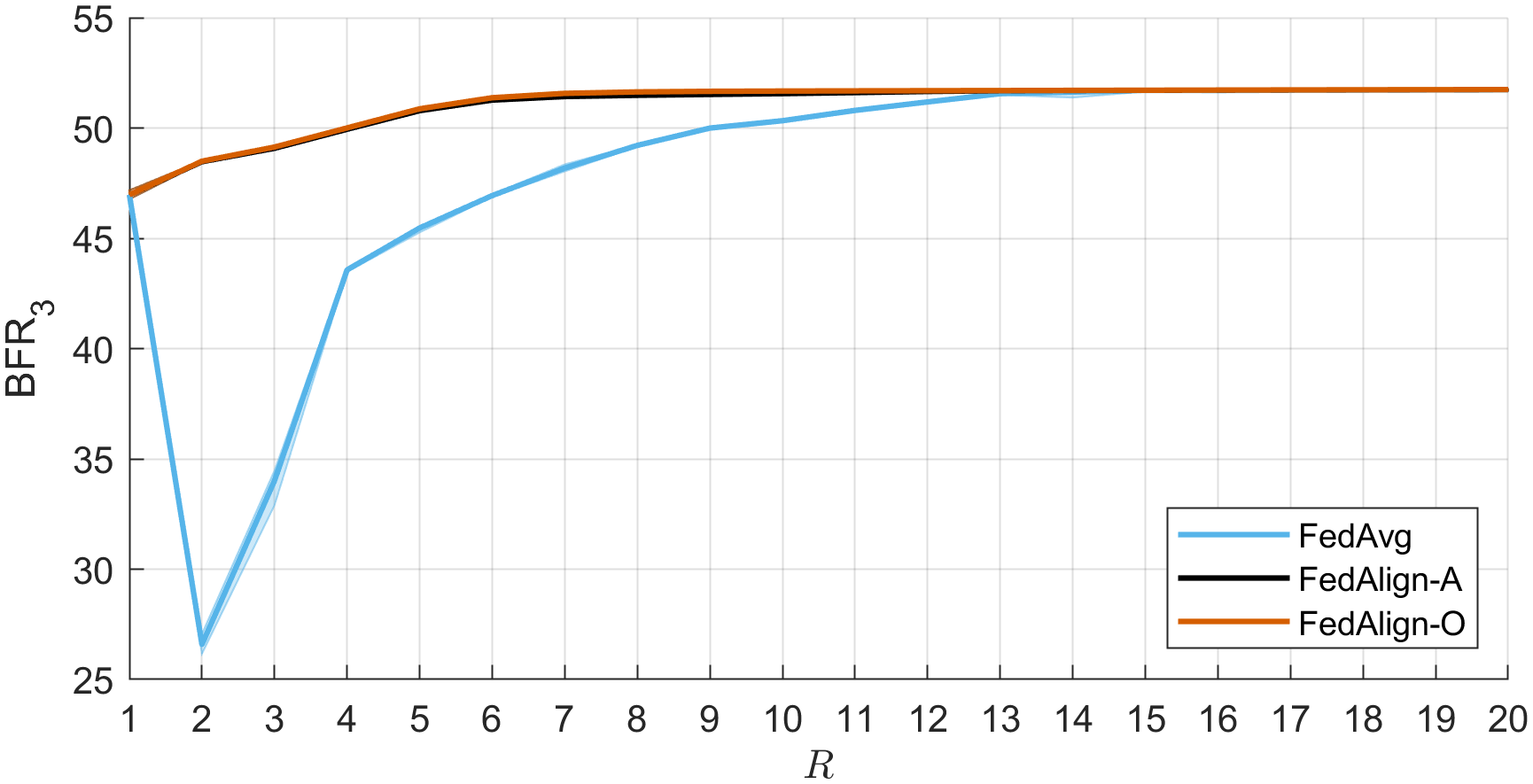}
        \end{minipage}
    }
    \hfill
    \subfloat[FedAlign for $iter=1$, FedAvg for $iter=20$\label{fig:evaporatorlfit-right}]{%
        \begin{minipage}{0.48\textwidth}
            \centering
            \includegraphics[width=\textwidth]{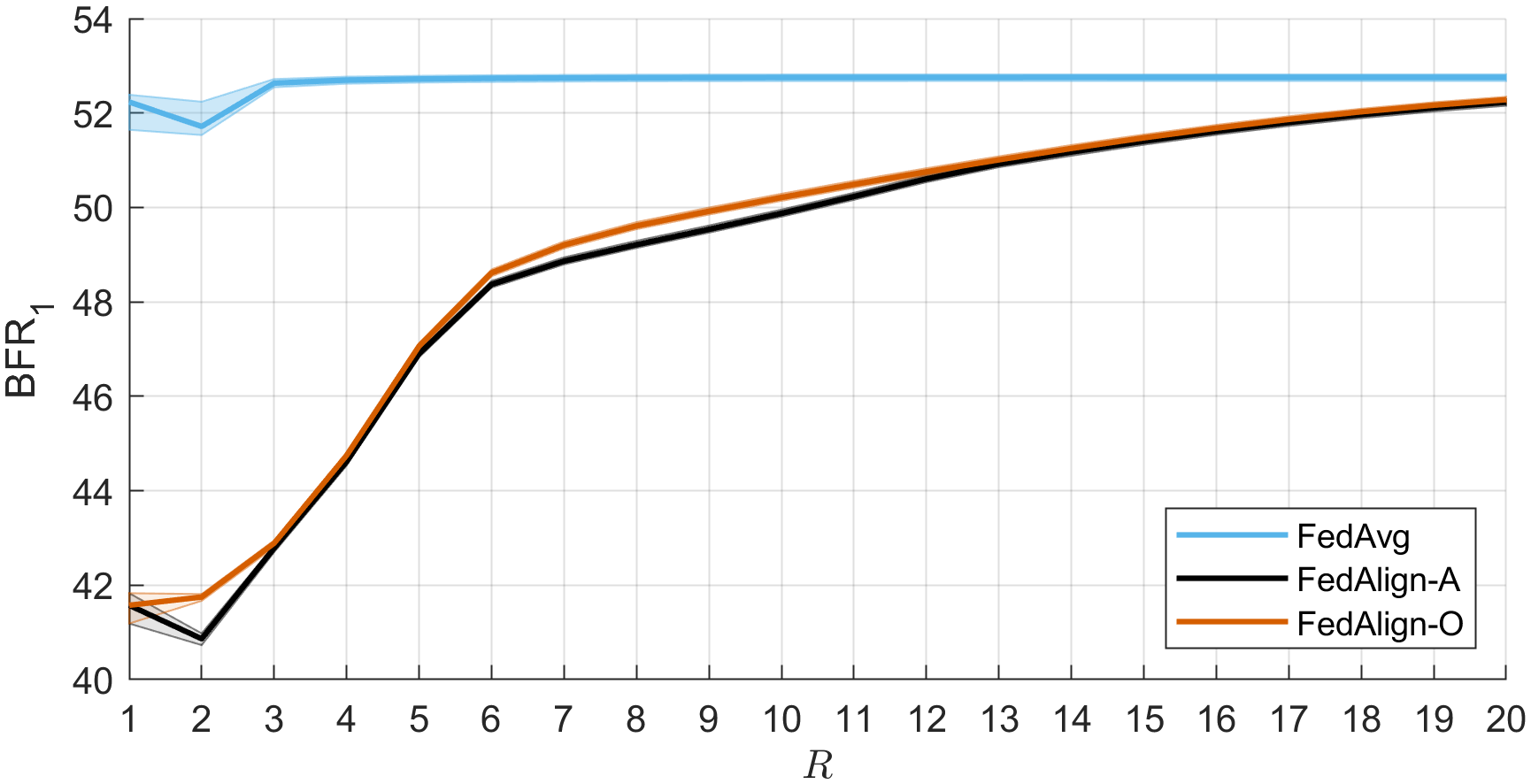}\\
            \vspace{1mm}
            \includegraphics[width=\textwidth]{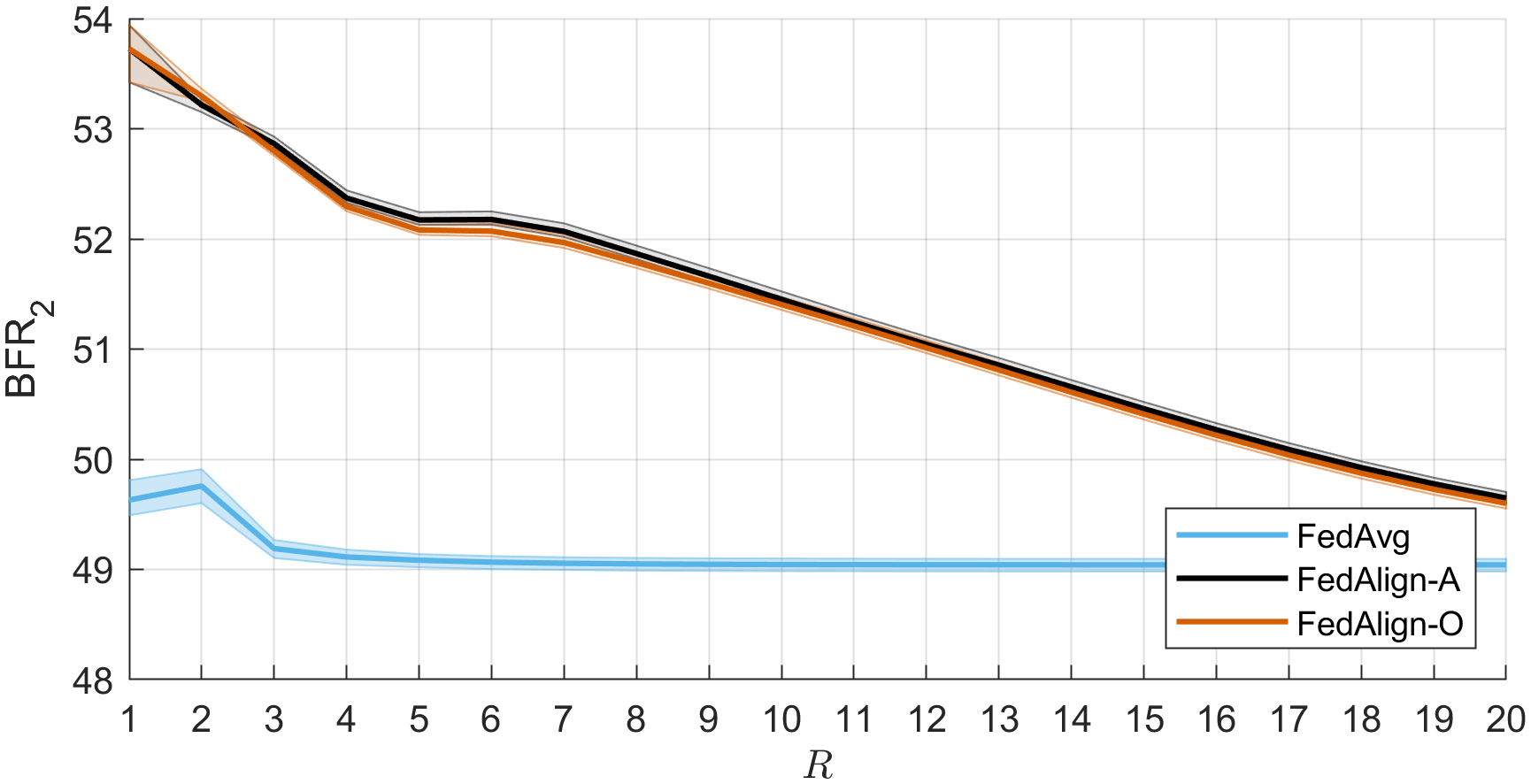}\\
            \vspace{1mm}
            \includegraphics[width=\textwidth]{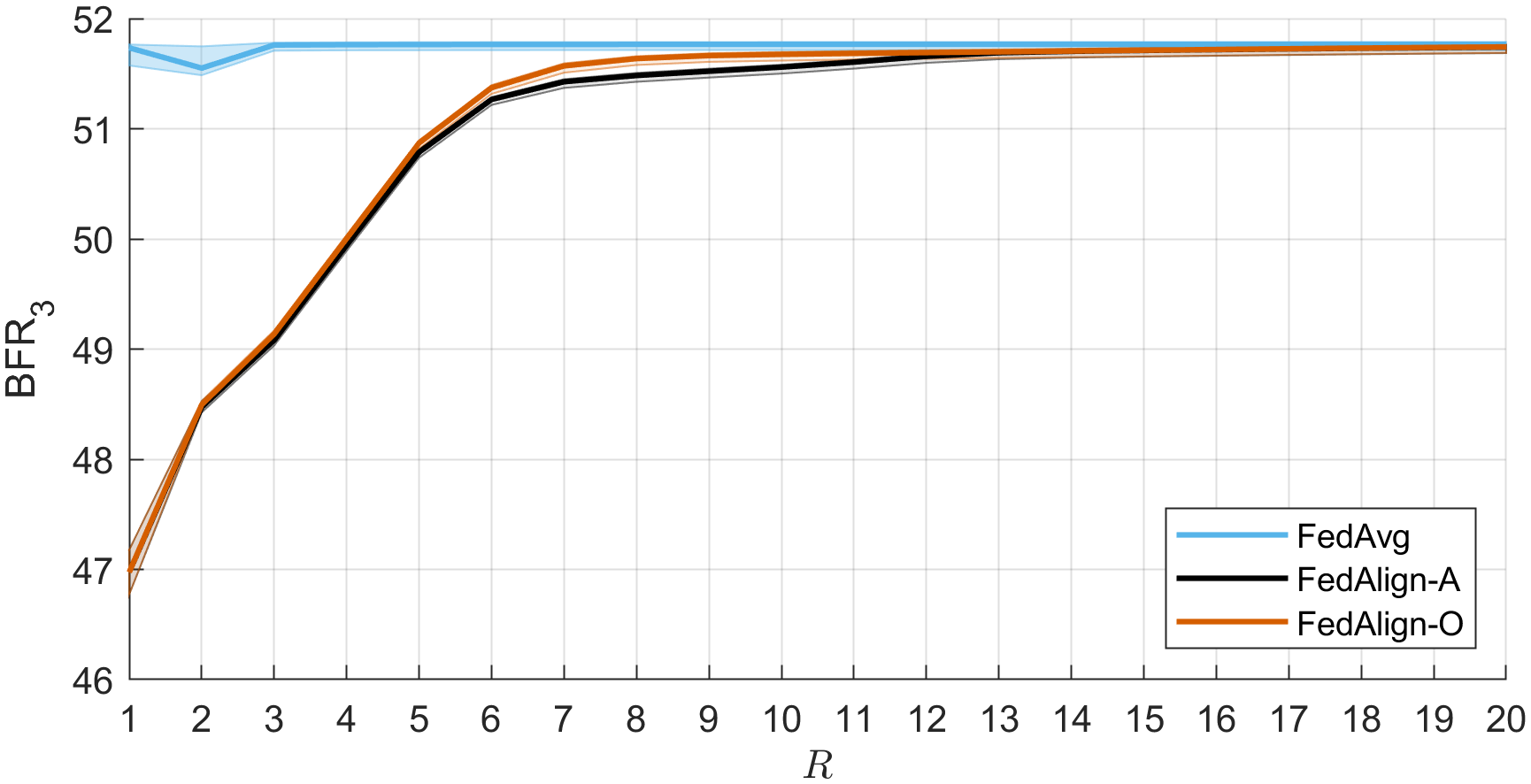}
        \end{minipage}
    }
    \caption{Comparison of FedAlign and FedAvg during training on Evaporator dataset: mean $\text{BFR}^{(i)}$ across local workers with solid lines while minimum, and maximum $\text{BFR}^{(i)}$ across local workers with shaded areas. Each row shows a different output: \(y_1\) (top row), \(y_2\) (middle row), and \(y_3\) (bottom row).}
    \label{fig:evaporatorlfit}
\end{figure}

Table \ref{tab:benchmarkMIMO} reports the mean \(\text{BFR}\) values with standard errors, \#UM and \#F2L. Moreover, we evaluated the global SSM’s SYSID performance against unseen data by evaluating \(\text{BFR}\) using \(D_{\text{test}}\). We demonstrated box plots $\text{BFR}^{(i)}$ values of FedAvg and FedAlign in Fig. \ref{fig:steamEngbox1} - Fig. \ref{fig:steamEngbox2} while we presented the $\text{BFR}^{(i)}$ during the communication rounds in Fig. \ref{fig:steamEnglfit} - Fig. \ref{fig:evaporatorlfit}. We observed that: 
\begin{itemize}
    \item For $iter = 1$, FedAlign evaluates higher BFR values with smaller deviations than FedAvg on the Steam Engine and CD Player datasets. As seen in Fig. \ref{fig:steamEngbox1} and Fig. \ref{fig:cdPlayerbox}, the outliers reduce the performance of FedAvg while increasing its variability. On the other hand, FedAvg performs on par with FedAlign on the Evaporator dataset, although FedAlign-O achieves the lowest standard deviation as shown in Fig. \ref{fig:evaporatorbox-left}.
    \item FedAvg shows only minor improvements compared to FedAlign for $iter=1$ on the Evaporator dataset, while slightly surpassing it on the Steam Engine dataset as illustrated in Fig. \ref{fig:evaporatorbox-right} and Fig. \ref{fig:steamEngbox2}. Nevertheless, FedAvg generates unstable global SSMs across all experiments on the CD Player dataset for $iter=20$.
    \item Figures \ref{fig:steamEnglfit} - \ref{fig:evaporatorlfit} illustrate that FedAvg experiences performance decreases at the initial communication rounds on all real-world MIMO datasets whereas FedAlign consistently preserves its performance, therefore providing faster convergence.  
    \item FedAvg yields unstable global SSMs in the CD Player and Evaporator datasets. On the contrary, FedAlign enables enhanced stability of the global SSMs by aligning local parameter basins.  
    \item Similar to the results on training data, FedAlign outperforms FedAvg against unseen data on all real-world MIMO datasets for $iter=1$. FedAvg only evaluates higher test BFR on the Steam Engine dataset with the help of increased local iterations ($iter=20$).
    \item FedAlign-A and FedAlign-O exhibit almost identical SYSID performance on all datasets, indicating both methods successfully establish a common parameter basin in the center server.

\end{itemize}

To conclude, FedAlign efficiently aligns local parameter basins and provides high SYSID performance against both training and test data. Moreover, due to its local parameter basin alignment, FedAlign enhances the stability of global SSM and converges faster.

\subsection{Statistical Comparison of FL-SYSID Performance} \label{stat}
    Here, we present the conducted Wilcoxon rank-sum tests to assess the statistical significance of the performance differences observed between the baselines and the proposed FedAlign variants. The tests were carried out using the test BFR values obtained from 20 independent experimental runs on the real-world SISO and MIMO datasets, as presented in Section \ref{sisocomp} and Section \ref{mimocomp}. Pairwise comparisons were made between FedAvg and each of the proposed FedAlign variants (FedAlign-A and FedAlign-O), as well as between FedAlign-A and FedAlign-O. For FedAvg, which has two configurations ($\texttt{iter}={1,20}$), we selected the configuration that yielded the best performance for each dataset. Experimental runs that resulted in unstable or failed-to-learn global SSMs were excluded from the analysis, as it has been done in descriptive statistics presented in Table \ref{tab:benchmark} and Table \ref{tab:benchmarkMIMO}. For the MIMO datasets, a single BFR score per experiment was computed by averaging the BFR values across all output channels to enable a consistent statistical comparison. We used a significance level of $p=0.05$ for all Wilcoxon tests. The resulting $p$-values are reported in Table \ref{tab:wilcoxon}. Based on the results, we observe that:  
    \begin{itemize}
        \item FedAlign-A and FedAlign-O demonstrate statistically significant performance improvement over FedAvg on most datasets. Only on the Piezoelectric dataset, FedAvg achieves a statistically significant advantage over FedAlign-A and FedAlign-O. 
        \item FedAlign-O shows statistically significant improvements over FedAlign-A on the MR Damper and Piezoelectric datasets. On the Hair Dryer, Steam Engine, and Evaporator datasets, their performances are statistically similar. In contrast, although the performances are nearly identical on the CD Player dataset, there is a statistically significant difference.     
    \end{itemize}
The results highlight the robustness of FedAlign variants across diverse SYSID scenarios, demonstrating statistically significant improvements and consistent performance.
\color{black}

\begin{table}[t]
\centering
\footnotesize
\renewcommand{\arraystretch}{1.4}
\setlength{\tabcolsep}{4pt}
\caption{Wilcoxon Rank-Sum Test \(p\)-Values Comparing Test BFR Values. † indicates statistical significance at \(p < 0.05\).}
\label{tab:wilcoxon}
\begin{tabular}{@{}l|c|c|c@{}}
\toprule
\textbf{Dataset} 
& \textbf{FedAvg vs. FedAlign-A} 
& \textbf{FedAvg vs. FedAlign-O} 
& \textbf{FedAlign-A vs. FedAlign-O} \\
\midrule
MR Damper     & 0.7180          & 0.0132$^\dagger$ & 0.0001$^\dagger$ \\
Hair Dryer    & 0.0001$^\dagger$ & 0.0001$^\dagger$ & 0.1417 \\
Piezoelectric & 0.0001$^\dagger$ & 0.0001$^\dagger$ & 0.0001$^\dagger$ \\
Steam Engine  & 0.0001$^\dagger$ & 0.0001$^\dagger$ & 0.1653 \\
CD Player     & 0.0012$^\dagger$ & 0.0029$^\dagger$ & 0.0350$^\dagger$ \\
Evaporator    & 0.0001$^\dagger$ & 0.0001$^\dagger$ & 0.8201 \\
\bottomrule
\end{tabular}
\end{table}
\color{black}

\section{Conclusion and Future Work} \label{conc}

In this paper, we propose FedAlign, an FL-SYSID framework designed to resolve alignment issues inherent in directly merging local SSMs via FedAvg. FedAlign overcomes these issues by aligning state representations of local SSMs through similarity transformation matrices. We developed two distinct methods in FedAlign to compute similarity transformation matrices: FedAlign-A, where we exploit control theory to analytically derive similarity transformation matrices, and FedAlign-O, which formulates the alignment problem as an optimization task to estimate similarity transformation matrices. Experiments conducted on various real-world SISO and MIMO datasets demonstrate that FedAlign improves SYSID performance, convergence speed, and stability of the global SSM even with fewer local iterations or reduced order modeling, underlining the effectiveness of the proposed FL-SYSID framework.

To sum up, while FedAlign has demonstrated significant improvements in SYSID, Table \ref{tab:fedalign_comparison} highlights several challenges and limitations that merit further investigation. In future work, we aim to address these challenges and extend the FedAlign framework to nonlinear state-space models based on neural networks, thereby enabling it to handle SYSID tasks for complex and hybrid systems. Furthermore, since FedAlign involves sharing local SSM parameters with a central server, it raises potential privacy concerns related to the exposure of sensitive system dynamics. To mitigate this risk, we plan to integrate privacy-preserving techniques—such as differential privacy \cite{mcmahan2017learning,truex2020ldp,wei2020federated}—into the FedAlign framework without compromising SYSID performance.

\begin{table}[t]
\centering
\footnotesize
\caption{Comparison of Challenges and Limitations of FedAlign Variants}
\label{tab:fedalign_comparison}
\renewcommand{\arraystretch}{1.3}
\begin{tabular}{@{}l p{5.5cm} p{5.5cm}@{}}
\toprule
\textbf{Aspect} & \textbf{FedAlign-A} & \textbf{FedAlign-O} \\
\midrule
Model Instability & Achieves global SSM stability by aligning state representations. Possible for MIMO systems due to numerical instabilities in the similarity matrices. & Ensures stable global SSM using state alignment, but imperfect optimization may still lead to unstable global models. \\
\cmidrule(lr){1-3}
Coordinate Sensitivity & Low for SISO systems due to unique CCF representation; sensitive for MIMO systems as the CCF representation is not unique. & Low, as it does not enforce strict representation. \\
\cmidrule(lr){1-3}
Aggregation Error & Low for SISO systems; may increase for MIMO systems with poor $\mu_\ell$ choice. & Generally low; may increase due to imperfect optimization. \\
\cmidrule(lr){1-3}
Computational Cost & Generally low due to analytical calculations, but complexity increases with the order of local SSMs. & Higher due to pseudo-data generation and solving optimization problems. \\
\cmidrule(lr){1-3}
Communication Overhead & Low; only local SSMs are transferred, with no additional burden. & Low; only local SSMs are transferred, with no additional burden. \\
\cmidrule(lr){1-3}
Real-time Deployment & Suitable due to fast and efficient analytical calculations. & Less suitable due to additional overhead from pseudo-data generation and optimization. \\
\cmidrule(lr){1-3}
Lack of Standardization & Low for SISO systems. Requires proper $\mu_\ell$ selection for MIMO systems; no established rules exist in classical control theory. & Present for both SISO and MIMO systems; pseudo-data generation and optimization setup may vary. \\
\bottomrule
\end{tabular}
\end{table}


\section*{Acknowledgment}

T. Kumbasar is supported by the BAGEP Award of the Science Academy. This study was funded by the Scientific Research Projects Commission of Istanbul Technical University, Project No: 47084.

\section*{Declaration of generative AI and AI-assisted technologies in the writing process}
During the preparation of this work, the authors used ChatGPT in order to refine the grammar and enhance the English language expressions. After using ChatGPT, the authors reviewed and edited the content as needed and take full responsibility for the content of the publication.



\bibliographystyle{elsarticle-num} 
\bibliography{bibliography}






\end{document}